%% file: main.tex
\documentclass[letterpaper,10pt,conference]{ieeeconf}

\IEEEoverridecommandlockouts %
\input{macros.tex}

\input{acronyms.tex}

\title{\LARGE \bf DYNUS: Uncertainty-aware Trajectory Planner \\ in Dynamic Unknown Environments}

\author{
    Kota Kondo, Mason Peterson, Nicholas Rober, Juan Rached Viso, Lucas Jia,  \\ 
    Jialin Chen, Harvey Merton, Jonathan P.\ How%
}

\begin{document}

\thispagestyle{plain}

\twocolumn[{%
\renewcommand\twocolumn[1][]{#1}%

\maketitle
\begin{center}
    \captionsetup{type=figure}
    \subfloat[UAV Hardware Experiment 4: We evaluate DYNUS in an environment with a dynamic obstacle. The UAV is equipped with a Livox Mid-360 LiDAR sensor and an Intel\textsuperscript{\texttrademark} NUC 13 for onboard computation. It successfully navigates the environment while avoiding the moving obstacle. All components (perception, planning, control, and localization) run onboard in real time.]{\includegraphics[width=\textwidth,trim=0 100 0 0,clip]{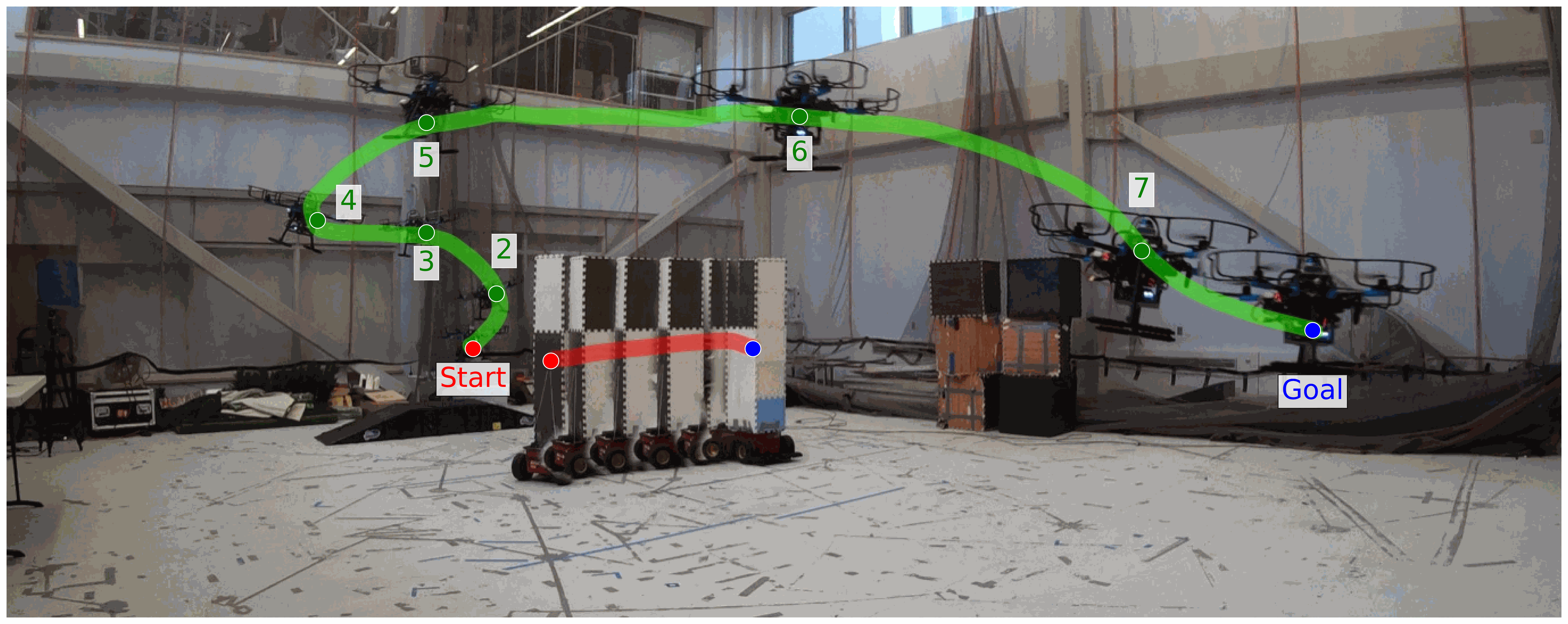}\label{fig:hardware_uav_dynamic_exp4}\vspace{0.5em}}
    \vspace{0.1em}
    \subfloat[Simulation: We run DYNUS in a confined office-like environment with multiple small rooms and dead ends. DYNUS successfully recovers from dead ends and traverses the building to reach the assigned goal. The trajectory is colored by speed, and the point cloud is visualized.]{\includegraphics[width=\textwidth]{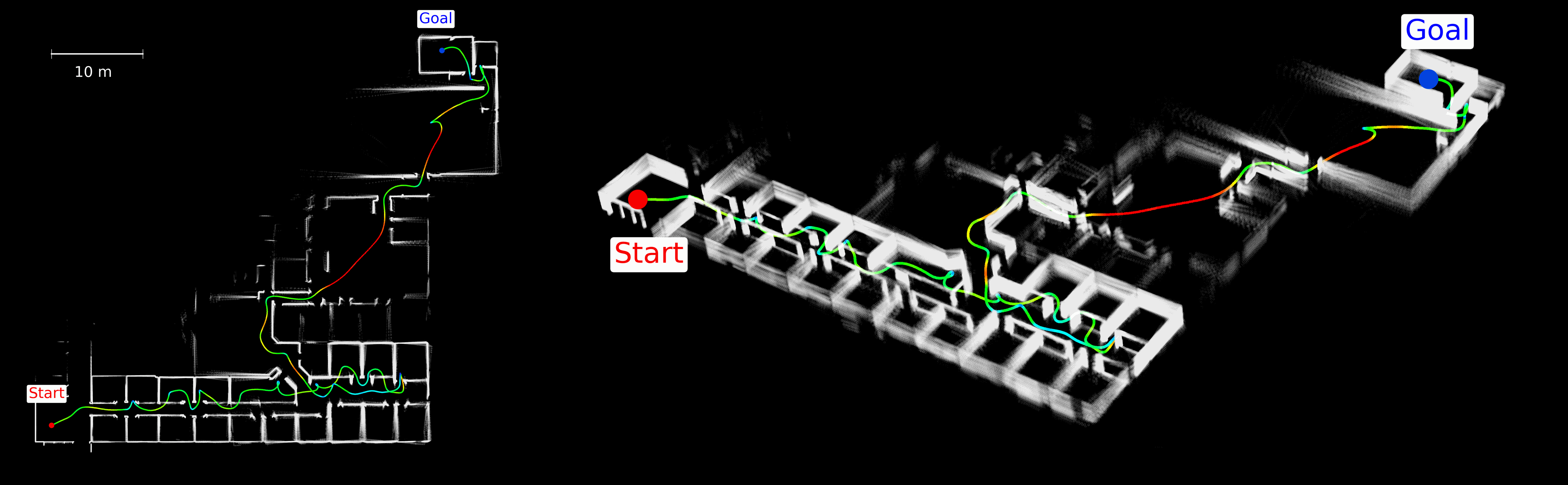}\label{fig:single_agent_performance_office_case3}
    \vspace{0.5em}}
    \caption{DYNUS is an uncertainty-aware trajectory planner capable of navigating in dynamic unknown environments.  
    (a) UAV hardware experiment in a dynamic environment.  
    (b) Simulation in a confined office space with small rooms and dead ends.}
    \label{fig:money_shot}
\end{center}
\vspace{-0.5em}
}]

\begingroup
\renewcommand\thefootnote{}\footnotetext{
The authors are with the Departments of Aeronautics and Astronautics, and Mechanical Engineering, Massachusetts Institute of Technology. \texttt{\{kkondo, masonbp, nrober, jrached, yixuany, jialinc7, hmer101, jhow\}@mit.edu.} This work is supported by DSTA.
}
\addtocounter{footnote}{-1}
\endgroup

\begin{abstract} 
\input{paper/abstract}
\end{abstract}

\section*{Supplementary Material}
\noindent\textbf{Video}: \href{https://youtu.be/SI8YbMS-wyw}{https://youtu.be/SI8YbMS-wyw} \\
\noindent\textbf{Code}: \href{https://github.com/mit-acl/dynus.git}{https://github.com/mit-acl/dynus.git}

\input{paper/01_intro}

\input{paper/02_dynus}
\input{paper/03_global_planner_and_safe_corridor_generation}

\input{paper/04_trajectory_optimization}

\input{paper/06_obst_tracking}
\input{paper/07_exploration}
\input{paper/10_simulation_results}

\input{paper/11_hardware_experiments}

\input{paper/12_conclusion}

\bibliographystyle{IEEEtran} %
\bibliography{main}

\end{document}

%% file: macros.tex
\PassOptionsToPackage{table}{xcolor} %

\usepackage{algorithm}
\usepackage{algpseudocode}

\usepackage{amsmath} %
\usepackage{amssymb} %
\usepackage{amsfonts}
\usepackage[sort,compress]{cite}
\usepackage[usenames,dvipsnames]{xcolor}
\usepackage{graphicx}
\usepackage{tabularx}
\usepackage{multirow}
\usepackage{mathtools}
\usepackage{esvect} %
\usepackage[]{siunitx}
\usepackage{caption}
\usepackage[nolist]{acronym}
\usepackage[pdftex,pdfstartview={FitV},pdfpagelayout={TwoColumnLeft},bookmarksopen=true,plainpages=false,colorlinks=true,linkcolor=black,citecolor=black,urlcolor=black,filecolor=black,pagebackref=false,hypertexnames=false,plainpages=false,pdfpagelabels]{hyperref}
\usepackage[T1]{fontenc} %
\usepackage{mathtools, cuted} %
\usepackage{bm}
\usepackage{xspace}
\usepackage{arydshln}
\usepackage{etoolbox}
\usepackage[normalem]{ulem}
\usepackage{empheq}
\usepackage{cleveref}
\usepackage{algorithm}
\usepackage{amsmath}
\usepackage{longtable}
\usepackage{textcomp}
\usepackage{svg}
\usepackage{graphicx}
\robustify\bfseries
\robustify\uline

\newcolumntype{H}{>{\setbox0=\hbox\bgroup}c<{\egroup}@{}}

\usepackage{enumerate}
\usepackage{balance} %
\usepackage{booktabs} %

\newcommand{\NoRed}{\textbf{\textcolor{red}{No}}}
\newcommand{\YesGreen}{\textbf{\textcolor{ForestGreen}{Yes}}}

\newcommand{\cg}{\cellcolor{LimeGreen!25} \YesGreen{}}
\newcommand{\crd}{\cellcolor{red!10} \NoRed{}}
\newcommand{\usetwod}{\cellcolor{red!10} \textbf{\textcolor{red}{2D}}}
\newcommand{\usethreed}{\cellcolor{LimeGreen!25} \textbf{\textcolor{ForestGreen}{3D}}}

\newcommand{\known}{\cellcolor{red!10} \textbf{\textcolor{red}{Known}}}
\newcommand{\unknown}{\cellcolor{LimeGreen!25} \textbf{\textcolor{ForestGreen}{Unknown}}}

\newcommand{\static}{\cellcolor{red!10} \textbf{\textcolor{red}{Static}}}
\newcommand{\dynamic}{\cellcolor{LimeGreen!25} \textbf{\textcolor{ForestGreen}{Both}}}
\newcommand{\open}{\cellcolor{red!10} \textbf{\textcolor{red}{Open}}}
\newcommand{\both}{\cellcolor{LimeGreen!25} \textbf{\textcolor{ForestGreen}{Both}}}

\newcommand{\best}[1]{\textbf{\textcolor{ForestGreen}{#1}}}

\usepackage{makecell}

\usepackage{amsmath,calc}

\usepackage{tikz} %
\usetikzlibrary{shapes,snakes} %
\usetikzlibrary{tikzmark} %
\usetikzlibrary{calc, arrows.meta, positioning} %
\usetikzlibrary{shapes.multipart} %
\usetikzlibrary{patterns}
\usepackage{subcaption} %

\usepackage{footnote}
\makesavenoteenv{tabular}
\makesavenoteenv{table}

\ifdefined\qtyproduct
\else
  \ifdefined\NewCommandCopy
    \NewCommandCopy\qtyproduct\SI
  \else
    \NewDocumentCommand\qtyproduct{O{}mm}{\SI[#1]{#2}{#3}}
  \fi
\fi

\DeclareCaptionLabelSeparator{periodspace}{.\quad}
\definecolor{opt_color}{RGB}{203,255,182}
\definecolor{check_color}{RGB}{195,218,255}
\definecolor{recheck_color}{RGB}{255,176,176}
\definecolor{delaycheck_color}{RGB}{255,228,181}

\definecolor{agent1_color}{RGB}{234,153,153}
\definecolor{agent2_color}{RGB}{151,104,175}
\definecolor{agent3_color}{RGB}{249,203,156}
\definecolor{agent4_color}{RGB}{194,115,156}
\definecolor{agent5_color}{RGB}{159,197,232}
\definecolor{agent6_color}{RGB}{117,186,117}
\definecolor{obstacle_color}{RGB}{160,117,109}

\usepackage{mdframed}

\newcommand{\bb}[1]{\boldsymbol{{#1}}}
\newcommand{\vect}[1]{\boldsymbol{{#1}}}

%% file: acronyms.tex
\newacro{slam}[SLAM]{Simultaneous Localization and Mapping}
\newacro{uav}[UAV]{Unmanned Aerial Vehicle}
\acrodefplural{uav}[UAVs]{unmanned aerial vehicles}
\newacro{gns}[GNS]{Global Navigation Satellite}
\newacro{gnss}[GNSS]{Global Navigation Satellite System}
\acrodefplural{gnss}[GNSS]{Global Navigation Satellite Systems}
\newacro{mcl}[MCL]{Monte-Carlo localization}
\newacro{imu}[IMU]{Inertial Measurement Unit}
\newacro{dof}[DOF]{degree-of-freedom}
\acrodefplural{dof}[DOFs]{degrees-of-freedom}
\newacro{ransac}[RANSAC]{random sample consensus}
\newacro{map}[MAP]{maximum a posteriori}
\newacro{mle}[MLE]{maximum likelihood estimation}
\newacro{rms}[RMS]{root-mean-square}
\newacro{dem}[DEM]{digital elevation model}
\newacro{vio}[VIO]{visual-inertial odometry}
\newacro{cnn}[CNN]{convolutional neural network}
\newacro{pdf}[pdf]{probability density function}
\newacro{ahrs}[AHRS]{attitude and heading reference system}
\newacro{lidar}[LIDAR]{light detection and ranging}
\newacro{relu}[ReLU]{rectified linear unit}
\newacro{rtk}[RTK]{real-time kinematic}
\newacro{gps}[GPS]{global positioning system}
\newacro{fcn}[FCN]{fully-connected network}
\newacro{brm}[BRM]{building ratio map}
\newacro{sfm}[SfM]{Structure-from-Motion}
\newacro{vpr}[VPR]{visual place recognition}
\newacro{fov}[FOV]{field of view}
\newacro{poc}[POC]{partially overlapping circular}

%% file: paper/abstract.tex
This paper introduces DYNUS, an uncertainty-aware trajectory planner designed for dynamic unknown environments. 
Operating in such settings presents many challenges{\textemdash}most notably, because the agent cannot predict the ground-truth future paths of obstacles, a previously planned trajectory can become unsafe at any moment, requiring rapid replanning to avoid collisions.
Recently developed planners have used soft-constraint approaches to achieve the necessary fast computation times; however, these methods do not guarantee collision-free paths even with static obstacles.
In contrast, hard-constraint methods ensure collision-free safety, but typically have longer computation times. 
To address these issues, we propose three key contributions. 
First, the \textit{DYNUS Global Planner (DGP) and Temporal Safe Corridor Generation} operate in spatio-temporal space and handle both static and dynamic obstacles in the 3D environment. 
Second, the \textit{Safe Planning Framework} leverages a combination of exploratory, safe, and contingency trajectories to flexibly re-route when potential future collisions with dynamic obstacles are detected. 
Finally, the \textit{Fast Hard-Constraint Local Trajectory Formulation} uses a variable elimination approach to reduce the problem size and enable faster computation by pre-computing dependencies between free and dependent variables while still ensuring collision-free trajectories.
We evaluated DYNUS in a variety of simulations, including dense forests, confined office spaces, cave systems, and dynamic environments. 
Our experiments show that DYNUS achieves a success rate of 100\% and travel times that are approximately 25.0\% faster than state-of-the-art methods. 
We also evaluated DYNUS on multiple platforms{\textemdash}a quadrotor, a wheeled robot, and a quadruped{\textemdash}in both simulation and hardware experiments.
\acresetall

%% file: paper/01_intro.tex
\section{Introduction}\label{sec:introduction}

Path and trajectory planning for autonomous navigation has been extensively studied~\cite{zhou2021ego-planner, zhou2021raptor, tordesillas2022faster,toumieh2024high,ren2025super,zhou2021ego-swarm, zhang2024soar, lee2024rapid, celestini2024transformer, zhang2024falcon, zhao2024learning, wang2024miner, wu2024trajectory, levy2024stitcher, xu2024navrl, bhattacharya2024monocular, liu2024lidar, huang2024safe, lehnert2024beyond, yu2025perception, su2025dynamic, Fan2025flying}.
In practical implementations of trajectory planning methods, it is crucial to avoid making overly strict prior assumptions about the environment, as these can limit the generalizability of an approach.
This work aims to develop a trajectory planner that utilizes a highly relaxed set of assumptions, which enables it to operate on a wide range of vehicles in a diverse set of environments.
The assumptions made by DYNUS are listed in Section~\ref{sec:system_overview}.
In Section~\ref{subsec:classification_of_environments}, we demonstrate that DYNUS is capable of operating in a wide range of environments.  
Sections~\ref{sec:lit_review_global_planning_and_safe_corridor} to~\ref{subsec:lit_review_frontier_exploration} review related work on global planning, local trajectory optimization, dynamic obstacle tracking, and exploration.  
We then summarize our key contributions.

\subsection{Classification of Environments}\label{subsec:classification_of_environments}

Environments are classified by three criteria: (1) known vs.\ unknown, (2) static vs.\ dynamic, and (3) open vs.\ confined.
\textit{Known environments} provide agents with prior information, while \textit{unknown environments} do not. 
\textit{Static environments} have fixed obstacles, whereas \textit{dynamic environments} could include moving ones. 
The primary difference is predictability: static obstacles do not change position once observed, while dynamic obstacles continuously move, making it a challenging planning environment.
Finally, \textit{open environments} are characterized by relatively sparse obstacle distributions, such as outdoor settings, whereas \textit{confined environments} involve occlusions and narrow spaces, such as indoor settings.

Table~\ref{tab:state_of_the_art_comparison} compares the capabilities of numerous UAV trajectory planners under these environmental assumptions.
State-of-the-art methods such as FASTER~\cite{tordesillas2022faster}, EGO-Planner~\cite{zhou2021ego-planner}, RAPTOR~\cite{zhou2021raptor}, HDSM~\cite{toumieh2024high}, and SUPER~\cite{ren2025super} are designed for unknown static environments. 
MADER~\cite{tordesillas2021mader} handles dynamic environments but requires the future trajectories of dynamic obstacles. 
PANTHER~\cite{tordesillas2022panther} can operate in dynamic unknown environments, but it generates convex hulls around each obstacle, and this approach is computationally expensive and less effective in environments with many static obstacles.
FHD~\cite{Fan2025flying} and STS~\cite{quan2025state} are the most recent works addressing dynamic unknown environments.  
FHD constructs a temporal Euclidean Signed Distance Field (ESDF) and plans around dynamic obstacles,  
while STS employs an end-to-end learning approach to navigate highly dynamic environments.  
However, both approaches assume a 2D environment and do not handle the challenge of planning in 3D dynamic unknown environments.
\vspace*{-0.05in}
\subsection{Global Planning and Safe Corridor Generation}~\label{sec:lit_review_global_planning_and_safe_corridor}
Many existing works implement a two-stage planning approach: global planning followed by local trajectory optimization.
Global planning generates a path that connects the start and goal positions, while local planning produces a dynamically feasible trajectory.
Hard-constraint safe corridor-based methods, such as FASTER~\cite{tordesillas2022faster}, use a global path to generate a safe corridor through convex decomposition techniques~\cite{liu2017planning}.
In contrast, soft-constrained methods often use a global path as an initial guess for trajectory optimization.

FASTER uses Jump Point Search (JPS)~\cite{harabor2011online} to generate a global path; however, it does not account for dynamic obstacles.
Several works have extended JPS, such as~\cite{traish2015optimization, zhao2023reducing}, which proposed JPS-based approaches for handling dynamic obstacles.  
In addition, D*~\cite{stentz1994optimal} has been developed as an extension of A* designed to improve its ability to handle dynamic obstacles.  
Although these methods address dynamic obstacles, they focus on obstacles that appear randomly during planning execution, which differs from the problem discussed in this paper.  
In our case, obstacles are detected and their future trajectories are predicted, and therefore the planner must generate a path that avoids these predicted obstacle trajectories in a spatio-temporal manner.

\begin{table*}[!t]
    \renewcommand{\arraystretch}{1.4}
    \scriptsize
    \begin{centering}
    \caption{\centering State-of-the-art UAV Trajectory Planners}
    \label{tab:state_of_the_art_comparison}
    \resizebox{1.0\textwidth}{!}{
    \begin{tabular}{>{\centering\arraybackslash}m{0.18\textwidth} 
                    >{\centering\arraybackslash}m{0.08\textwidth} 
                    >{\centering\arraybackslash}m{0.1\textwidth} 
                    >{\centering\arraybackslash}m{0.1\textwidth} 
                    >{\centering\arraybackslash}m{0.1\textwidth} 
                    >{\centering\arraybackslash}m{0.1\textwidth}
                    >{\centering\arraybackslash}m{0.1\textwidth}}
    \toprule 
    \multirow{2}{*}{\textbf{Method}} & \multirow{2}{*}{\makecell{\textbf{Uncertainty-} \\ \textbf{aware}}} & \multirow{2}{*}{\makecell{\textbf{Exploration} \\ \textbf{capability}}} & \multicolumn{4}{c}{\textbf{Environments}} \tabularnewline
    \cline{4-7}
    &&& \textbf{2D/3D} & \textbf{Known/Unknown} & \textbf{Static/Dynamic} & \textbf{Open/Confined} \tabularnewline
    \midrule
    \textbf{MADER} \cite{tordesillas2021mader} (2021) & \crd & \crd & \usethreed & \known & \dynamic & \open \tabularnewline
    \cline{0-0}
    \textbf{EGO-Planner} \cite{zhou2021ego-planner} (2021) & \crd& \crd & \usethreed & \unknown & \static & \both \tabularnewline
    \cline{0-0}
    \textbf{RAPTOR} \cite{zhou2021raptor} (2021) & \crd & \crd & \usethreed & \unknown & \static & \both \tabularnewline
    \cline{0-0}
    \textbf{FASTER} \cite{tordesillas2022faster} (2022) & \crd & \crd & \usethreed & \unknown & \static & \both \tabularnewline
    \cline{0-0}
    \textbf{PANTHER} \cite{tordesillas2022panther} (2022) & \crd & \crd & \usethreed & \unknown & \dynamic & \open \tabularnewline
    \cline{0-0}
    \textbf{HDSM} \cite{toumieh2024high} (2024) & \crd & \crd & \usethreed & \unknown & \static & \both \tabularnewline
    \cline{0-0}
    \textbf{SUPER} \cite{ren2025super} (2025) & \crd & \cg & \usethreed & \unknown & \static & \both \tabularnewline
    \cline{0-0}
    \textbf{FHD} \cite{Fan2025flying} (2025) & \crd & \crd & \usetwod & \unknown & \dynamic & \both \tabularnewline
    \cline{0-0}
    \textbf{STS} \cite{quan2025state} (2025) & \cg & \crd & \usetwod & \unknown & \dynamic & \both \tabularnewline
    \cline{0-0}
    \textbf{DYNUS (proposed)} & \cg& \cg & \usethreed & \unknown & \dynamic & \both \tabularnewline
    \bottomrule
    \end{tabular}}
    \par\end{centering}
    \vspace{-2em}
\end{table*}

\subsection{Local Trajectory Optimization}~\label{subsec:position_trajectory_optimization}
\subsubsection{Position Trajectory Optimization}~\label{subsec:lit_review_position_trajectory_optimization}
Local trajectory optimization generates a dynamically feasible trajectory that is either constrained within a safe corridor (if hard-constrained) or loosely follows the global path (if soft-constrained).
Many existing works employ spline-based trajectory optimization methods to produce smooth trajectories~\cite{zhou2019robust, zhou2021ego-planner, tordesillas2022faster, tordesillas2021mader, zhou2021raptor}.
One key challenge in trajectory optimization is the trade-off between safety and computational efficiency.
Many works~\cite{zhou2021ego-swarm,zhou2021ego-planner,ren2025super,zhou2021raptor} use soft-constrained methods.
While these methods are computationally efficient, they do not guarantee safety\textemdash collision avoidance is encouraged by the soft constraints, but not enforced.
On the other hand, hard-constrained methods, such as \cite{tordesillas2022panther, tordesillas2022faster,tordesillas2021mader, kondo2023robust_ral, toumieh2024high}, guarantee collision-free safety, but can be computationally expensive.

\subsubsection{Yaw Optimization and Uncertainty-awareness in the Direction of Motion}~\label{subsec:lit_review_yaw_optimization}
When exploring dynamic environments, tracking dynamic obstacles is crucial to avoid collisions.
There are two main approaches for yaw optimization: (1) coupled and (2) decoupled with position optimization.
Coupled approaches~\cite{tordesillas2022panther, falanga2018pampc, wu2024trajectory, kondo2024puma} optimize position and yaw trajectory simultaneously, allowing position optimization to be influenced by perception quality, which improves tracking of dynamic obstacles. However, this approach is known to be computationally expensive.
Decoupled approaches~\cite{zhou2021raptor,spasojevic2020perception} optimize position and yaw separately, reducing computation cost.
Operating in dynamic unknown environments requires agents account for the possibility of encountering dynamic obstacles in unobservable areas.
Thus, whether coupled or decoupled, one of the primary goals of yaw optimization must be to balance dynamic obstacle tracking with maintaining visibility in the direction of motion. 
Focusing solely on tracking dynamic obstacles, as in PANTHER~\cite{tordesillas2022panther}, may result in unexpected collisions with obstacles in the direction of motion, whereas prioritizing the direction of motion may lead to collisions with dynamic obstacles.
PUMA~\cite{kondo2024puma} balances the tracking of dynamic obstacles with looking in the direction of motion; however, its implicit tracking approach with its position-yaw coupled optimization results in a computationally expensive optimization.

\subsection{Dynamic Obstacle Estimation and Prediction}

There are various approaches to estimating dynamic obstacles' location and predicting their future trajectories.
Many works employ a Kalman filter-based approach combined with either a constant velocity or constant acceleration prediction model~\cite{tordesillas2022panther}, while others adopt learning-based methods~\cite{lee2017desire}.
When using a Kalman filter-based approach, the prediction model noise and measurement noise are often assumed to be known and fixed.
In contrast, learning-based approaches can learn the noise model from the data but may require a large amount of training data to achieve accurate predictions.
It is also important to note that prediction models cannot perfectly predict the future states of dynamic obstacles.
To address this, uncertainty-aware prediction methods~\cite{liu2024spf, zhou2024dynamic, chi2024raltper} can be used to account for errors in the prediction and estimation models of dynamic obstacles.

\subsection{Frontier-based Exploration}\label{subsec:lit_review_frontier_exploration}

Frontiers represent the boundary between known and unknown space, and frontier-based exploration is a method used to explore unknown areas.
The approach in~\cite{cieslewski2017rapid} selects frontiers to maximize the speed of exploration, while~\cite{zhou2021fuel} introduces a trajectory optimization approach that leverages a frontier information structure.
Similarly to~\cite{cieslewski2017rapid}, the work in~\cite{batinovic2021multi} employs an Octomap~\cite{hornung2013octomap}-based frontier detection method.
In recent years, several works~\cite{shi2024enhancing, zhou2024oto} have proposed frontier-based exploration techniques to tackle exploration in complex unknown environments.

\subsection{DYNUS Contributions}\label{subsec:dynus_contributions}

To operate safely in dynamic unknown environments, this paper presents DYNUS (\textbf{DYN}amic \textbf{U}nknown \textbf{S}pace) and outlines the following key contributions:

\begin{enumerate}\label{list:contributions}
    \item \textbf{DYNUS Global Planner (DGP) and Temporal Safe Corridor Generation:} A fast and safe global planner operating in spatio-temporal space and generating temporal safe corridors.
    \item \textbf{A safe local planning framework} that leverages an exploratory trajectory, safe trajectories, and a contingency plan to handle dynamic unknown environments.
    \item \textbf{Variable Elimination-Based Hard-Constraint Optimization Formulation:} An optimization technique that pre-computes dependencies between the free and dependent variables to reduce problem complexity while ensuring collision-free trajectories.
    \item \textbf{A yaw optimization method} that considers potential encounters with obstacles and balances  between tracking dynamic obstacles and maintaining visibility in the direction of motion. 
    \item \textbf{An efficient frontier-based exploration} approach.
    \item \textbf{Numerous simulations} demonstrating the effectiveness of DYNUS in various complex domains, including unknown, open, confined, static, and dynamic environments in both 2D and 3D.
    \item \textbf{Hardware experiments} on three different platforms: a quadrotor, a wheeled robot, and a quadruped robot.
\end{enumerate}

%% file: paper/02_dynus.tex
\section{DYNUS}\label{sec:dynus}

\begin{figure*}[htbp]
    \centering
    \includegraphics[width=\textwidth, trim=20 0 20 0, clip]{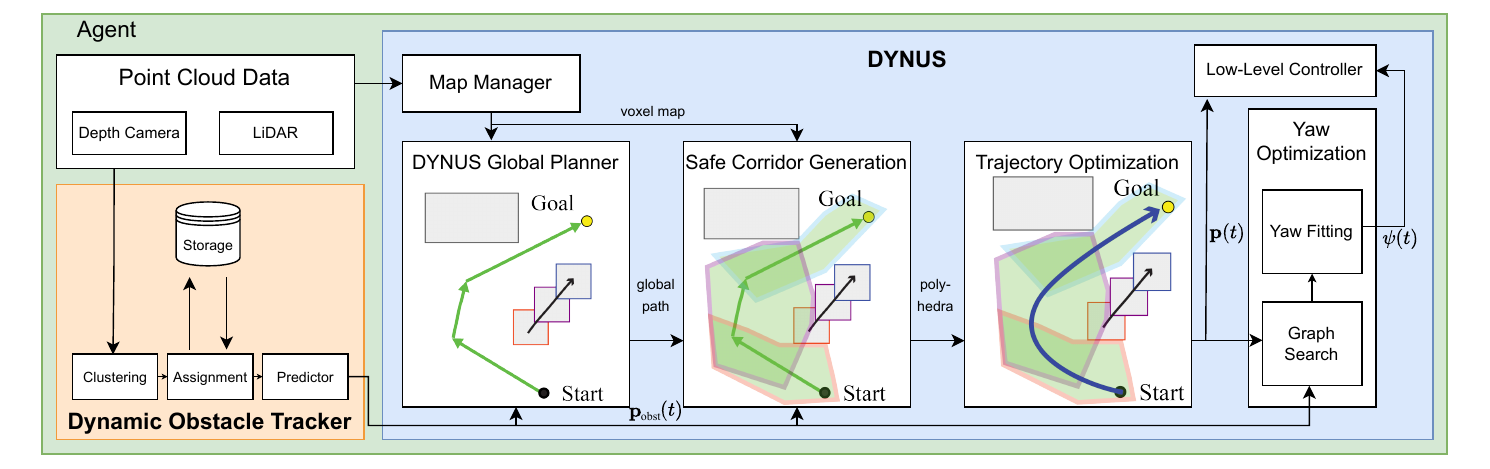}
    \caption{DYNUS system overview: Point cloud data from both the LiDAR and depth camera are processed by the Octomap-based Map Manager. Point cloud data is also sent to the Dynamic Obstacle Tracker, where dynamic obstacles are detected, clustered, and matched with previously observed obstacles. The AEKF estimates the current positions of dynamic obstacles, and a constant acceleration model is used to predict their future trajectories. 
    The voxel map and predicted future trajectories of dynamic obstacles ($\mathbf{p}_{\text{obst}}(t)$) are sent to the DGP and Safe Corridor Generation modules. 
    The DGP module computes a global path, and the Safe Corridor Generation module produces a series of overlapping polyhedra. 
    These polyhedra are then used in the Trajectory Optimization module to ensure the safety of the planned trajectory. 
    Note that the obstacles and safe polyhedra are colored red, purple, and blue according to their predicted time stamps. 
    The optimized trajectory ($\mathbf{p}(t)$) is passed to the Yaw Optimization module, where a sequence of yaw angles is generated by the Graph Search module and then smoothed using the Yaw Fitting module. 
    $\mathbf{p}(t)$ and $\mathbf{\psi}(t)$ are then sent to the low-level controller.}
    \label{fig:system_overview}
    \vspace{-2em}
\end{figure*}

This section provides an overview of DYNUS, as illustrated in Fig.~\ref{fig:system_overview}.  
Each component of DYNUS is designed to handle not only static but \textbf{dynamic unknown obstacles}, which requires:  
(1) planning in spatio-temporal space,  
(2) adapting trajectories flexibly as dynamic obstacles move unpredictably, and  
(3) computing safe trajectories quickly.

\subsection{DYNUS Global Planner (DGP)}

Traditional global planning algorithms such as JPS, A*, and RRT*~\cite{karaman2011rrt} can find paths in 3D static space but do not consider dynamic obstacles.  
To address this limitation, we first introduce \textbf{Dynamic A*}, a time-aware graph search algorithm that estimates both velocity and the time needed to reach each node during node expansion in graph search. 
These time estimates allow us to check for potential conflicts with dynamic obstacles. 
However, this time-aware approach can be computationally expensive, as it requires estimating velocity and travel time at every node expansion.  
Thus, to plan efficiently, while remaining robust to dynamic obstacles, our global planner, \textbf{DGP}, combines two components:  
(1) \textbf{JPS}, which is computationally efficient but cannot handle dynamic obstacles, and  
(2) \textbf{Dynamic A*}, which accounts for dynamic obstacles, but is computationally slow.  
DGP uses Dynamic A* only when necessary, reducing computation time without compromising safety.  
See Section~\ref{sec:global_planner_and_safe_corridor_generation} for further details on DGP.

\subsection{Trajectory Planning Framework}\label{subsec:trajectory_planning_framework}

\begin{figure}[htbp]
    \centering
    \begin{tikzpicture}[scale=1]
        \node[anchor=south west,inner sep=0] (image) at (0,0) {
            \includegraphics[width=\columnwidth, trim=0 0 0 0, clip]{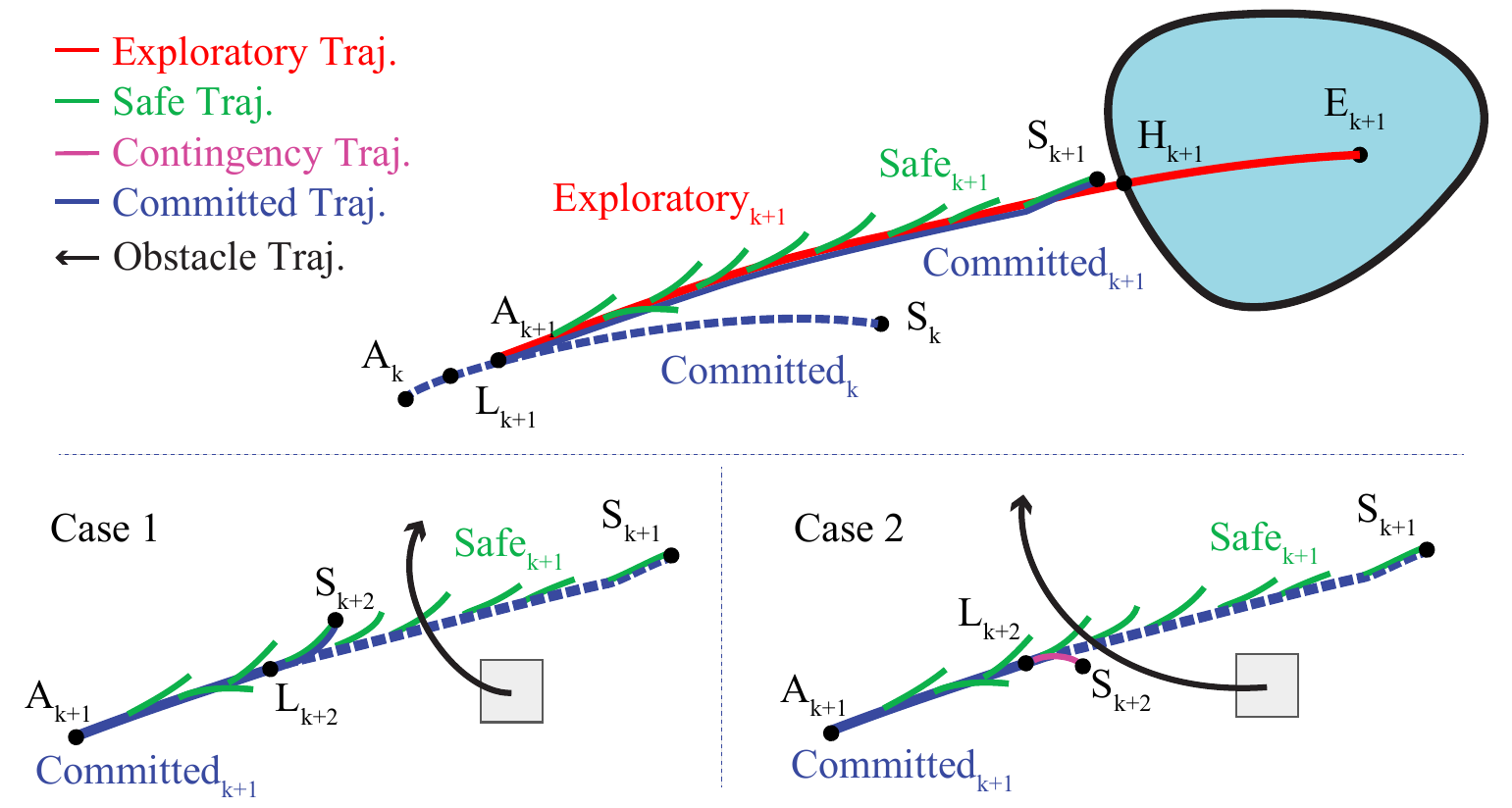}
        };
        \node at (7.2,3.3) {\small $\mathcal{U}$}; %
    \end{tikzpicture}
    \caption{Trajectory planning framework: $\mathcal{U}$ represents unknown space, 
    Point L is the current agent position, Point A is the replanning starting point, Point H is the point where the exploratory trajectory enters the unknown space, Point E is the goal for the exploratory trajectory, and Point S is the endpoint of the committed safe trajectory.  
    The subscripts denote the replanning iteration.
    The agent first generates an exploratory trajectory that may go into the unknown space $\mathcal{U}$.
    The agent then attempts to generate safe trajectories along the exploratory trajectory from Point A to H. 
    The agent initially commits to a trajectory that begins at point A as an exploratory trajectory, but switches to a safe trajectory that ends at point S.
    If a dynamic obstacle deviates from their predicted motion, they may collide with the committed future trajectory of the agent. 
    In such cases, the agent discards the previously committed trajectory and commits to the next available safe trajectory. 
    Case~1 illustrates this behavior: the agent, located at $L_{k+2}$ between the third and fourth safe trajectories, commits to the fourth safe trajectory.  
    In some situations, even the next available safe trajectory is in collision. 
    In such cases (Case~2), the agent generates a contingency trajectory from its current location ($L_{k+2}$), as detailed in Section~\ref{subsubsec:contingency_trajectory}.
}
    \label{fig:trajectory_planning_framework}
    \vspace{-2em}
\end{figure}

When planning a trajectory, an agent must consider three categories of space within its environment: unknown space, known-free space, and known-occupied space. 
Some approaches assume that an unknown space is free space and plan trajectories within it; however, this assumption could lead to collisions.
FASTER~\cite{tordesillas2022faster} and SUPER~\cite{ren2025super} instead plan a trajectory into unknown space but commit to a trajectory that is known to be safe to avoid the conservatism of planning only in safe known space.
Although this approach is efficient and ensures safety in a static world, it assumes that the known-free space remains free.
That assumption can be violated in dynamic unknown environments because the obstacles can move unpredictably, causing known-free space to suddenly become occupied.
To address this limitation, DYNUS introduces a novel planning framework that leverages three types of trajectories: exploratory, safe, and contingency.

Fig.~\ref{fig:trajectory_planning_framework} illustrates the trajectory planning framework.
The exploratory trajectory is generated from point $A$ using the MIQP-based method described in Section~\ref{subsec:trajectory_optimization_for_position}.
We then identify the point where the exploratory trajectory enters the unknown space (point $H$).
From $A$ to $H$, we sample points along the exploratory trajectory and generate safe trajectories that starts at these sampled points using a fast closed-form solution, which is described in Section~\ref{subsec:trajectory_optimization_for_position}.
We parallelize the computation of safe trajectories to reduce the computation time, which leads to fast computation times, as reported in Section~\ref{sec:simulation-results}.
We then commit to the trajectory that is the combination of the exploratory and safe trajectories that ends closest to $H$, denoted as Point $S$ in Fig.~\ref{fig:trajectory_planning_framework}. 
If we fail to find more than a certain number of safe trajectories, we discard the exploratory trajectory, since this means the environment does not allow the agent to have safe backup plans.
As the agent moves, we continuously check if the committed trajectory is still safe, and if not, we switch to the closest safe trajectory from the current position (Case 1 in Fig.~\ref{fig:trajectory_planning_framework}).
However, if a future potential collision is detected and there is no safe trajectory available (e.g., even the safe trajectory is now in collision), we immediately generate and commit to a contingency trajectory using the closed-form solution that avoids the collision (Case 2 in Fig.~\ref{fig:trajectory_planning_framework}).

As discussed in detail in Section~\ref{subsubsec:contingency_trajectory}, each of these safe and contingency trajectories can be generated in microseconds using our closed-form solution, making the framework suitable for real-time operation.
Note that DGP generates a global path that avoids both dynamic and static obstacles.  
Safe corridors (polyhedra) are then generated around this path, and the local trajectory planner generates trajectories that remain within these corridors, accounting for both types of obstacles.  
However, because dynamic obstacles are unpredictable and may change their motion at any moment, we designed our replanning framework, which enables DYNUS to flexibly adapt its trajectories.

\subsection{Local Trajectory Optimization}\label{subsec:local_traj_opt}

Since dynamic obstacles can change their motion unpredictably, fast trajectory optimization is needed.  
As discussed in Section~\ref{subsec:lit_review_position_trajectory_optimization}, soft constraint-based approaches, while computationally efficient, do not guarantee safety even with static obstacles.  
On the other hand, hard constraint-based approaches can guarantee safety with static obstacles but are typically slower.  
To address this trade-off, we introduce a \textit{variable elimination technique} that reduces the number of decision variables and constraints in the optimization problem, significantly accelerating hard-constraint-based optimization.  
Section~\ref{sec:trajectory_optimization} provides a detailed explanation of this method.

\subsection{Yaw Optimization}

The yaw optimization module consists of two components: yaw graph search and yaw fitting.  
First, we perform a graph search using utility values that consider factors such as collision probability and the time since each obstacle was last observed.  
This search produces a sequence of discrete yaw angles.  
Then, a B-spline fitting process is applied to generate a smooth yaw trajectory from these discrete values.  
The full algorithm is described in Section~\ref{subsec:yaw_optimization}.

\subsection{System Overview Summary}\label{sec:system_overview}

DYNUS processes point cloud data from both a LiDAR and a depth camera.  
The point cloud is processed using an Octomap-based Map Manager, which generates a voxel map in a sliding window as detailed in Section~\ref{subsec:map_representation}.  
Point cloud data is also used by the Dynamic Obstacle Tracker, where dynamic obstacles are detected, clustered, and associated with previously observed obstacles.  
As described in Section~\ref{sec:obstacle_tracking}, the Adaptive Extended Kalman Filter (AEKF) is applied within this module to estimate the current positions of dynamic obstacles, and a constant acceleration model is used to predict their future trajectories.

The voxel map and predicted obstacle trajectories ($\mathbf{p}_{\text{obst}}(t)$) are provided as inputs to the DYNUS Global Planner (DGP) and the Safe Corridor Generation module.  
The DGP computes a global path from the agent's start position to a subgoal, while the Safe Corridor Generation module generates a sequence of overlapping polyhedra.  
These polyhedra serve as constraints in the trajectory optimization process, ensuring collision-free paths.
The resulting position trajectory ($\mathbf{p}(t)$) is passed to the Yaw Optimization module.  
Initially, the Graph Search module generates a sequence of discrete yaw angles, which are then smoothed using B-spline fitting in the Yaw Fitting module, where yaw rate constraints are enforced.  
The position trajectory ($\mathbf{p}(t)$) and yaw trajectory ($\mathbf{\psi}(t)$) are then transmitted to the low-level controller for execution.  
Note that point cloud processing, dynamic obstacle tracking, map management, and DYNUS's planning modules are all parallelized.

Although DYNUS operates effectively in dynamic, unknown, open, and confined 3D environments, the framework relies on the following assumptions:
\begin{enumerate}
    \item Dynamic obstacles are non-adversarial.
    \item Trajectories generated by DYNUS, which satisfy dynamic constraints, can be tracked by the low-level controller.
    \item Static and dynamic obstacles can be detected by onboard sensors.
    \item The depth of any bug trap (dead end) is within the maximum range of the local map. Specifically, if a bug trap exceeds the local map's maximum sensing range, the agent may fail to recover, even if the global map captures the entire trap.
    \item Dynamic obstacles have relatively consistent motion patterns. As described in Section~\ref{sec:obstacle_tracking}, the process and sensor noise covariances for newly detected obstacles are initialized using the average covariance from previously observed obstacles. This implicitly assumes that new obstacles behave similarly to those encountered before. However, since these covariances evolve over time, the effect of this initialization diminishes. Alternatively, one could use predefined covariance values; however, this approach does not incorporate any information about the environment. Therefore, we choose to use the average of the covariances from previously encountered obstacles.
\end{enumerate}

%% file: paper/03_global_planner_and_safe_corridor_generation.tex
\section{Global planner and Safe Corridor}\label{sec:global_planner_and_safe_corridor_generation}

\subsection{Global Planner}\label{subsec:global_planner}

To handle static obstacles and predicted trajectories of dynamic obstacles in a computationally efficient manner, we propose the DYNUS Global Planner (DGP). 
DGP combines two global planning algorithms: Dynamic A* and JPS. 
We first introduce Dynamic A* and then discuss how DGP integrates these two methods.

\subsubsection{Dynamic A*}\label{subsubsec:dynamic_a_star}

The main challenge in global planning for \emph{dynamic} environments is to efficiently deconflict with the \emph{predicted} future positions of moving obstacles.  
We use a graph-search approach where each node is defined as
\[
    n_i = [\,x_i,\,y_i,\,z_i,\,v_{x,i},\,v_{y,i},\,v_{z,i},\,t_i,\,f_i,\,o_i\,]^{\mathsf T}
\]
encoding a 3D position, velocity, travel time $t_i$, A* cost $f_i$, and occupancy status $o_i$.  
Since each node specifies a unique position at a specific time, occupancy $o_i$ can be directly queried in the spatio-temporal space.  
Velocities $(v_{x,i},\,v_{y,i},\,v_{z,i})$ are included because different routes require different velocities, directly influencing travel times.  
The cost $f_i$, computed via the standard A* formulation (sum of travel and heuristic costs), determines node expansion priority.

\paragraph{Why travel‑time matters}
Because dynamic obstacles are timestamped, the occupancy of a location depends on \emph{when} the agent gets there.  
Hence, during expansion we must estimate the \emph{travel time} to a child node so that its timestamp~\(t_{i+1}\) is known before the collision check.

\paragraph{Axis‑wise time estimate}
Let \(p_i=[x_i,y_i,z_i]^{\mathsf T}\) and \(p_{i+1}=[x_{i+1},y_{i+1},z_{i+1}]^{\mathsf T}\), and then the displacement on each axis \(k\in\{x,y,z\}\) is
\[
    \delta_k = p_{k,i+1} - p_{k,i}, \qquad d_k = |\,\delta_k\,|
\]
Note that all computations below are performed \emph{independently on each axis}, and for clarity, we omit the axis subscript (e.g. $v_{k,i}$ is now denoted as $v_{i}$). 

To estimate travel time, we introduce a kinematics-based time estimation approach, which is incorporated into node expansion during the graph search. 
When a node is expanded, we estimate the time required to reach it from its parent.
We first show the simple example case.
If the initial velocity \(v_i\) at node \(i\) is positive ($v_i$ > 0), and also the node expansion is positive ($\delta$ > 0).
Then the agent's velocity at node \(i+1\) is computed by \(v_{i+1} = \sqrt{v_i^2 + 2a_{\text{max}}\,d}\)
and the travel time from node \(i\) to node \(i+1\) is given by \(\Delta t = (v_{i+1} - v_i) / a_{\text{max}}\).
If the agent is initially moving opposite to the desired direction (i.e., if the current velocity \(v_i\) and the displacement have opposite signs), the algorithm first computes the time required to decelerate to zero before accelerating in the intended direction (See Algorithm~\ref{alg:compute_node_time} for details).

We also consider \emph{cruising phase}. 
When the agent reaches the maximum velocity \(v_{\text{max}}\) we need to maintain this maximum velocity and compute travel time accordingly.
To this end, we first calculate the candidate velocity, \(v_{\text{cand}} = \sqrt{v_i^2 + 2a_{\text{max}}\,d}\), and we check if \(v_{\text{cand}}\) exceeds \(v_{\text{max}}\). 
If so, the motion is divided into two phases:
\begin{enumerate}
    \item \textbf{Acceleration Phase:} Accelerate from \(v_i\) to \(v_{\text{max}}\). 
    The distance required for this phase is \(d_{\text{accel}} = (v_{\text{max}}^2 - v_i^2)/(2a_{\text{max}})\), and the time for acceleration is \(t_{\text{accel}} = (v_{\text{max}} - v_i) / a_{\text{max}}\).
    \item \textbf{Cruising Phase:} Cover the remaining distance \(d_{\text{cruise}} = d - d_{\text{accel}}\)
    at a constant speed \(v_{\text{max}}\) with a time of \(t_{\text{cruise}} = d_{\text{cruise}} / v_{\text{max}}\).
\end{enumerate}
Thus, the total travel time is given by \(\Delta t = t_{\text{accel}} + t_{\text{cruise}}\) with the final velocity at the node updated to \(v_{i+1} \leftarrow v_{\text{max}}\). 
A similar approach is applied for negative displacements (\(\delta \leq 0\)) with appropriate sign adjustments. 
Finally, since the computation is executed independently for each of the three axes (\(x\), \(y\), and \(z\)), the overall travel time to a node $i+1$ from a node $i$ is the maximum of the three computed times:
\[
\Delta t = \max(\Delta t_x, \Delta t_y, \Delta t_z).
\]
This ensures that the estimated arrival times consider both the agent's kinematic capabilities and the imposed velocity limits.
Algorithm~\ref{alg:compute_node_time} provides the pseudocode for time estimation in Dynamic A* \textemdash to simplify the presentation, only the positive displacement case is presented. 
For negative displacements, similar logic applies with appropriate deceleration and reversal operations.

\begin{figure}[htbp]
    \vspace{-2em}
    \begin{algorithm}[H]
    \caption{Time Estimation in Dynamic A*}\label{alg:compute_node_time}
    \begin{algorithmic}[1]
        \State \textbf{Input:} Positions $p_i$ and $p_{i+1}$, velocity at node $i$ ($v_i$), maximum acceleration $a_{\text{max}}$, maximum velocity $v_{\text{max}}$, and tolerance $\epsilon$.
        \State \textbf{Output:} Travel time $\Delta t$ and updated velocity $v_{i+1}$.
        \State $d \gets |p_{i+1} - p_i|$, $\delta \gets p_{i+1} - p_i$ \small{\color{gray}{// Compute displacement.}}
        \If{$\delta > 0$} \small{\color{gray}{// Positive displacement.}}
            \If{$v_i \ge 0$} \small{\color{gray}{// Moving in the right direction.}}
                \If{$v_i \ge (v_{\text{max}} - \epsilon)$} \small{\color{gray}{// Already cruising near $v_{\text{max}}$.}}
                    \State $v_{i+1} \gets v_{\text{max}}$ \small{\color{gray}{// Set final velocity to the maximum.}}
                    \State $\Delta t \gets d/v_{\text{max}}$
                \Else
                    \State \small{\color{gray}{// Compute candidate velocity.}}
                    \State $v_{\text{cand}} \gets \sqrt{v_i^2 + 2\,a_{\text{max}}\,d}$
                    \If{$v_{\text{cand}} \le v_{\text{max}}$} \small{\color{gray}{// No cruising phase needed.}}
                        \State $v_{i+1} \gets v_{\text{cand}}$
                        \State $\Delta t \gets (v_{i+1} - v_i)/a_{\text{max}}$
                    \Else \small{\color{gray}{// Candidate velocity exceeds maximum}}
                        \State \small{\color{gray}{// Distance required to accelerate to $v_{\text{max}}$.}}
                        \State $d_{\text{accel}} \gets (v_{\text{max}}^2 - v_i^2)/(2\,a_{\text{max}})$
                        \State \small{\color{gray}{// Time to accelerate up to $v_{\text{max}}$.}}
                        \State $t_{\text{accel}} \gets (v_{\text{max}} - v_i)/a_{\text{max}}$
                        \State \small{\color{gray}{// If a cruise phase is allowed.}}
                        \If{$d > d_{\text{accel}}$} 
                            \State \small{\color{gray}{// Remaining distance at cruising speed.}}
                            \State $d_{\text{cruise}} \gets d - d_{\text{accel}}$
                            \State $t_{\text{cruise}} \gets d_{\text{cruise}}/v_{\text{max}}$ 
                            \State $\Delta t \gets t_{\text{accel}} + t_{\text{cruise}}$ \small{\color{gray}{// Total travel time.}}
                        \EndIf
                        \State $v_{i+1} \gets v_{\text{max}}$ \small{\color{gray}{// Final velocity set to $v_{\text{max}}$.}}
                    \EndIf
                \EndIf
            \Else
                \State \small{\color{gray}{// $v_{i} < 0$ case is omitted for brevity.}}
            \EndIf
        \EndIf
    \end{algorithmic}
    \end{algorithm}
    \vspace{-2em}
\end{figure}

\subsubsection{DYNUS Global Planner (DGP)}

Although Dynamic A* can handle dynamic obstacles, it requires computing the travel time whenever a node is expanded.
Since this can increase computation time significantly, we propose DGP.
The planner first considers only static obstacles and finds a path using JPS.
Next, it checks whether the JPS-generated path intersects with the predicted trajectory of dynamic obstacles.
If no collision is detected, the JPS path is returned since it is guaranteed to be collision-free.
However, if the path intersects with a predicted dynamic obstacle trajectory, the planner identifies the colliding node (node A) and the nearest subsequent collision-free node (node B).
Dynamic A* is then used to generate a subpath between node A and node B that avoids dynamic obstacles.
This new subpath is merged with the original JPS path, and the planner re-evaluates the entire path to check for potential collisions with dynamic obstacles.
This re-evaluation is necessary since the updated path may be longer than the original path, and new collisions could arise.
This process is repeated until a fully collision-free path is obtained, after which the planner returns the final path.
Unlike Dynamic A*, DGP does not need to consider dynamic obstacles at every search step, as dynamic obstacles are only accounted for when the JPS-generated path intersects with their predicted trajectories.

\subsubsection{Path Adjustment}\label{subsubsec:path_adjustment}

\begin{figure}
    \centering
    \subfloat{\includegraphics[width=0.32\columnwidth, trim=0 0 0 0, clip]{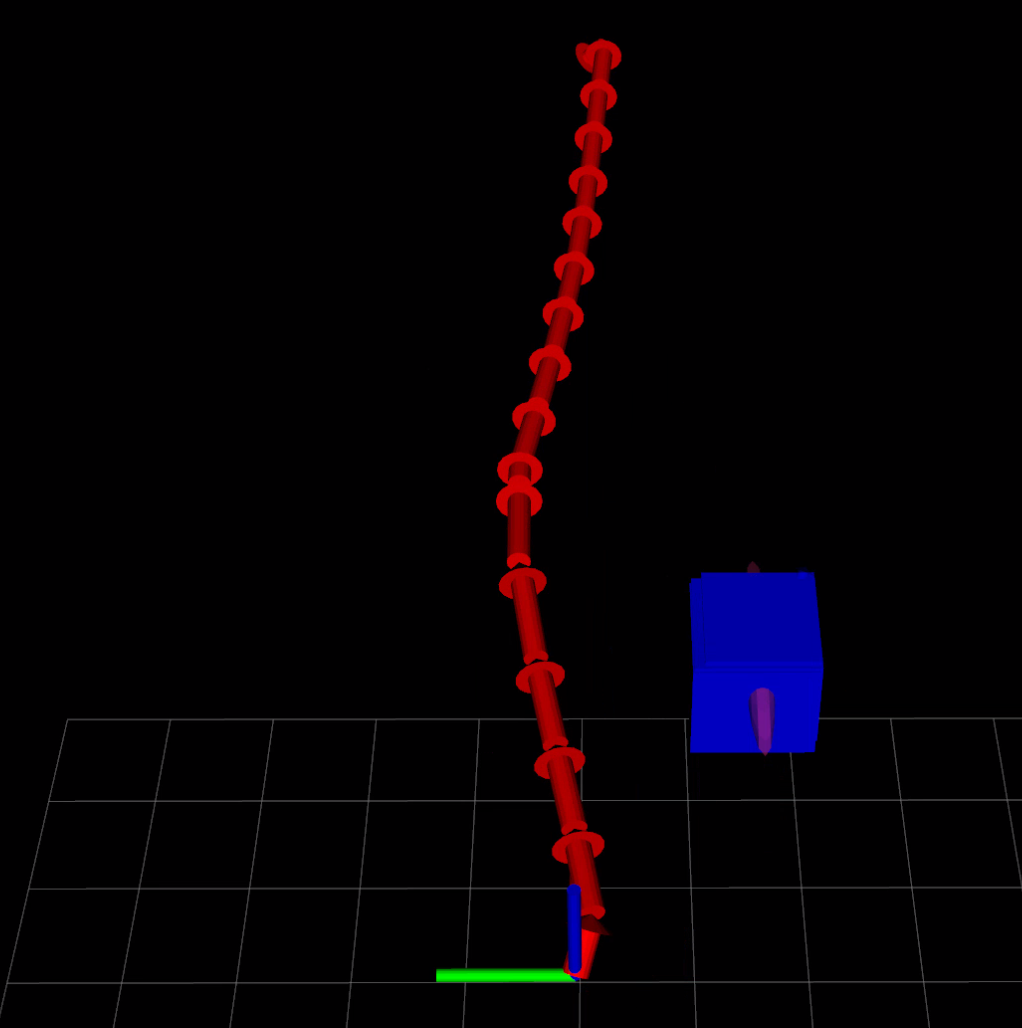}}
    \ 
    \subfloat{\includegraphics[width=0.32\columnwidth, trim=0 0 0 0, clip]{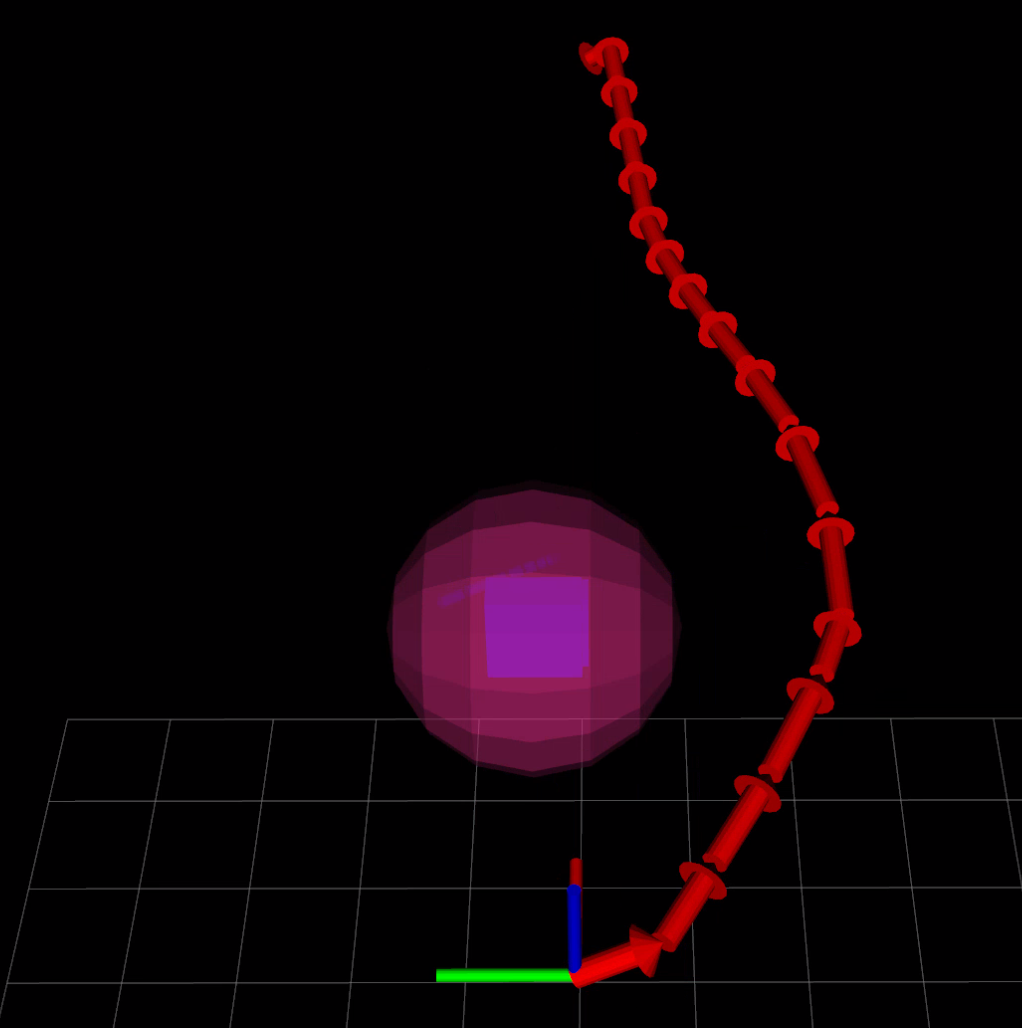}}
    \
    \subfloat{\includegraphics[width=0.32\columnwidth, trim=0 0 0 0, clip]{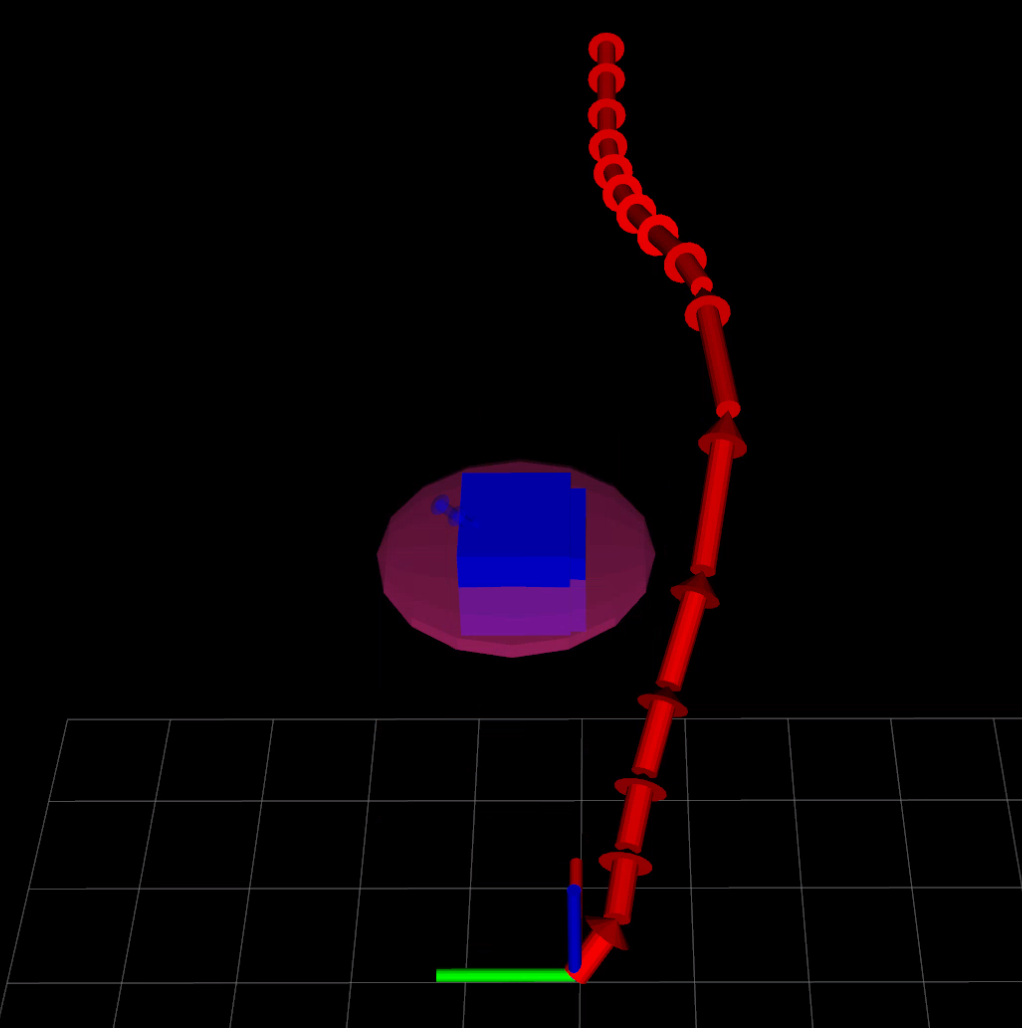}} \\
    \vspace{0.5em}
    \subfloat{\includegraphics[width=0.32\columnwidth, trim=0 0 0 0, clip]{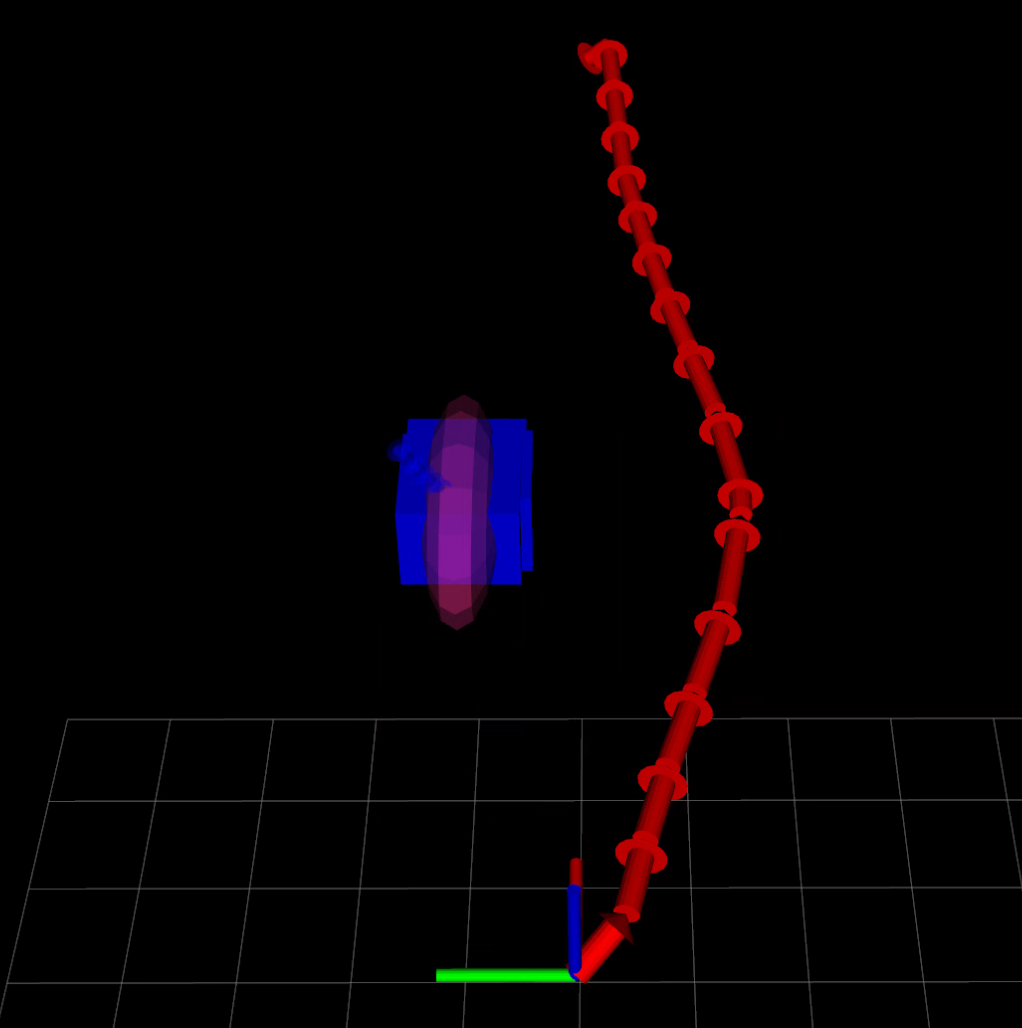}}
    \
    \subfloat{\includegraphics[width=0.32\columnwidth, trim=0 0 0 0, clip]{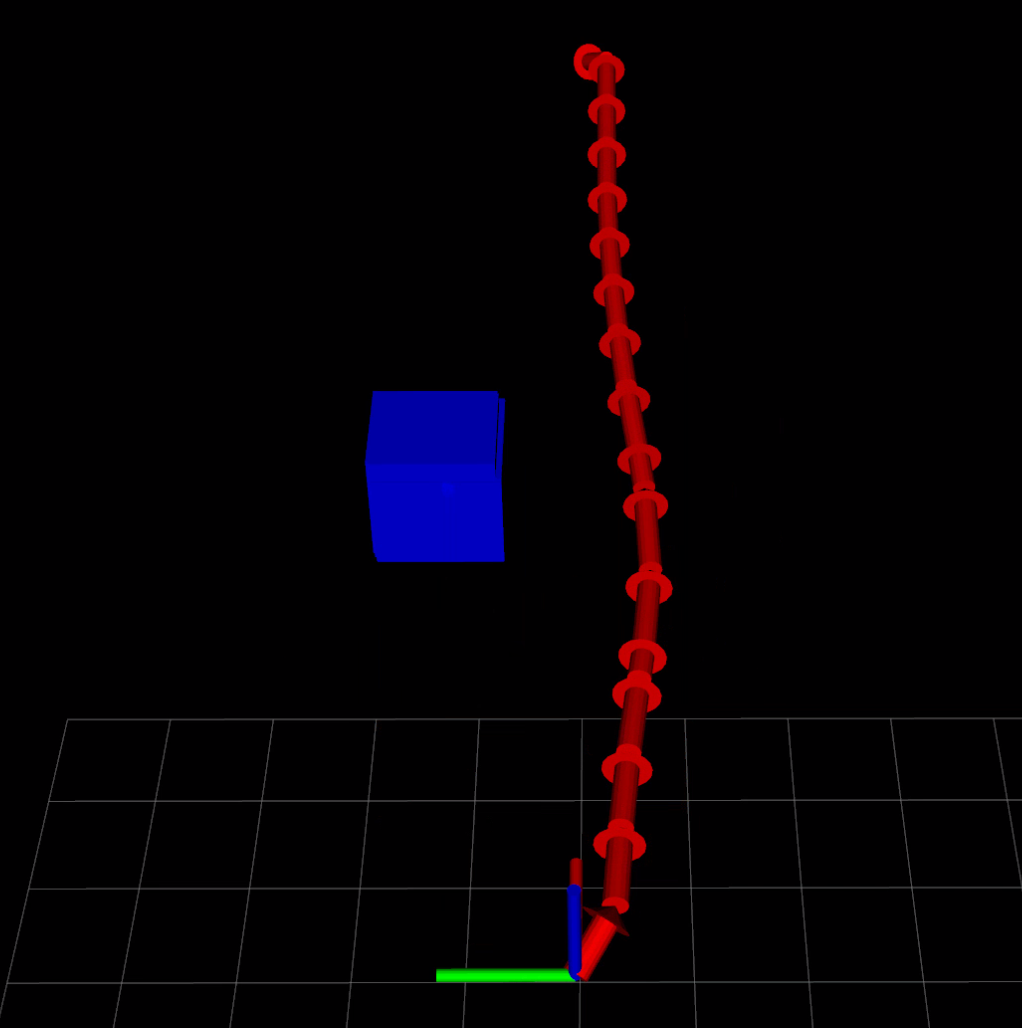}}
    \
    \subfloat{\includegraphics[width=0.32\columnwidth, trim=0 0 0 0, clip]{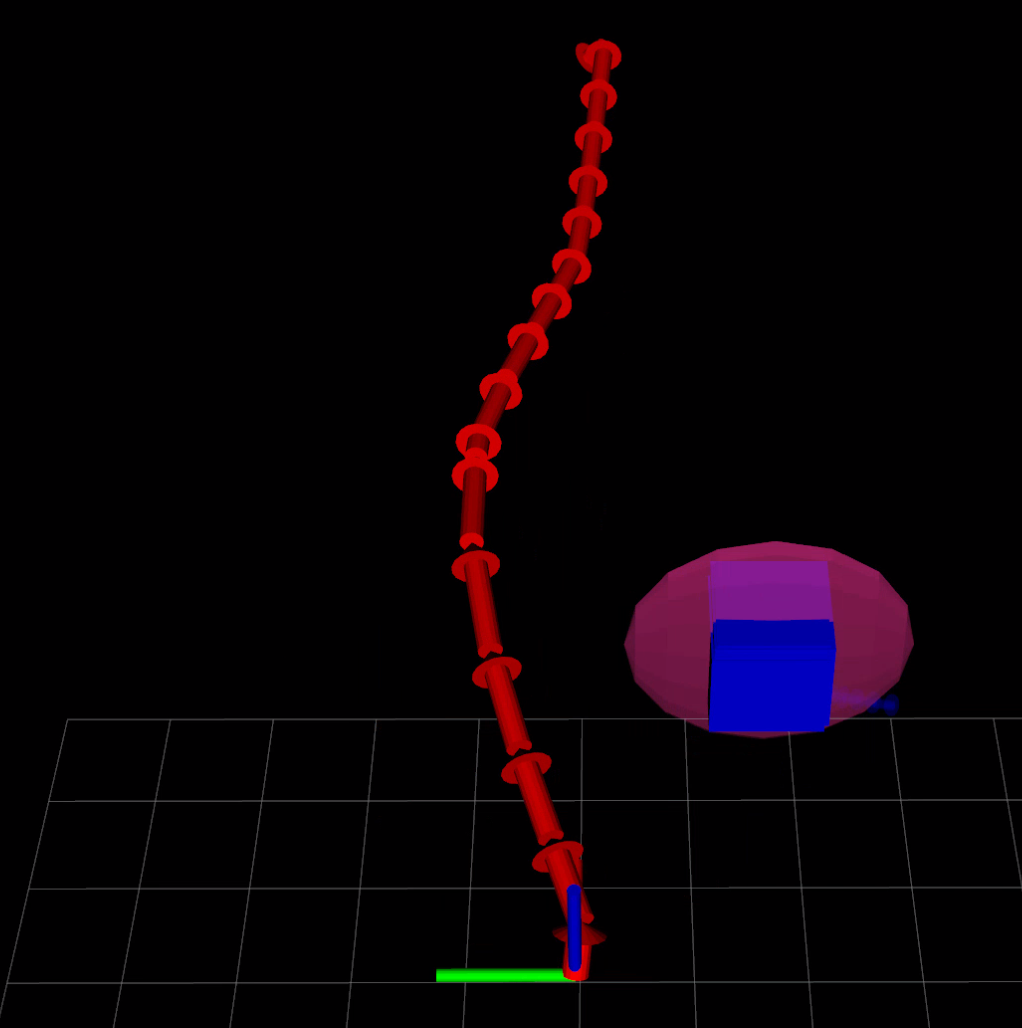}} \\
    \caption{Illustration of DGP with Dynamic Obstacles: The global path (red arrows) avoids dynamic obstacles.  
    The blue box represents a tracked dynamic obstacle, while the red ellipsoid indicates the estimation uncertainty.  
    The size of the ellipsoid along the $x$, $y$, and $z$ axes reflects the uncertainty magnitude \textemdash larger ellipsoids correspond to greater uncertainty, resulting in stronger repulsion forces.}
    \label{fig:dynamic_path_push_rviz}
    \vspace{-1em}
\end{figure}

To account for uncertainty in the prediction of dynamic obstacles and potential encounters with obstacles from behind occlusions, we propose a path adjustment algorithm that improves the visibility of unknown areas by pushing the global path away from both static obstacles and the predicted trajectories of dynamic obstacles.  
First, Algorithm~\ref{alg:dynamic_path_adjustment} summarizes the process of adjusting paths to account for dynamic obstacles, and Fig.~\ref{fig:dynamic_path_push_rviz} illustrates the path adjustment process.
The algorithm begins by initializing an adjusted path, copying the initial point of the global path. 
For each subsequent point in the global path, the algorithm evaluates the repulsion force based on the predicted uncertainty of obstacles, which is represented by the estimated covariance from the AEKF estimation module (Section~\ref{sec:obstacle_tracking}).  
Fig.~\ref{fig:dynamic_path_push_rviz} shows that as the uncertainty increases, the repulsion force becomes stronger, enforcing a larger safety distance from the obstacle.

\begin{figure}[htbp]
    \vspace{-2em}
    \begin{algorithm}[H]
        \caption{Dynamic Obstacle Path Adjustment}\label{alg:dynamic_path_adjustment}
        \begin{algorithmic}[1]
            \State \textbf{Input:} $\mathbf{\mathcal{P}}$ (global path)
            \State \quad \quad \quad $\mathcal{T}_{\mathrm{obs}}$ (set of predicted obstacle trajectories)
            \State \textbf{Output:} $\mathbf{\mathcal{P}}_{\text{new}}$ (adjusted new path)
            \State $\mathbf{\mathcal{P}}_{\text{new}} \gets [\mathbf{\mathcal{P}}[0]]$ \small{\color{gray}{// Initialize $\mathbf{\mathcal{P}}_{\text{A}}$}}
            \For{$i \gets 1$ to $\text{size}(\mathbf{\mathcal{P}}) - 1$} \small{\color{gray}{// Skip the first point}}
                \State $\mathbf{p} \gets \mathbf{\mathcal{P}}[i]$ \small{\color{gray}{// Current global path point}}
                \State $\mathbf{p}_{\mathrm{new}} \gets \mathbf{\mathcal{P}}[i]$ \small{\color{gray}{// Initialize adjusted point}}
                \For{$\tau \in \mathcal{T}_{\mathrm{obs}}$} \small{\color{gray}{// Iterate through each obstacle trajectory}}
                    \State \small{\color{gray}{// Compute push force}}
                    \State \small{\color{gray}{// $k$: const., $\alpha_{P}$: scaling factor, $\tau.P$: estimate covariance}}
                    \State $F_{\mathrm{push}} \gets k + \alpha_{P} \cdot \|\tau.P\|$ 
                    \State $\mathbf{o} \gets \tau.\text{pos}(t_i)$ \small{\color{gray}{// Obstacle position at $t=t_i$}}
                    \State $\mathbf{d} \gets \mathbf{p} - \mathbf{o}$ \small{\color{gray}{// Direction vector to obstacle}}
                    \State $r \gets \|\mathbf{d}\|$ \small{\color{gray}{// Distance to obstacle}}
                    \If{$r < C$} \small{\color{gray}{// Check if within collision clearance}}
                        \State \small{\color{gray}{// Compute repulsion force}}
                        \State $\mathbf{f}_{\mathrm{rep}} \gets F_{\mathrm{push}} \cdot \big(1 - \frac{r}{C}\big) \cdot \frac{\mathbf{d}}{r}$ 
                        \State $\mathbf{p}_{\mathrm{new}} \gets \mathbf{p}_{\mathrm{new}} + \mathbf{f}_{\mathrm{rep}}$ \small{\color{gray}{// Update adjusted point}}
                    \EndIf
                \EndFor
                \State \small{\color{gray}{// Append adjusted point to adjusted new path}}
                \State $\mathbf{\mathcal{P}}_{\text{new}}.\mathrm{append}(\mathbf{p}_{\mathrm{new}})$
            \EndFor
            \State \Return $\mathbf{\mathcal{P}}_{\text{new}}$
        \end{algorithmic}
    \end{algorithm}
    \vspace{-2em}
\end{figure}

\begin{figure}
    \centering
    \subfloat[\centering\label{fig:static_push_0}]{\includegraphics[width=0.32\columnwidth, trim=15 25 15 45, clip]{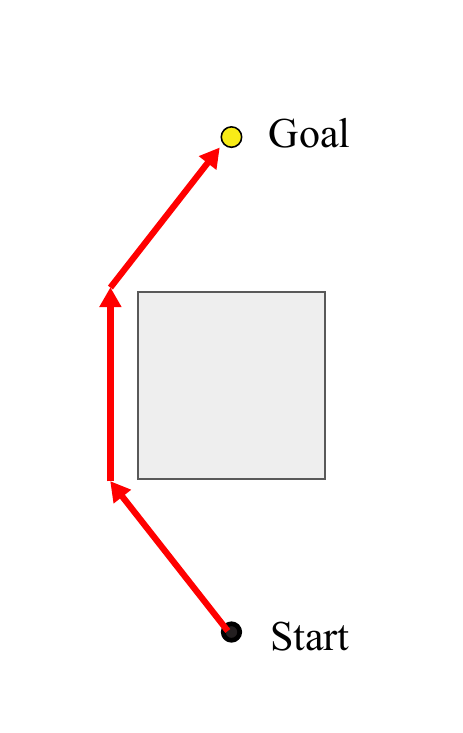}}
    \ 
    \subfloat[\centering\label{fig:static_push_1}]{\includegraphics[width=0.32\columnwidth, trim=15 25 15 45, clip]{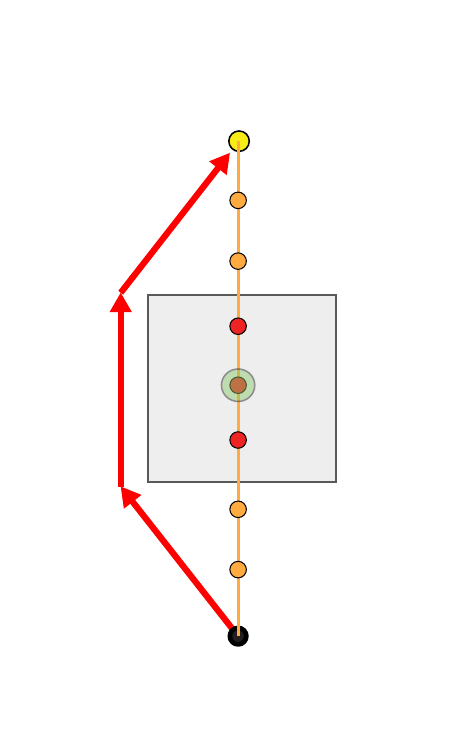}}
    \
    \subfloat[\centering\label{fig:static_push_2}]{\includegraphics[width=0.32\columnwidth, trim=15 25 15 45, clip]{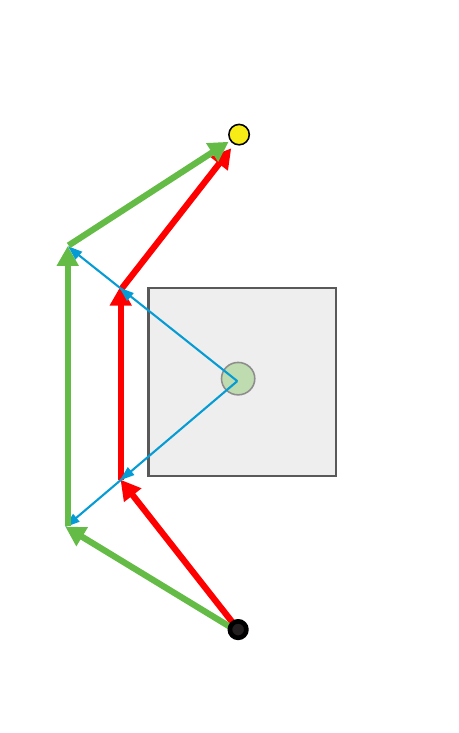}}
    \\
    \subfloat{\includegraphics[width=0.98\columnwidth, trim=13 5 0 5, clip]{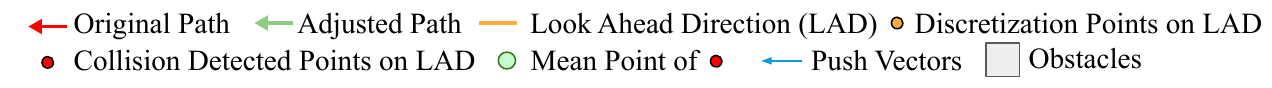}}
    \setcounter{subfigure}{0}
    \caption{Static Obstacle Trajectory Push: (a) Original global path (red) from DGP. (b) Look-ahead direction, its discretized points, and the detected occupied points and their mean position. (c) Final adjusted path, pushed away from the obstacle. This mean position is stored and used to push the path at the next iteration.}
    \label{fig:static_path_adjustment}
    \vspace{-2em}
\end{figure}

We now describe the path adjustment process for static obstacles.  
Although the path generated by DGP is collision-free, it may pass close to static obstacles, limiting the visibility of unknown areas.
To address this issue, we propose a path adjustment algorithm that pushes the global path away from static obstacles, improving the visibility of previously unknown areas.
Fig.~\ref{fig:static_path_adjustment} illustrates the static obstacle path adjustment algorithm (see Algorithm~\ref{alg:static_path_adjustment}).  
As shown in Fig.~\ref{fig:static_push_1}, the adjustment process begins by finding a straight line that connects the start and \( N_{\text{LAD}} \)-th points of the global path, where \( N_{\text{LAD}} \) denotes the number of look-ahead discretization points.  
The algorithm then discretizes this path and checks for occupied points in the static map, and if occupied points are detected, the mean position of these points is computed. 
Lastly, the points on the original global path are pushed away from this mean position by a fixed distance (Fig.~\ref{fig:static_push_2}).  
After adjusting the path, DGP verifies that the updated trajectory remains within known-free space, and if not, the push distance is decreased until a collision-free path is obtained.
This mean position of the occupied points is stored and used to push the path at the next iteration as well.

\begin{figure}[htbp]
    \vspace{-2em}
    \begin{algorithm}[H]
        \caption{Static Obstacle Path Adjustment}\label{alg:static_path_adjustment}
        \begin{algorithmic}[1]
            \State \textbf{Input:} $\mathbf{\mathcal{P}}$ (global path)
            \State \quad \quad \quad $d_{\mathrm{disc.}}$ (discretization distance)
            \State \textbf{Output:} $\mathbf{\mathcal{P}_{\text{new}}}$ (adjusted new path)
            \State \small{\color{gray}{// Find the path connecting the start and $N_{\text{LAD}}$-th points}}
            \State $\mathbf{p}_{\mathrm{start}} \gets \mathbf{\mathcal{P}}[0]$, $\mathbf{p}_{\mathrm{LAD}} \gets \mathbf{\mathcal{P}}[N_{\text{LAD}}]$, $\mathbf{d} \gets \mathbf{p}_{\mathrm{LAD}} - \mathbf{p}_{\mathrm{start}}$
            \State \small{\color{gray}{// Discretize the path and check for occupied points}}
            \State $n_{\mathrm{steps}} \gets \lceil \|\mathbf{d}\| / d_{\mathrm{disc.}} \rceil$, $\mathbf{s} \gets \mathbf{d} / n_{\mathrm{steps}}$
            \For{$j \gets 0$ to $n_{\mathrm{steps}}$}
                \State $\mathbf{p} \gets \mathbf{p}_{\mathrm{start}} + j \cdot \mathbf{s}$
                \If{$\mathbf{p}$ is Occupied}
                    \State $\mathbf{\mathcal{O}}.\mathrm{append}(\mathbf{p})$
                \EndIf
            \EndFor
            \small{\color{gray}{// Compute the mean position of occupied points}}
            \State $\mathbf{\bar{p}}_{\mathrm{mean}} \gets \frac{1}{|\mathbf{\mathcal{O}}|} \sum_{\mathbf{o} \in \mathbf{\mathcal{O}}} \mathbf{o}$
            \small{\color{gray}{// Compute push vectors}}
            \For{$i \gets 1$ to $|\mathbf{\mathcal{P}}|$}
                \State $\mathbf{\mathcal{V}}_{\mathrm{push}}.\mathrm{append}(\mathbf{\mathcal{P}}[i] - \mathbf{\bar{p}}_{\mathrm{mean}})$
            \EndFor
            \State \small{\color{gray}{// Push the path points away from $\mathbf{\bar{p}}_{\mathrm{mean}}$}}
            \For {$i \gets 0$ to $N_{\text{LAD}}$}
                \State $\mathbf{\mathcal{P}_{\text{new}}}[i] \gets \mathbf{\mathcal{P}}[i] + \alpha_{\mathrm{push}} \cdot \mathbf{\mathcal{V}}_{\mathrm{push}}[i]$
            \EndFor
            \State \Return $\mathbf{\mathcal{P}_{\text{new}}}$
        \end{algorithmic}
    \end{algorithm}
    \vspace{-2em}
\end{figure}

\subsection{Temporal Safe Corridor Generation}\label{subsec:safe_corridor_generation}

The safe corridor generation module constructs a series of overlapping convex hulls, or polyhedra, that allow the agent to navigate safely through the environment.  
The inputs for this module consist of the path generated by the global planner, the static occupancy map, and the predicted trajectories of dynamic obstacles.  
A polyhedron is generated around each segment of the piecewise linear path by first inflating an ellipsoid aligned with the segment, followed by computing tangent planes at the contact points of the ellipsoid with obstacles.  
See~\cite{liu2017planning} for a detailed explanation of this method.   

To account for timestamped dynamic obstacles, we estimate travel times along the global path using the double integrator model~\cite{feher2017constrained}.
Note that the path found by DGP is adjusted by Algorithms~\ref{alg:dynamic_path_adjustment} and ~\ref{alg:static_path_adjustment}, and therefore we use the double integrator to re-estimate travel times along the global path.
We then update the occupancy of the map and generate a snapshot of the environment for each segment.
The safe corridor generation module uses this snapshot to construct a safe corridor, accouting for dynamic obstacles.

\subsection{Starting Point Selection}\label{subsec:starting_point_selection}

This section describes how DYNUS selects the starting point (denoted as $A$ in Fig.~\ref{fig:trajectory_planning_framework}).  
DYNUS continuously replans its trajectory as it moves and adjusts the starting point based on the computation time of the global planner.  
If the starting point is too close to the current position, the global planner may not have enough time to compute the trajectory before the agent reaches that point.
Conversely, if the starting point is too far, the agent will have to execute a trajectory based on outdated information, making it suboptimal.
DYNUS uses an exponential moving average (EMA) of the computation time to adjust the starting point.
The EMA is computed as:
\begin{equation}
    \delta t_{A} \gets \alpha_{\delta t_{A}} \cdot \delta t_{A} + (1 - \alpha_{\delta t_{A}}) \cdot \delta t_{A}^{\text{new}}
\end{equation}
where $\delta t_{A}$ is the estimated computation time for the global planner to compute the trajectory from the starting point A, $\delta t_{A}^{\text{new}}$ is the new computation time, and $\alpha_{\delta t_{A}}$ is the EMA coefficient.

%% file: paper/04_trajectory_optimization.tex
\section{Trajectory Optimization}\label{sec:trajectory_optimization}

This section describes the trajectory optimization process for both the position and yaw of the agent.  
For position trajectory optimization, we utilize hard constraint with Mixed-Integer Quadratic Programming (MIQP).  
Although a MIQP-based hard constraint approach introduces additional variables (binary variables for interval-polyhedron assignment) and increases computational cost, it guarantees safety against static obstacles and improves the likelihood of finding a feasible solution.  
To reduce computational complexity, we introduce a variable elimination technique that pre-computes the dependencies of variables and eliminates dependent variables from the optimization problem.
Additionally, to balance efficiency and feasibility, we change the number of intervals based on the replanning results. 
A lower number of intervals mean fewer variables and constraints, which reduces computational cost but may lead to infeasible solutions. 
In contrast, more intervals increase the likelihood of finding feasible solutions but also increase computational cost.
We therefore starts with smaller number of intervals and increase the number of intervals if the solution is infeasible.

\subsection{Position Trajectory Optimization}\label{subsec:trajectory_optimization_for_position}

This section first describes DYNUS's MIQP formulation for position trajectory optimization.
We use triple integrator dynamics with the state vector: \(\mathbf{x}^T = \left[\vect{x}^T ~\vect{v}^T ~\vect{a}^T\right]\), where $\vect{x}$, $\vect{v}$, and $\vect{a}$ represent the position, velocity, and acceleration, respectively.

We formulate trajectory optimization using an $N$-interval composite B\'ezier curve with $P$ polyhedra.  
Let $n\in\{0:N-1\}$ denote a specific interval of the trajectory, $p\in\{0{:}P-1\}$ represent a specific polyhedron, and $dt$ denote the time allocated for each interval{\textemdash}the same for all intervals.  
To clarify the notation, $\vect{j}_n(\tau)$ denotes the jerk vector at the $n$-th interval at time $\tau$ within that interval.  
The control input, jerk, remains constant within each interval, allowing the position trajectory of each interval to be represented as a cubic polynomial:
\begin{equation}\label{eq:cubic_spline}
    \boldsymbol{x}_{n}(\tau)=\mathbf{a}_{n}\tau^{3}+\mathbf{b}_{n}\tau^{2}+\mathbf{c}_{n}\tau+\mathbf{d}_{n},\; \; \tau\in[0,dt]
\end{equation}
where $\mathbf{a}_{n}$, $\mathbf{b}_{n}$, $\mathbf{c}_{n}$, and $\mathbf{d}_{n}$ are the coefficients of the cubic spline in interval $n$.

We now discuss constraints for the optimization formulation.
Continuity constraints are added between adjacent intervals to ensure the trajectory is continuous:
\begin{equation}\label{eq:continuity_constraints}
    \mathbf{x}_{n+1}(0)=\mathbf{x}_{n}(dt) \quad \text{for} \quad n\in\{0{:}N-2\} \\
\end{equation}
The B\'ezier curve control points $\bb{p}_{nj}$ $(j \in \{0{:}3\})$ associated with each interval $n$ are:
\begin{align}
	&\boldsymbol{p}_{n0}=\mathbf{d}_{n}, 
	\quad \boldsymbol{p}_{n1}=\frac{\mathbf{c}_{n}dt+3\mathbf{d}_{n}}{3} \nonumber\\
	&\boldsymbol{p}_{n2}=\frac{\mathbf{b}_{n}dt^{2}+2\mathbf{c}_{n}dt+3\mathbf{d}_{n}}{3} \\
	&\boldsymbol{p}_{n3}=\mathbf{a}_{n}dt^{3}+\mathbf{b}_{n}dt^{2}+\mathbf{c}_{n}dt+\mathbf{d}_{n} \nonumber
\end{align}
and, to assign intervals to polyhedra and ensure the control points for each interval are within the corresponding polyhedron, we introduce binary variables $b_{np}$, where $b_{np}=1$ if interval $n$ is assigned to polyhedron $p$, and $b_{np}=0$ otherwise.
This condition is enforced through the following constraint:
\begin{align}
    & b_{np}=1\implies \mathbf{A}_{p}\boldsymbol{p}_{nj}\leq\mathbf{l}_{p}, \, \text{for} \quad j\in\{0{:}3\}, \, \forall n, \forall p
    \label{eq:polyhedron_constraints_MIQP}
\end{align} 
where polyhedra are denoted as $\{(\mathbf{A}_p, \mathbf{l}_p)\},\; p \in \{0{:}P-1\}$.
Each interval must be assigned to at least one polyhedron, which is ensured by the constraint:
\begin{align}
	& \sum_{p=0}^{P-1}b_{np}\ge1, \quad \forall n
    \label{eq:polyhedron_assignment_constraints}
\end{align}
To ensure the trajectory starts at an initial state and ends at a final state, we impose the following constraints:
\begin{align} \label{eq:initial_and_final_state}
    \mathbf{x}_{0}(0) &= \mathbf{x}_{\text{init}}, \quad \quad \mathbf{x}_{N}(dt) = \mathbf{x}_{\text{final}}
\end{align}
where $\mathbf{x}_{\text{init}}$ is the initial state, and $\mathbf{x}_{\text{final}}$ is the final state.
Note that the final state is chosen to be the mean of the last polyhedron's vertices.

For dynamic constraints, we require the velocity, acceleration, and jerk control points to satisfy the following constraints for all intervals $n$.
We define the control points for the $n$-th interval as:
\begin{align*}
    &\text{Velocity:} && \boldsymbol{v}_{nj}, \quad j \in \{0{:}2\} \\
    &\text{Acceleration:} && \boldsymbol{a}_{nj}, \quad j \in \{0{:}1\} \\
    &\text{Jerk:} && \boldsymbol{j}_{n}
\end{align*}
These control points are constrained by:
\begin{align}
	\left\Vert \boldsymbol{v}_{nj} \right\Vert_{\infty} &\le v_{\text{max}}, \quad j \in \{0{:}2\} \nonumber\\
	\left\Vert \boldsymbol{a}_{nj} \right\Vert_{\infty} &\le a_{\text{max}}, \quad j \in \{0{:}1\} \label{eq:velocity_accel_jerk_constraints} \\
	\left\Vert \boldsymbol{j}_{n} \right\Vert_{\infty} &\le j_{\text{max}} \nonumber 
\end{align}
where $v_{\text{max}}$, $a_{\text{max}}$, and $j_{\text{max}}$ denote the maximum allowable velocity, acceleration, and jerk, respectively.

The objective function consists of two components: (1) control input cost and (2) reference tracking cost. 
These are combined with respective weights to form the final cost:
\begin{equation}
    J = w_{\text{ctrl}} J_{\text{ctrl}} + w_{\text{ref}} J_{\text{ref}}
\end{equation}
\textbf{Control Input Cost:} We penalize the squared jerk along the trajectory for smooth motion:
\begin{equation}
    J_{\text{ctrl}} = \sum_{n=0}^{N-1} \left\| \mathbf{j}_n \right\|^2
\end{equation}
\textbf{Reference Tracking Cost:} We penalize deviation from reference points $\mathbf{x}_{\text{ref},i}$, which are computed as the mean of each polyhedron's vertices. 
These reference points are time-aligned along the trajectory using total trajectory time $T = N \cdot dt$:
\begin{equation}
    J_{\text{ref}} = \sum_{i=1}^{P-2} \left\| \mathbf{x}(t_i) - \mathbf{x}_{\text{ref},p} \right\|^2
\end{equation}
where $t_i = \frac{i}{P - 1} T$ is the time associated with reference point $\mathbf{x}_{\text{ref},i}$, and $\mathbf{x}(t_i)$ is the position evaluated at time $t_i$.
The reason for excluding the first and last polyhedra is that they are already constrained by Eq.~\ref{eq:initial_and_final_state}, and there is no need to have reference points for them.

The complete MIQP problem is then formulated as:
\begin{equation}\label{eq:MIQP_formulation}
\begin{split}
    \min_{\mathbf{a}_n, \mathbf{b}_n, \mathbf{c}_n, \mathbf{d}_n, b_{np}} \quad & J \\
    \text{s.t.} \quad \quad \quad \ & \text{Eqs.~\eqref{eq:continuity_constraints}, \eqref{eq:polyhedron_constraints_MIQP}, \eqref{eq:polyhedron_assignment_constraints}, \eqref{eq:initial_and_final_state}, and \eqref{eq:velocity_accel_jerk_constraints}}
\end{split}
\end{equation}

\subsubsection{Variable Elimination}\label{subsubsec:variable_elimination}

In our MIQP formulation (see Eq.~\eqref{eq:MIQP_formulation}), four sets of coefficients,
\(\mathbf{a}_n\), \(\mathbf{b}_n\), \(\mathbf{c}_n\), and \(\mathbf{d}_n\), are introduced per segment \(n\). 
Since the total number of decision variables grows linearly with the number of segments, the computational burden increases accordingly.

To address this challenge, we employ a variable elimination technique that leverages the structure of cubic splines. 
Specifically, by symbolically solving the equality constraints imposed by the initial/final conditions and the continuity conditions between segments, we can express most of the spline coefficients in terms of a small set of free variables. 
This reparameterization has two main benefits:
\begin{enumerate}
    \item It significantly reduces the number of decision variables.
    \item It eliminates the need to include equality constraints explicitly in the optimization.
\end{enumerate}

\textbf{Problem Setup:} Each segment is a cubic polynomial, as shown in Eq.~\ref{eq:cubic_spline}, and the overall optimization is subject to two sets of equality constraints:
\begin{itemize}
    \item \textbf{Continuity Constraints (Eq.~\eqref{eq:continuity_constraints}):} These ensure continuity of position, velocity, and acceleration at the junctions between consecutive segments, resulting in \(3(N-1)\) constraints per axis.
    \item \textbf{Boundary Conditions (Eq.~\eqref{eq:initial_and_final_state}):} These impose the initial and final values for position, velocity, and acceleration (a total of 6 constraints per axis).
\end{itemize}
Thus, for \(N\) segments, there are \(4N\) decision variables and \(3N+3\) equality constraints per axis. 

\textbf{Case Study 1 (\(N = 3\)):}
\begin{itemize}
    \item \emph{Decision Variables:} \(4N = 12\) per axis (i.e., 36 total for \(x\), \(y\), and \(z\)).
    \item \emph{Equality Constraints:} \(3N+3 = 12\) per axis (i.e., 36 total).
\end{itemize}
This implies that the system is fully determined, resulting in a unique closed-form solution for all spline coefficients as functions of the boundary conditions. 
While this formulation enables very fast computation, it does not guarantee the satisfaction of inequality constraints (e.g., Eqs.~\eqref{eq:velocity_accel_jerk_constraints} and \eqref{eq:polyhedron_constraints_MIQP}). 
These constraints can be verified in a post-optimization step, making the \(N = 3\) formulation appropriate for safe and contingency trajectory generation (see Section~\ref{subsec:trajectory_planning_framework}).

\textbf{Case Study 2 (\(N = 4\)):}
\begin{itemize}
    \item \emph{Decision Variables:} \(4N = 16\) per axis (i.e., 48 total for all axes).
    \item \emph{Equality Constraints:} \(3N+3 = 15\) per axis (i.e., 45 total).
\end{itemize}
which indicates the existence of one free parameter per axis. 
Symbolic elimination reveals that this free variable is \(\mathbf{d}_3\) (the first control point of the final segment, where \(n \in \{0,1,2,3\}\)). 
Consequently, all other spline coefficients can be expressed as explicit affine functions of \(\mathbf{d}_3\). This reformulation provides the following advantages:
\begin{itemize}
    \item The equality constraints are implicitly satisfied, enabling their removal from the optimization problem.
    \item Only a single decision variable per axis, \(\mathbf{d}_3\), remains explicitly.
    \item All other control points are formulated as affine functions of \(\mathbf{d}_3\).
\end{itemize}
This leads to a revised MIQP formulation featuring significantly fewer decision variables and no equality constraints:
\begin{equation}\label{eq:MIQP_formulation_variable_elimination}
\begin{aligned}
    \min_{\mathbf{d}_3,\, b_{np}} \quad & J \\
    \text{s.t.} \quad \,\, & \text{Eqs.~\eqref{eq:polyhedron_constraints_MIQP}, \eqref{eq:polyhedron_assignment_constraints}, and \eqref{eq:velocity_accel_jerk_constraints}}.
\end{aligned}
\end{equation}

\subsubsection{Contingency Trajectory Generation}\label{subsubsec:contingency_trajectory}

This section discusses how we generate the contingency trajectory (See Section~\ref{subsec:trajectory_planning_framework} for details), using the closed-form solution ($N=3$).
Given the current state of the agent and the predicted future collision point $\mathbf{x}_{\text{col}}$ at time $t_{\text{col}}$, we first compute a velocity-aware safety distance:
\[
d_{\text{safe}} = d_{\min} + (d_{\max} - d_{\min}) \cdot \frac{\|\mathbf{v}_{\text{curr}}\|}{v_{\max}}
\]
where $d_{\min}, d_{\max}$ are pre-defined distance bounds, and $\mathbf{v}_{\text{curr}}$ is the current velocity of the agent. 
We then define a contingency goal $\mathbf{x}_{\text{goal}}$ in the direction of motion: \(\mathbf{x}_{\text{goal}} = \mathbf{x}_{\text{curr}} + d_{\text{safe}} \cdot \hat{\mathbf{v}}_{\text{curr}}\), where $\mathbf{x}_{\text{curr}}$ is the current position of the agent, and $\hat{\mathbf{v}}_{\text{curr}}$ is the unit vector in the direction of current velocity. 
To explore alternative directions, we build a plane orthogonal to $\hat{\mathbf{v}}_{\text{curr}}$ and define evenly spaced lateral directions (at $45^\circ$ increments), generating eight additional candidate goals around $\mathbf{x}_{\text{center}}$.
Each candidate's goal is scored based on its distance from the predicted collision point, and the planner iteratively attempts to connect to the farthest candidate using the closed-form solution. 
As soon as a feasible plan is found, it is committed and replaces the previous trajectory.
If none of the contingency goals result in a feasible trajectory, the agent executes an emergency stop. 
In this case, the planner replaces the committed trajectory with a static hover command at the current position and zero velocity, acceleration, and jerk.

\subsubsection{Time Allocation and Parallelization}\label{subsec:time_allocation}

\begin{figure}
    \centering
    \includegraphics[width=\columnwidth]{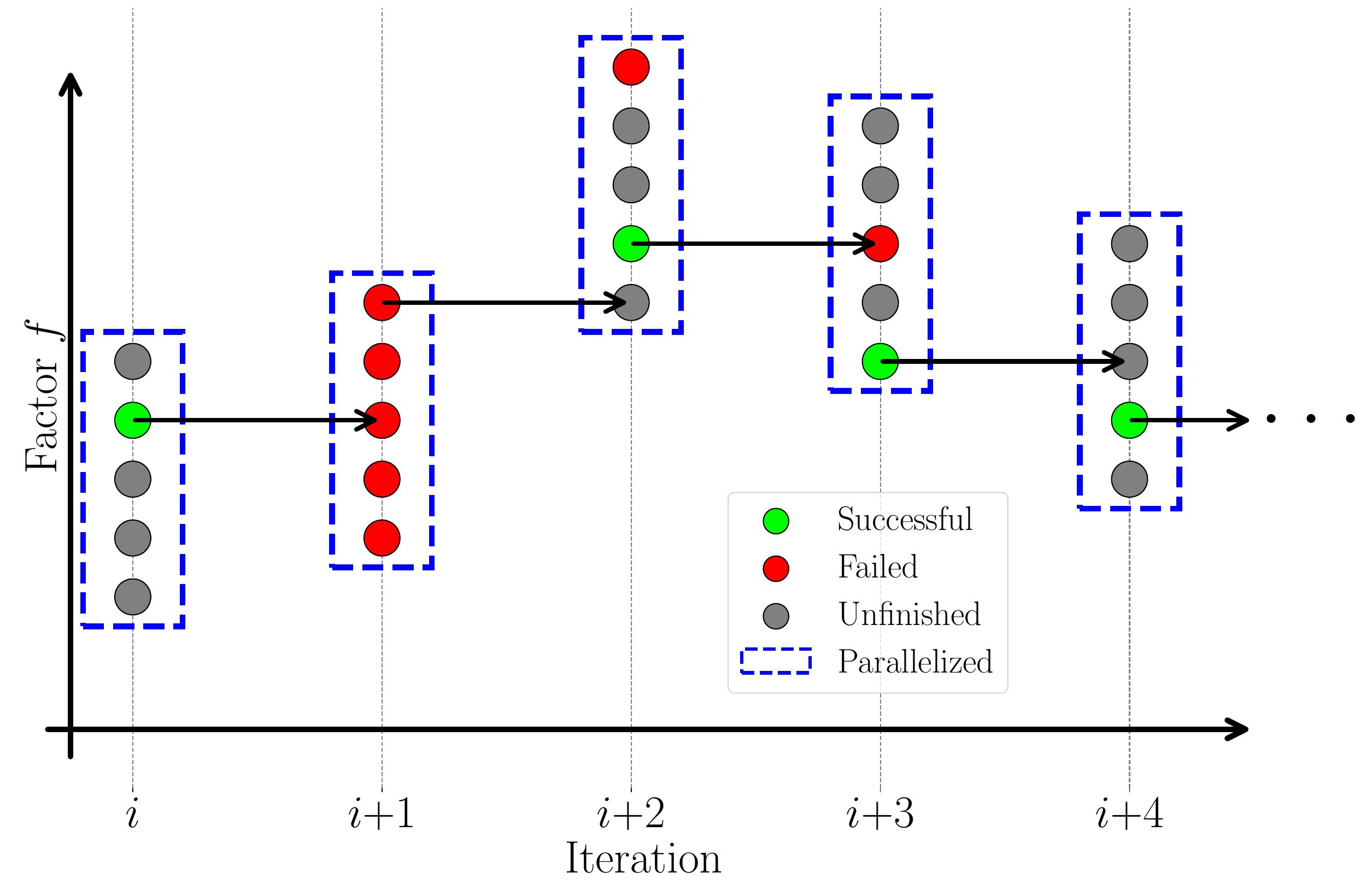}
    \caption{Time allocation and parallelization: At iteration $i$, the second largest factor is the fastest successful factor; therefore, at iteration $i+1$, the factors are set such that this second largest factor is the median of the factors. At iteration $i+1$, all the factors failed, so we initialize the factor at iteration $i+2$ with the largest factor of iteration $i+1$ to be the smallest factor at iteration $i+2$. Note that each iteration has a different set of conditions (e.g., obstacles, initial/final states); therefore, we try the largest factor from iteration $i+1$ again at $i+2$. At iteration $i+2$, the second smallest factor is the fastest successful factor; therefore, at iteration $i+3$, the factors are set such that this factor is the median of the factors. This process continues until the agent reaches the goal.}
    \label{fig:time_allocation_and_parallelization}
    \vspace{-2em}
\end{figure}

To find the optimal time allocation, we compute the infinity norm of the difference between the initial and final positions and divide it by the maximum velocity:
\begin{equation}
    T_{\text{total}} = \frac{\left\| \mathbf{x}_{\text{final}} - \mathbf{x}_{\text{init}} \right\|_{\infty}}{v_{\text{max}}}
\end{equation}
and we solve the variable-eliminated optimization problem in parallel with different factors $f$ to find the optimal time allocation:
\begin{equation}
    dt = f \cdot \frac{T_{\text{total}}}{N}
\end{equation}
where $f$ is a factor that we vary in parallel to efficiently explore different time allocations.
Fig.~\ref{fig:time_allocation_and_parallelization} shows how we initialize the time allocation with different factors and solve the optimization problems in parallel to find the optimal trajectory.
As soon as any of the optimization problems running in parallel find a feasible solution, the other optimization problems are stopped, and that solution is used.  The factors for the next iteration are chosen so that the previously successful factor is the median of the new ones. 
If none of the parallelized optimization problems find a feasible solution, we initialize the minimum factor to be the largest factor in the previous iteration.

\subsubsection{Adaptive Number of Intervals}\label{subsubsec:adaptive_number_of_intervals}

QP-based methods~\cite{liu2017planning} use an equal number of intervals and polyhedra ($P = N$) while using non-uniform time allocation.
This approach is computationally efficient but may lead to infeasible solutions.
In contrast, MIQP-based methods~\cite{tordesillas2022faster,toumieh2024high} use $P < N$ with uniform time allocation.
This approach improves feasibility, but it also increases computational cost.
DYNUS adaptively changes the number of intervals based on the replanning results{\textemdash}when the solution is infeasible, the number of intervals is increased.

\subsection{Map Representation}\label{subsec:map_representation}

\begin{figure}
    \centering
    \begin{tikzpicture}[scale=1]

        \definecolor{unknown}{HTML}{98dfe8} %
        \definecolor{occupied}{HTML}{ff9d2a} %
        \definecolor{free}{HTML}{89d086} %
        \definecolor{goal}{HTML}{ffe747} %
        \definecolor{sliding_window}{HTML}{70b491} %

        \node[anchor=south west,inner sep=0] (image) at (0,0) {
            \includegraphics[width=\columnwidth, trim=1 1 1 1, clip]{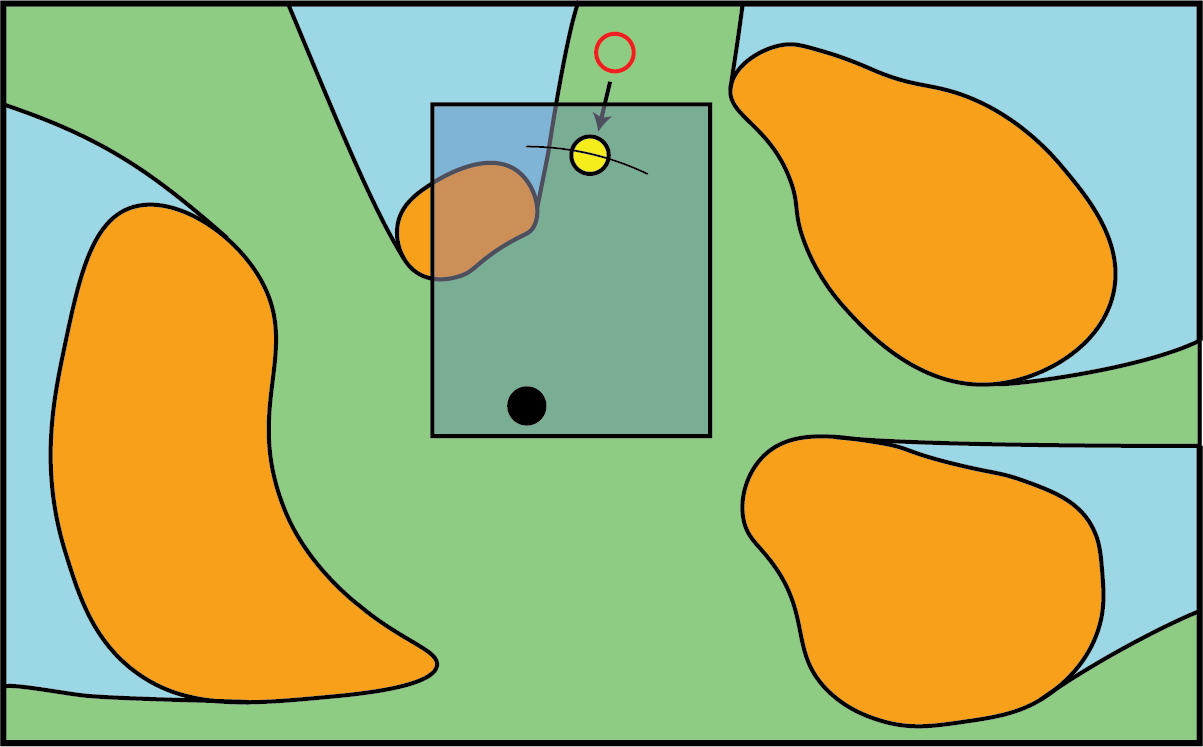}
        };
        
        \node at (8.0,4.7) {\Large $\mathcal{U}$}; %
        \node at (1.2,2.0) {\Large $\mathcal{O}$}; %
        \node at (6.8,3.5) {\Large $\mathcal{O}$}; %
        \node at (6.8,1.0) {\Large $\mathcal{O}$}; %
        \node at (4.5,0.5) {\Large $\mathcal{F}$}; %

        \draw[black] (0.12,-0.1) rectangle (0.62,-0.6);
        \fill[unknown] (0.12,-0.1) rectangle (0.62,-0.6);
        \node[anchor=center] at (0.37,-0.36) {$\mathcal{U}$};
        \node[right] at (0.62,-0.36) {Unknown};
        
        \draw[black] (2.5,-0.1) rectangle (3.0,-0.6);
        \fill[occupied] (2.5,-0.1) rectangle (3.0,-0.6);
        \node[anchor=center] at (2.75,-0.36) {$\mathcal{O}$};
        \node[right] at (3.0,-0.38) {Occupied};

        \draw[black] (4.8,-0.1) rectangle (5.3,-0.6);
        \fill[free] (4.8,-0.1) rectangle (5.3,-0.6);
        \node[anchor=center] at (5.05,-0.36) {$\mathcal{F}$};
        \node[right] at (5.3,-0.38) {Known-free};

        \draw[black, very thick] (0.12,-0.7) rectangle (0.62,-1.2);
        \fill[sliding_window]  (0.12,-0.7) rectangle (0.62,-1.2);
        \node[right] at (0.62,-0.97) {Sliding window};

        \draw[line width=1.5pt, black] (3.3,-0.95) circle (0.1);
        \fill[black] (3.3,-0.95) circle (0.1); 
        \node[right] at(3.4,-0.97) {Agent's location};

        \draw[line width=1.5pt, black] (0.37,-1.55) circle (0.1);
        \fill[goal] (0.37,-1.55) circle (0.1);
        \node[right] at(0.62,-1.55) {Projected Sub-goal};

        \draw[line width=1pt, red] (3.8,-1.55) circle (0.1);
        \node[right] at(3.9,-1.55) {Terminal goal};

    \end{tikzpicture}
    \caption{DYNUS map representation: the global map is stored as an Octomap, while the sliding window maintains a local map around the agent's current position. The terminal goal is projected onto the agent's horizon, and the size of the sliding window is determined based on the agent's current position, previous path, and the projected goal.}
    \label{fig:map}
    \vspace{-2em}
\end{figure}

To achieve efficient map storage, we use an octomap~\cite{hornung2013octomap}.  
Additionally, as illustrated in Fig.~\ref{fig:map}, we use a sliding window with varying size.
We first project the terminal goal onto DYNUS's planning horizon and adjust the sliding window's size based on the projected goal's location.  
Specifically, we increase the window's size in the direction of the projected goal.
The octomap stores the entire map (global map), while the sliding window maintains a local map for planning.  
Note that, as discussed in Section~\ref{sec:frontier}, when DYNUS is assigned exploration tasks, we do not perform projection and use the best frontier as the sub-goal.

We also implemented the removal of residual obstacle traces or ``smears'' in the octomap.
When sensors detect dynamic obstacles, residual data may persist even after the obstacles have moved away, creating false obstacles on the map.
To address this, we introduce a smear-removal mechanism that periodically checks the last detection timestamp of obstacles and removes occupied space if the obstacle has not been detected for a certain period.

\subsection{Yaw Optimization}\label{subsec:yaw_optimization}

As discussed in Section~\ref{subsec:lit_review_yaw_optimization}, coupled approaches are computationally expensive and may not be suitable for real-time applications.
Thus, DYNUS adopts a decoupled approach, where the position and yaw trajectories are optimized separately.
First, we run a graph search to find the optimal sequence of discrete yaw angles along the position trajectory.
Next, we perform B-spline fitting to smooth the discrete yaw angles, ensuring a continuous and feasible yaw trajectory.

\subsubsection{Graph Search for Discrete Yaw Angles}\label{subsubsec:graph_search_for_discrete_yaw_angles}

To find a sequence of yaw angles that optimally balance (1) tracking obstacles and (2) looking in the direction of motion, we employ a graph search algorithm that balances multiple objectives: (1) collision likelihood, (2) velocity of dynamic obstacles, (3) proximity to obstacles, (4) time since last observed, and (5) minimization of yaw changes. 
The algorithm operates over a discretized time horizon and incrementally explores potential yaw states by evaluating their utility.

The graph search begins by initializing an open set containing the root node, which corresponds to the initial position, yaw, and time. 
At each iteration, the algorithm selects the node with the lowest cost from the open set and expands it by generating potential next yaw angeles. 
For each new state, the utility is computed as a weighted sum of the following components.

\noindent
\textbf{Collision Likelihood}:
Using the Mahalanobis distance, the collision likelihood evaluates the probability of a collision between the agent and obstacles by considering position uncertainty. 
This uncertainty is represented by the covariance matrix obtained from the Adaptive Extended Kalman Filter (AEKF) obstacle tracker, which is discussed in detail in Section~\ref{sec:obstacle_tracking}. 
The collision likelihood is defined as:
\begin{align} 
    U_{\text{collision}} &= \frac{1}{K} \sum_{i=0}^{K-1} \exp\left(-\frac{1}{2} d_M(t_i)^2\right) \\
    d_M(t_i) &= \| \vect{x}_a(t_i) - \vect{x}_o(t_i) \|_{\Sigma(t_i)^{-1} }
\end{align}
where $\vect{x}_a(t)$ and $\vect{x}_o(t)$ represent the positions of the agent and the obstacle at time $t$, respectively, $K$ is the number of sampled time points, and $\Sigma(t)$ is the blended covariance at time $t$, defined as:
\begin{equation} 
    \Sigma(t) = \left(1 - \frac{t - t_{\text{cur}}}{T_{\text{total}}}\right) \Sigma_{\text{EKF}} + \frac{t - t_{\text{cur}}}{T_{\text{total}}} \Sigma_{\text{poly}} 
\end{equation}
where $T_{\text{total}} = t_{\text{end}} - t_{\text{cur}}$ is the total duration of the trajectory, with $t_{\text{end}}$ representing the final time and $t_{\text{cur}}$ the current time, and $\Sigma_{\text{EKF}}$ is the AEKF estimate covariance, and $\Sigma_{\text{poly}}$ has the dynamic obstacles' predicted future trajectory (polynomial)'s fitting residuals as the diagonal elements.
We blended the covariances since, at $t = t_{\text{cur}}$, the estimation uncertainty primarily arises from the AEKF estimate covariance, while as time progresses, the uncertainty from the trajectory prediction becomes more significant.

\noindent
\textbf{Velocity}:
The velocity cost encourages an agent to track dynamic obstacles that are moving faster. 
\begin{align} 
    U_{\text{velocity}} &= \frac{1}{M} \sum_{i=0}^{M-1} |\vect{v}_o(t_i)|
\end{align}
where $\vect{v}_o(t)$ is the velocity of the obstacle at time $t$, and $M$ is the number of sampled time points.

\noindent
\textbf{Time Since Observed}:
To prioritize obstacles (and the direction of motion) that have not been observed recently, the time since the last observation is calculated for each obstacle:
\begin{align} 
    U_{\text{observed}} &= \sum_{o \in \mathcal{O}} t_{\text{slo}, o}
\end{align}
where $\mathcal{O}$ represents the set of obstacles within the cutoff distance, and $t_{\text{slo}, o}$ denotes the time elapsed since the last observation of obstacle $o$.
For clarity, a smaller value of this term for a particular yaw angle indicates that all obstacles, as well as the direction of motion, have been observed recently, indicating a good yaw angle.  
Note that in DYNUS, the direction of motion is computed using a point that is $t_{lookup}$ seconds ahead along the planned trajectory.

\noindent
\textbf{Proximity}:
The proximity score prioritizes nearby obstacles by summing the reciprocal of the distances between the agent and obstacles within the cutoff distance:
\begin{align} 
    U_{\text{proximity}} &= \sum_{o \in \mathcal{O}} \frac{1}{\|\vect{x}_a(t_{\text{cur}}) - \vect{x}_o(t_{\text{cur}})\|}
\end{align}

\noindent
\textbf{Yaw Change}:
A penalty is introduced for differences between consecutive yaw angles to enforce smooth transitions:
\begin{align}
    U_{\text{yaw}} &= \left(\psi(t_{i+1}) - \psi(t_i)\right)^2
\end{align}
where $\psi(t_i)$ and $\psi(t_{i+1})$ denote the yaw angles at times $t_i$ and $t_{i+1}$, respectively.

\noindent
\textbf{Overall Utility Function}:
The total utility function for a node is defined as:
\begin{align}
    U &= \sum_{i \in \mathcal{U}} w_i \cdot U_i
\end{align}
where $\mathcal{U} = \{\text{collision}, \text{velocity}, \text{observed}, \text{proximity}, \text{yaw}\}$ represents the set of utility components.
At the terminal node, an additional penalty is applied to minimize the difference between the final yaw and a reference terminal yaw:
\begin{equation} 
    U_{\text{final}} = w_{\text{final}} \cdot (\psi_{\text{final}} - \psi_{\text{terminal}})^2 
\end{equation}
To reconstruct the optimal yaw sequence, the algorithm tracks the parent node for each explored state, enabling backtracking once the terminal node is reached.
The resulting yaw sequence is then smoothed using a B-spline fitting process, which is detailed in the following.

\subsubsection{Yaw B-Spline Fitting}\label{subsubsec:yaw_fitting}

To achieve a smooth yaw trajectory, we employ a clamped uniform cubic B-spline fitting.
This approach minimizes the squared error between the optimized discrete yaw sequence and the fitted B-spline trajectory while enforcing constraints on the yaw rate.

\noindent
\textbf{Problem Formulation}:
The yaw trajectory is represented as a cubic B-spline defined by a set of control points $\mathbf{q} = \{q_0, q_1, \dots, q_{n}\}$. 
The objective is to determine the optimal control points that minimize the squared error between the B-spline values and the yaw sequence $\{\psi_{\text{opt},i}\}$ for $i \in \{0{:}S-1\}$, where $S$ is the number of yaw angles in the sequence.

\noindent
\textbf{Objective Function}:
The optimization minimizes the squared error between the B-spline values $\psi_{\text{B-spline}}(t_i)$ and the discrete yaw values $\psi_{\text{opt},i}$: 
\begin{equation} 
    \min_{\mathbf{q}} \sum_{i=0}^{S-1} \left(\psi_{\text{opt},i} - \psi_{\text{B-spline}}(t_i)\right)^2
\end{equation} 
where $\psi_{\text{B-spline}}(t)$ represents the B-spline value at time $t$, and the time increment between two consecutive points, $t_{i+1} - t_i$, is given by $T_{\text{total}}/S$, with $T_{\text{total}}$ denoting the total trajectory duration.
Since the B-spline is defined over a clamped uniform knot vector, the initial and final yaw values are inherently satisfied.

\noindent
\textbf{Constraints}: To ensure smoothness and enforce yaw rate constraints, the yaw rate control points are expressed as:
\begin{equation} 
    \dot{\psi}_i = \frac{p}{t_{i+p+1} - t_{i+1}} (q_{i+1} - q_{i})
\end{equation} 
where $\dot{\psi}(t)$ represents the yaw rate control point, $p$ is the degree of the B-spline (here $p=3$), and $t_i$ denotes the time at point $i$ in knots.
For all control points $q_i$, the yaw rate constraints are enforced as 
$    |\dot{\psi}_i| \leq \omega_{\text{max}}$, 
where $\omega_{\text{max}}$ is the maximum allowable yaw rate.
The optimization is solved using Gurobi~\cite{gurobi}.

%% file: paper/06_obst_tracking.tex
\section{Dynamic Obstacle Tracking and Prediction}\label{sec:obstacle_tracking}

\noindent
\textbf{Dynamic Obstacle Tracking}: 
Many methods assume predefined noise models for both process and measurement noise, and they typically use fixed noise covariances, which could represent inaccurate noise models.  
To overcome this challenge, we employ an Adaptive Extended Kalman Filter (AEKF)~\cite{akhlaghi2017adaptive}, which dynamically adjusts the process noise and measurement noise covariances.
By continuously updating these covariance values, the AEKF accounts for uncertainty in estimation.
Below, we briefly describe the covariance update steps of the AEKF algorithm for both process noise and sensor noise covariances.

The measurement noise covariance \( R_k \) is updated adaptively based on the residual, which is the difference between the actual and estimated measurement:
\begin{equation}
    \begin{aligned}
        \epsilon_k &= z_k - h(\hat{x}_k^+, u_k) \\
        R_k &= \alpha R_{k-1} + (1 - \alpha) \epsilon_k \epsilon_k^T
    \end{aligned}
\end{equation}
where \( \epsilon_k \) is the residual, \( z_k \) is the actual measurement, \( \hat{x}_k^+ \) is the updated state estimate, \( h(\cdot) \) is the measurement model, \( u_k \) is the control input, which is not used in our case, and \( \alpha \) is a forgetting factor (with \( 0 < \alpha \leq 1 \)) that controls the influence of past values on the new covariance estimate.

The process noise covariance \( Q_k \) is updated adaptively using the innovation:
\begin{equation}
    \begin{aligned}
        d_k &= z_k - h(\hat{x}_k^-, u_k) \\
        Q_k &= \alpha Q_{k-1} + (1 - \alpha) K_k d_k d_k^T K_k^T
    \end{aligned}
\end{equation}
where \( d_k \) is the innovation, \( \hat{x}_k^- \) is the predicted state, and \( K_k \) is the Kalman gain.
We implement this AEKF-based filtering approach to estimate and smooth the history of dynamic obstacles' positions, velocities, and accelerations.  
Additionally, when DYNUS detects a new dynamic obstacle, it initializes $Q_0$ and $R_0$ by averaging the process noise and sensor noise covariances of previously detected dynamic obstacles.  
This approach enables DYNUS to capture the uncertainty in both the prediction and sensor noise of the dynamic obstacles.

\noindent
\textbf{Dynamic Obstacle Prediction}: To predict the future positions of dynamic obstacles, we use a constant acceleration model.
After predicting the future positions, we fit it into a polynomial to smooth the prediction.  
Additionally, we compute residual values to capture prediction inaccuracies, which are used in the yaw optimization step detailed in Section~\ref{subsec:yaw_optimization}.

%% file: paper/07_exploration.tex
\section{Frontier-based exploration}\label{sec:frontier}
In some missions, target goals are predefined, while in others, the agent must explore the environment autonomously.  
To enable autonomous exploration, we designed a frontier-based exploration approach.

\subsection{Frontier-based Exploration Algorithm}~\label{subsubsec:frontier_selection}

DYNUS selects the optimal frontier based on a multi-objective cost function. 
The algorithm evaluates a set of candidate frontiers and selects the one with the lowest cost. 
The algorithm includes Frontier Filtering, Cost Function Evaluation, and Frontier Selection.

\subsubsection{Frontier Filtering}
Frontiers $\mathcal{F} = \{\vect{f}_1, \vect{f}_2, \dots, \vect{f}_n\}$ are identified by detecting voxels in the octomap that lie between known-free and unknown voxels and are within the agent's sliding window map.

\subsubsection{Cost Function Evaluation}
For each detected frontier $\vect{f}_i$, a cost function $C(\vect{f}_i)$ is computed as:
\begin{align}
    C(\vect{f}_i) &= \sum_{j \in \mathcal{C}} w_j \cdot C_j(\vect{f}_i)
\end{align}
where $\mathcal{C} = \{\text{vel}, \text{camera}, \text{continuity}, \text{forward}, \text{info}\}$, $w_j$ is the weight associated with cost $C_j$.

\noindent
\textbf{Velocity}: Inspired by the work of~\cite{cieslewski2017rapid}, the frontier that allows the agent to maintain the desired velocity is preferred:
\begin{align}
    C_\text{vel}(\vect{f}_i) &= \frac{1}{\|\vect{v}_\text{des}(\vect{f}_i)\| + \epsilon} \\
    \vect{v}_\text{des}(\vect{f}_i) &= \frac{\vect{f}_i - \vect{x}_\text{agent}}{d_\text{max}} \cdot v_\text{max}
\end{align}
where $\vect{x}_\text{agent}$ is the agent's current position, $d_\text{max}$ is the maximum distance to the frontier, $v_\text{max}$ is the maximum velocity, and $\epsilon$ is a small positive constant.

\noindent
\textbf{Camera}: To align the selected frontier with the forward direction of the camera, the cost is calculated as:
\begin{equation}
    C_\text{camera}(\vect{f}_i) = \sqrt{x_\text{camera}^2 + y_\text{camera}^2}
\end{equation}
where $(x_\text{camera}, y_\text{camera}, z_\text{camera})$ are the coordinates of $\vect{f}_i$ in the camera frame.

\noindent
\textbf{Continuity}: To maintain continuity with the previously selected best frontier, $\vect{f}_\text{best}$, the cost function is defined as:
\begin{equation}
    C_\text{continuity}(\vect{f}_i) = \|\vect{f}_i - \vect{f}_\text{best}\|
\end{equation}

\noindent
\textbf{Forward}: To ensure the frontier lies within the camera's field of view, the following cost function is used:
\begin{equation}
    C_\text{forward}(\vect{f}_i) = -z_\text{camera}
\end{equation}

\noindent
\textbf{Information}: To maximize information gain, the following cost function is considered:
\begin{equation}
    C_\text{info}(\vect{f}_i) =  - |\{ \vect{f}_j \in \mathcal{F} \mid \|\vect{f}_i - \vect{f}_j\| < d_\text{thresh} \}|
\end{equation}
where $|\cdot|$ denotes the cardinality of the set, and $d_\text{thresh}$ is a threshold distance. 
This term encourages selecting frontiers that are clustered together, which indicates more information gain.

\subsubsection{Frontier Selection}
The frontier that minimizes the total cost is selected as the best frontier:
\begin{equation}
    \vect{f}_\text{best} = \arg\min_{\vect{f}_i \in \mathcal{F}} C(\vect{f}_i)
\end{equation}

%% file: paper/10_simulation_results.tex
\section{Simulation Results}\label{sec:simulation-results}

We performed all the simulations on an \texttt{AlienWare Aurora R8} desktop computer with an Intel\textsuperscript{\textregistered} Core~\textsuperscript{TM} i9\-9900K CPU @ 3.60GHz$\times$16, 64 GB of RAM.
The operating system is Ubuntu 22.04 LTS, and we used ROS2 Humble.

To simulate LiDAR data, we used the \texttt{livox\_ros\_driver2} package, which provides a ROS 2 interface for the Livox MID-360 LiDAR sensor.
For depth camera data, we utilized the ROS 2 interface provided by the \texttt{realsense-ros} package for the Intel\textsuperscript{\textregistered} RealSense\textsuperscript{TM} D435 depth camera.

\subsection{Benchmarking against State-of-the-Art Methods in Static Environments}

We performed benchmarking experiments to compare DYNUS against state-of-the-art methods: FASTER~\cite{tordesillas2022faster}, SUPER~\cite{ren2025super}, and EGO-Swarm~\cite{zhou2021ego-swarm}.
FASTER and EGO-Swarm use a depth camera, while SUPER uses a LiDAR sensor.
As illustrated in Fig.~\ref{fig:system_overview}, DYNUS uses both the LiDAR sensor and the depth camera for mapping.
Since all of them assume a static environment, we evaluated them in a static forest setting, as shown in Fig.~\ref{fig:global_planner_benchmarking_gazebo}.
The simulation environment was generated with randomly generated static cylinder obstacles with a radius ranging from \SI{0.2}{\m} to \SI{1.0}{\m} and a height ranging from \SI{1.0}{\m} to \SI{5.0}{\m}.
The obstacles are spawned within \SI{100}{\m} $\times$ \SI{20}{\m}, and the agent starts at position $(0.0, 0.0, 3.0)$ \SI{}{\m}, and the goal position is $(105.0, 0.0, 3.0)$ \SI{}{\m}.
Each planner was executed in its preferred operating system: FASTER on Ubuntu 18.04 with ROS1 Melodic, and SUPER and EGO-Swarm on Ubuntu 20.04 with ROS1 Noetic. 
To ensure a fair comparison, all algorithms were containerized using Docker and executed on the same machine.
The reason why we used EGO-Swarm (multiagent planner) over EGO-Planner (single-agent planner) is that EGO-Planner's GitHub code specifically states that EGO-Swarm is more robust and safe than EGO-Planner.

\begin{figure}
  \centering
  \includegraphics[width=\columnwidth]{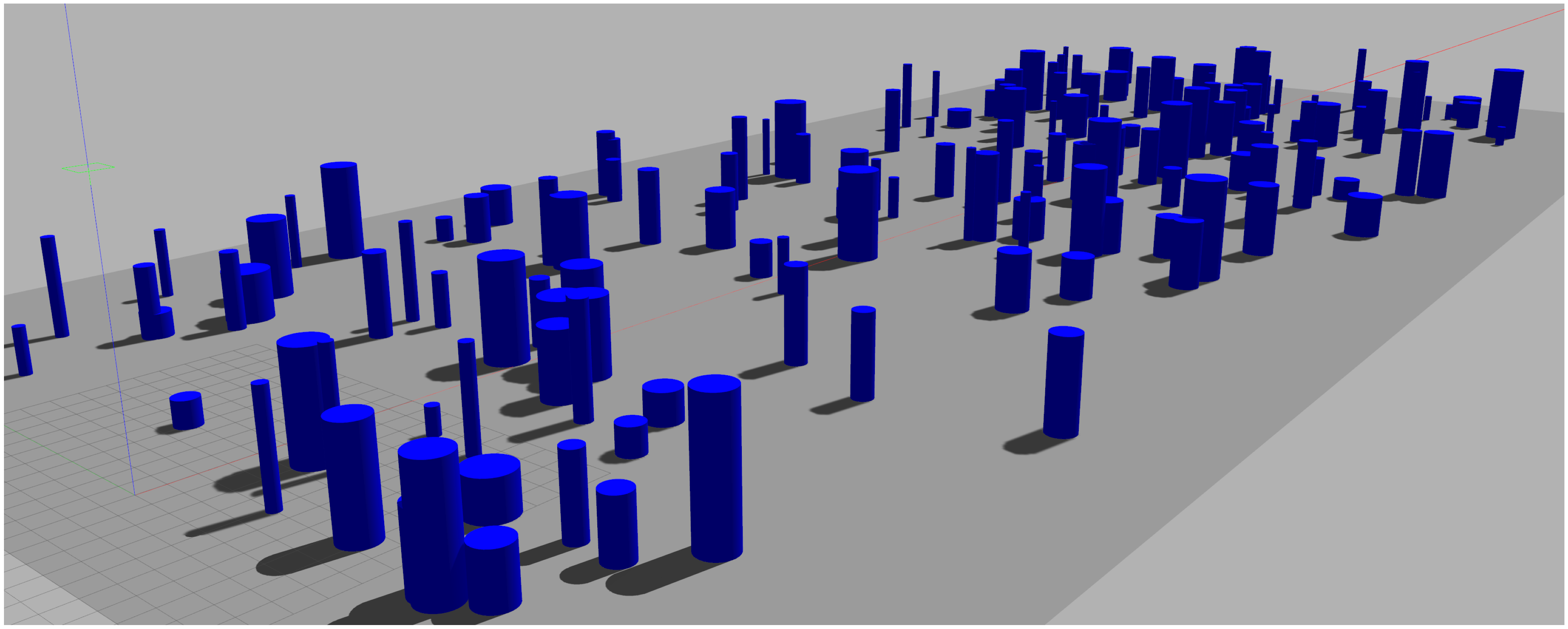}
  \caption{Static Forest: The static forest Gazebo environment.}
  \label{fig:sota_benchmarking_gazebo}
  \vspace{-1em}
\end{figure}

We performed 10 simulations with different dynamic constraints in a static forest environment, as shown in Fig.~\ref{fig:sota_benchmarking_gazebo}.
The dynamic constraints of Case 1 are set to $\vect{v_{\text{max}}} = 2.0$ \SI{}{\m/\s}, $\vect{a_{\text{max}}} = 5.0$ \SI{}{\m/\s\squared}, and $\vect{j_{\text{max}}} = 30.0$ \SI{}{\m/\s\cubed}, and the dynamic constraints of Case 2 are set to $\vect{v_{\text{max}}} = 10.0$ \SI{}{\m/\s}, $\vect{a_{\text{max}}} = 20.0$ \SI{}{\m/\s\squared}, and $\vect{j_{\text{max}}} = 30.0$ \SI{}{\m/\s\cubed}.
The evaluation metrics include the following:
\begin{itemize}
  \item \textbf{Success:} The number of successful runs without hitting obstacles and getting stuck.
  \item \textbf{Travel Time [s]:} Travel time taken by the agent to reach the goal position.
  \item \textbf{Path Length [m]:} Total length of the path traveled.
  \item \textbf{Global Path Planning Computation Time [ms]:} Average computation time for global path planning.
  \item \textbf{Exploratory Trajectory Optimization Computation Time [ms]:} Average computation time for exploratory trajectory optimization.
  \item \textbf{Safe Trajectory Optimization Computation Time [ms]:} Average computation time for safe trajectory optimization.
  \item \textbf{Total Planning Computation Time [ms]:} Total computation time for planning of the three components.
  \item \textbf{Replanning Total Computation Time [ms]:} Total computation time for replanning \textemdash for instance, DYNUS includes safe corridor generation, sub-goal computation, yaw planning, etc. 
\end{itemize}

\begin{figure}
  \centering
  \includegraphics[width=\columnwidth]{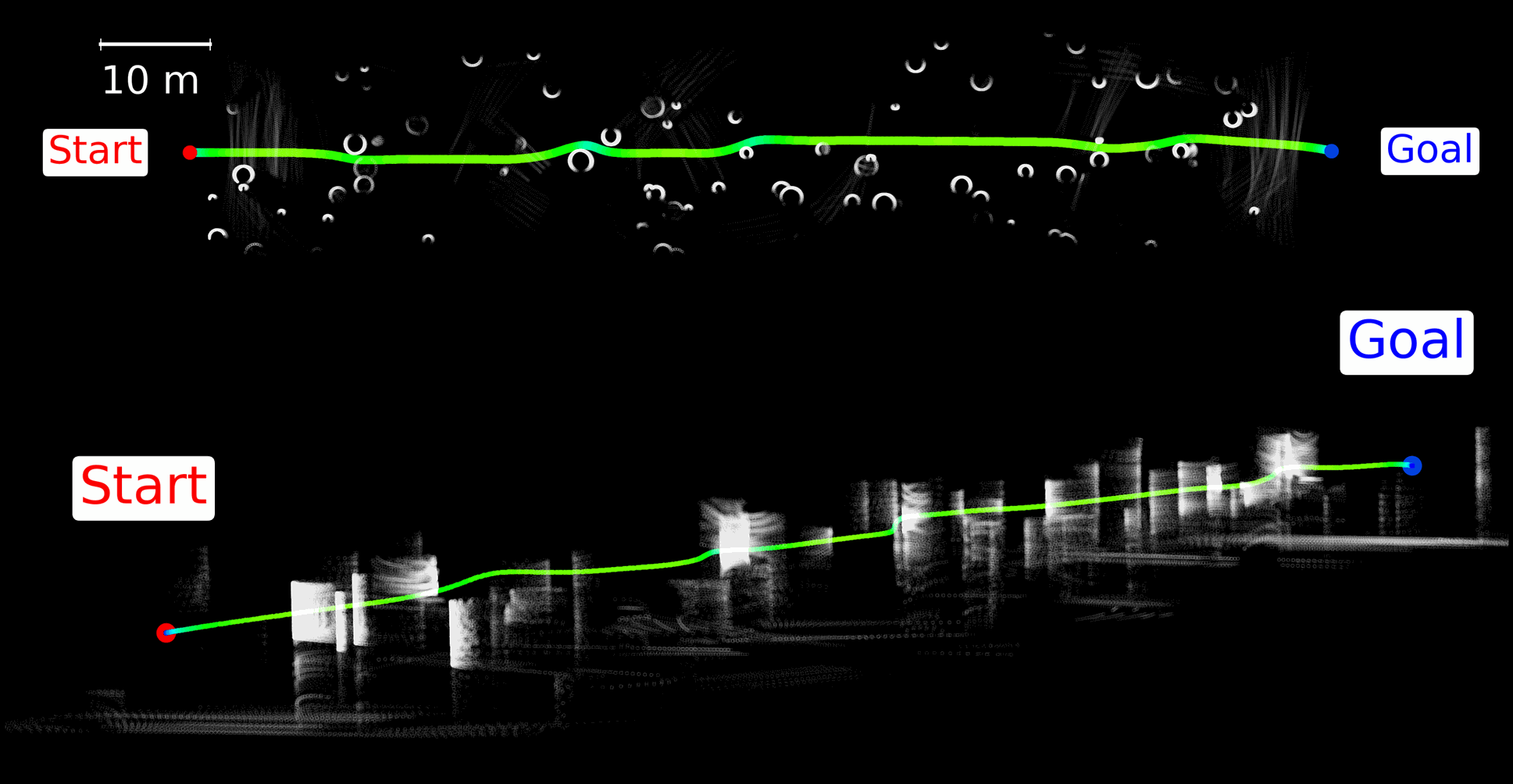}
  \caption{Static Forest: This figure shows DYNUS's trajectory color-coded according to the velocity profile.}
  \label{fig:sota_benchmarking}
  \vspace{-2em}
\end{figure}

Table~\ref{tab:local_trajectory_optimization_benchmarking} summarizes the benchmarking results, and Fig.~\ref{fig:sota_benchmarking} shows one of DYNUS's velocity profiles in Case 2.
Note that SUPER and EGO-Swarm are soft constraint methods, while FASTER and DYNUS are hard constraint methods.
The data reported in Table~\ref{tab:local_trajectory_optimization_benchmarking} is based on the average of 10 simulations, excluding failed runs.
Note that SUPER performs global path planning for both exploratory and safe trajectories, so we report both values as {Exploratory | Safe} under the Global Path Planning Computation Time column.

In Case 1, SUPER and FASTER suffer from a low success rate.
A common failure mode for SUPER is that the LiDAR does not provide sufficient point cloud coverage below the drone (the downward angle of Livox Mid-360 is -7.22 degrees), causing the drone to crash into the lower parts of obstacles.
EGO-Swarm and DYNUS achieved a 100\% success rate; however, DYNUS outperformed EGO-Swarm in terms of travel time and path length.
In terms of planning computation time, EGO-Swarm achieves the fastest computation time, followed by DYNUS.
Note that compared to FASTER, which also uses a hard constraint MIQP-based trajectory optimization method, DYNUS achieves a shorter total planning time due to its variable elimination technique.
Also note that FASTER and SUPER generate only a single safe trajectory, while DYNUS parallelizes safe trajectory optimization and generates up to 15 safe trajectories. (See Fig.~\ref{fig:trajectory_planning_framework} for details.)

In Case 2, FASTER and DYNUS achieved a 100\% success rate, while SUPER and EGO-Swarm suffered from a low success rate. 
This is consistent with the observation in HDSM~\cite{toumieh2024high} that the performance of EGO-Swarm degrades in high-speed scenarios. 
Further note that DYNUS achieved the fastest travel time and a shorter computation time compared to FASTER.

\begin{table*}
  \caption{Benchmark results against state-of-the-art methods in static environments. DYNUS outperforms the other methods in terms of travel time and achieves a 100\% success rate. Since SUPER performs global path planning for both exploratory and safe trajectories, we list the corresponding computation times as {Exploratory | Safe} in the Global Path Planning Computation Time column. Note that DYNUS generates 15 safe trajectories, and the reported computation time reflects the total time required to generate all of them.}
  \label{tab:local_trajectory_optimization_benchmarking}
  \begin{centering}
  \renewcommand{\arraystretch}{1.3}
  \resizebox{\textwidth}{!}{
  \begin{tabular}{>{\centering\arraybackslash}m{0.08\columnwidth} 
                  >{\centering\arraybackslash}m{0.25\columnwidth} 
                  >{\centering\arraybackslash}m{0.10\columnwidth} 
                  >{\centering\arraybackslash}m{0.1\columnwidth} 
                  >{\centering\arraybackslash}m{0.15\columnwidth} 
                  >{\centering\arraybackslash}m{0.16\columnwidth} 
                  >{\centering\arraybackslash}m{0.15\columnwidth} 
                  >{\centering\arraybackslash}m{0.15\columnwidth}
                  >{\centering\arraybackslash}m{0.15\columnwidth}
                  >{\centering\arraybackslash}m{0.15\columnwidth}
                  >{\centering\arraybackslash}m{0.17\columnwidth}}
  \toprule
  & & \multirow{3}{*}{\makecell{\textbf{Opt.} \\ \textbf{Type}}} & \multirow{3}{*}{\textbf{Success}} & \multicolumn{2}{c}{\textbf{Performance}} & \multicolumn{4}{c}{\textbf{Planning Comp. [ms]}} & \multirow{3}{*}{\makecell{\textbf{Replan Total} \\ \textbf{Comp. [ms]}}}\tabularnewline
  \cmidrule(lr){5-6} \cmidrule(lr){7-10}
  & & & & \textbf{Travel Time [s]} & \textbf{Path Length [m]} & \textbf{Global Path Plan} & \textbf{Exp. Traj. Opt.} & \textbf{Safe Traj. Opt} & \textbf{Total}  \tabularnewline
  \midrule
  \multirow{4}{*}{Case 1} 
    & \textbf{EGO-Swarm~\cite{zhou2021ego-swarm}}   & Soft & \best{10} / 10 & {80.5} & {110.0} & {-} & {-} & {-} & \best{0.69} & \best{0.69} \tabularnewline
\cline{2-11}
    & \textbf{SUPER~\cite{ren2025super}}            & Soft & \textcolor{red}{\textbf{3}} / 10 & {62.0} & {110.0} & {2.2 | 1.6} & {4.3} & {16.3} & {24.4} & {40.7} \tabularnewline
\cline{2-11}
    & \textbf{FASTER~\cite{tordesillas2022faster}}  & Hard & \textcolor{red}{\textbf{7}} / 10 & {52.5} & \best{104.9} & {6.0} & {15.8} & {9.6} & {31.4} & {33.2} \tabularnewline
\cline{2-11}
    & \textbf{DYNUS}                                & Hard & \best{10} / 10 & \best{46.2} & {105.8} & {3.1} & {5.9} & {3.8} & {12.8} & {37.4} \tabularnewline
  \midrule
  \multirow{4}{*}{Case 2} 
    & \textbf{EGO-Swarm~\cite{zhou2021ego-swarm}}   & Soft & \textcolor{red}{\textbf{1}} / 10 & {24.7} & {113.4} & {-} & {-} & {-} & \best{5.0} & \best{5.0} \tabularnewline
\cline{2-11}
    & \textbf{SUPER~\cite{ren2025super}}            & Soft & \textcolor{red}{\textbf{1}} / 10 & {20.6} & {109.9} & {2.1 | 1.3} & {1.3} & {11.5} & {16.2} & {27.9} \tabularnewline
\cline{2-11}
    & \textbf{FASTER~\cite{tordesillas2022faster}}  & Hard & \best{10} / 10 & {20.4} & \best{105.2} & {5.6} & {11.8} & {17.4} & {34.8} & {32.6} \tabularnewline
\cline{2-11}
    & \textbf{DYNUS}                                & Hard & \best{10} / 10 & \best{15.3} & {105.4}  & {4.2} & {2.4} & {1.8} & {8.4} & {31.1} \tabularnewline
  \bottomrule
  \end{tabular}}
  \par\end{centering}
  \vspace{-1em}
\end{table*}

\subsection{DYNUS in Dynamic Environments}

\begin{figure}[t]
  \centering
  \includegraphics[width=\columnwidth]{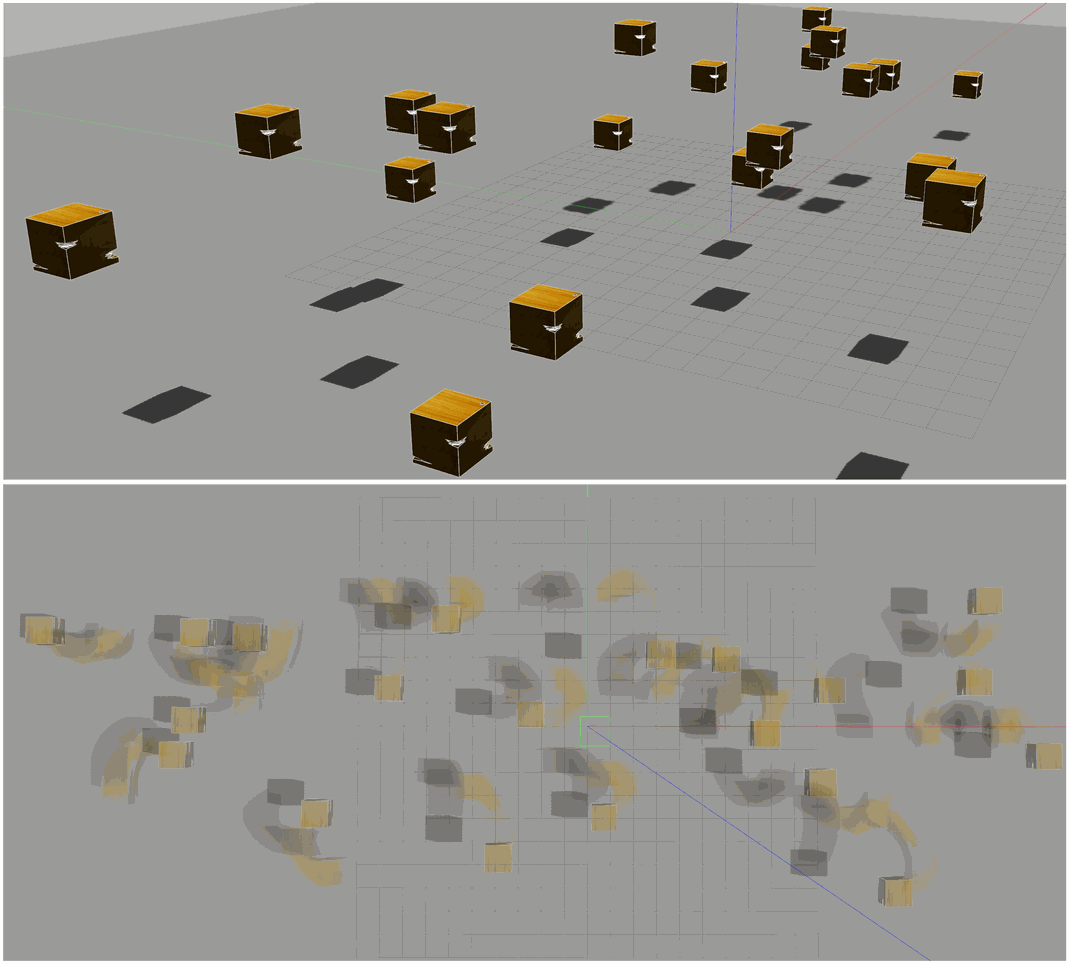}
  \caption{Dynamic Obstacles: Shows the Gazebo simulation environment with 20 dynamic obstacles following a trefoil trajectory. The second figure captures dynamic obstacles' movement for the duration of 8 seconds.}
  \label{fig:global_planner_benchmarking_gazebo}
  \vspace{-1em}
\end{figure}

\begin{figure}[h]
  \centering
  \includegraphics[width=\columnwidth]{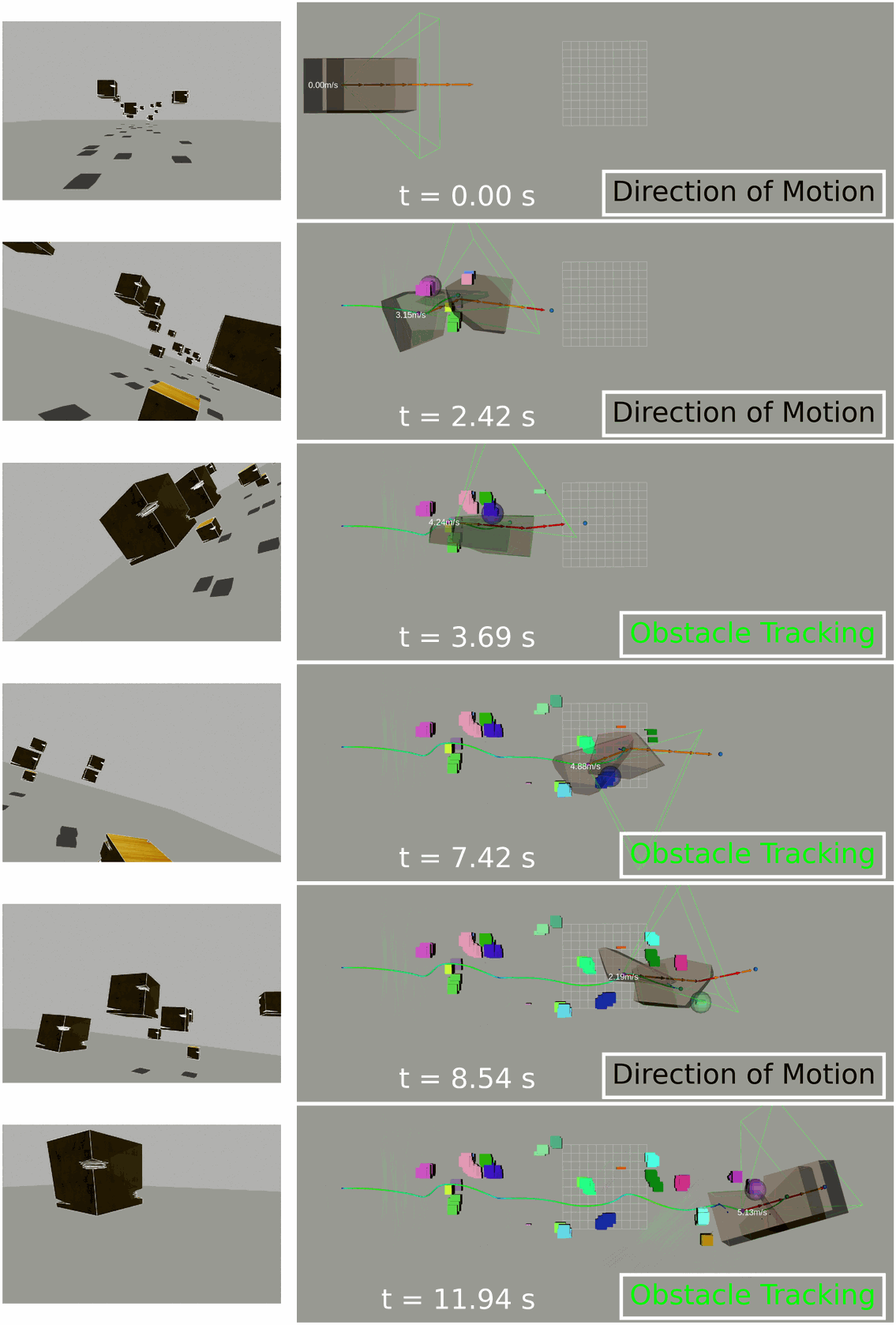}
  \caption{Dynamic Obstacles: Shows the result of one of DYNUS's simulations in a dynamic environment with 20 trefoil knot obstacles showing that the approach maintains high-speed motion while avoiding collisions with dynamic obstacles.
  The left figures are the camera view, and the right figures show tracked dynamic obstacles, DYNUS' field of view, and the agent's trajectory at each time step. The colored boxes represent the dynamic obstacles tracked by AEKF, and the green pyramid represents the agent's field of view. The transparent red and green boxes around the agent are the temporal safe corridors (convex hulls) for exploratory and safe trajectories, respectively. DYNUS successfully navigates through the dynamic obstacles while balancing tracking obstacles and visibility in the direction of motion.}
  \label{fig:global_planner_benchmarking_result}
  \vspace{-2em}
\end{figure}

This section showcases the ability of DYNUS to navigate in dynamic unknown environments and compares DGP to Dynamic A*.   
Fig.~\ref{fig:global_planner_benchmarking_gazebo} shows the simulation environment in which we generated dynamic obstacles following a trefoil knot trajectory, with randomized parameters including initial position, scale, time offset, and speed. 
A total of 20 obstacles (modeled as $1\,\mathrm{m}$ cubes) were spawned at evenly spaced intervals along the $x$-axis, with their $y$ and $z$ positions uniformly sampled from a predefined range. 
Each obstacle follows a parametric trefoil trajectory with different spatial and time scaling. 
The dynamic constraints are $\vect{v_{\text{max}}} = 10.0$ \SI{}{\m/\s}, $\vect{a_{\text{max}}} = 20.0$ \SI{}{\m/\s\squared}, and $\vect{j_{\text{max}}} = 30.0$ \SI{}{\m/\s\cubed}.
Table~\ref{tab:dynus_global_planner_benchmarking} summarizes the benchmarking results, and Fig.~\ref{fig:global_planner_benchmarking_result} shows the detailed time sequence of DYNUS's navigation, where the agent optimally balances tracking obstacles and visibility in the direction of motion.
Both DGP and Dynamic A* achieved a 100\% success rate, and DGP outperformed Dynamic A* in terms of travel time and computation time.
Table~\ref{tab:dynus_global_planner_benchmarking} also shows DYNUS's fast yaw computation time \textemdash its decoupled yaw optimization approach achieves faster computation than coupled yaw optimization methods such as PUMA~\cite{kondo2024puma}.

\begin{table*}[htbp]
  \caption{Global Planner Benchmarking in Dynamic Environments: Both DGP and Dynamic A* achieve a \SI{100}{\%} success rate. However, DGP achieves significantly faster global planning times by selectively running either JPS or Dynamic A*, depending on potential collisions with dynamic obstacles. Furthermore, the decoupled yaw optimization in DYNUS enables much faster computation compared to coupled approaches~\cite{kondo2024puma, tordesillas2022panther}.}
  \label{tab:dynus_global_planner_benchmarking}
  \begin{centering}
  \renewcommand{\arraystretch}{1.3}
  \resizebox{\textwidth}{!}{
    \begin{tabular}{>{\centering\arraybackslash}m{0.2\columnwidth} 
                    >{\centering\arraybackslash}m{0.1\columnwidth} 
                    >{\centering\arraybackslash}m{0.15\columnwidth} 
                    >{\centering\arraybackslash}m{0.18\columnwidth} 
                    >{\centering\arraybackslash}m{0.15\columnwidth} 
                    >{\centering\arraybackslash}m{0.18\columnwidth}
                    >{\centering\arraybackslash}m{0.15\columnwidth}
                    >{\centering\arraybackslash}m{0.15\columnwidth}
                    >{\centering\arraybackslash}m{0.15\columnwidth}
                    >{\centering\arraybackslash}m{0.15\columnwidth}
                    >{\centering\arraybackslash}m{0.15\columnwidth} 
                    >{\centering\arraybackslash}m{0.17\columnwidth}}
      \toprule
      & \multirow{3}{*}{\textbf{Success}} & \multicolumn{2}{c}{\textbf{Performance}} & \multicolumn{7}{c}{\textbf{Computation Time [ms]}} \tabularnewline 
      \cmidrule(lr){3-4} \cmidrule(lr){5-12}
      && \textbf{Travel Time [s]} & \textbf{Path Length [m]} & \textbf{Global Planning} & \textbf{Replan Total Comp.} & \textbf{CVX Decomp.} & \textbf{Exp. Traj.} & \textbf{Safe Trajs.} & \textbf{Yaw Seq.} & \textbf{Yaw Fit.} \tabularnewline
      \midrule
      Dynamic A* & \best{10} / 10 & {16.6} & \best{57.7} & {40.9} & {101.8} & {10.2} & {5.7} & {2.0} & {5.6} & {6.4} \tabularnewline
      \midrule
      DGP & \best{10} / 10 & \best{15.5} & {58.6} & \best{13.5} & \best{76.6} & {11.0} & {5.7} & {1.9} & {7.2} & {6.4} \tabularnewline
      \bottomrule
    \end{tabular}
  }
  \par\end{centering}
  \vspace{-1em}
\end{table*}

\begin{figure}
  \centering
  \includegraphics[width=\columnwidth]{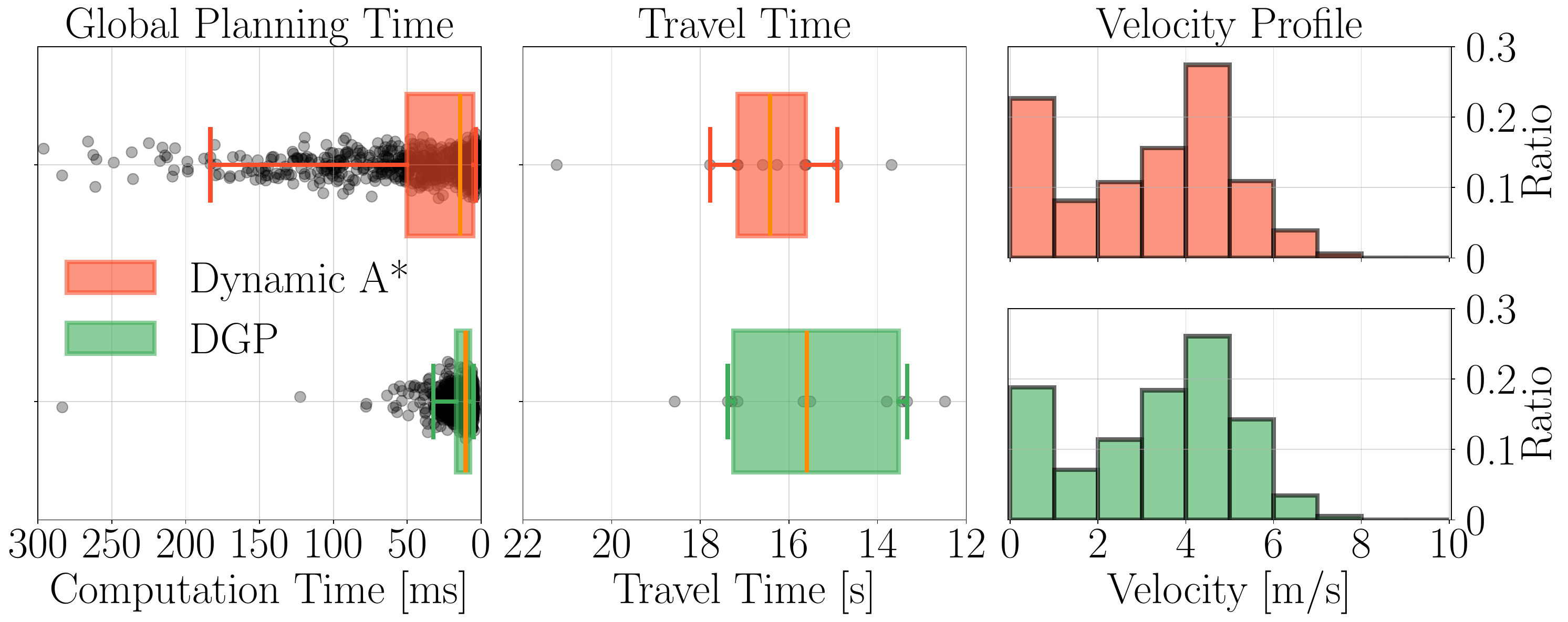}
  \caption{Global planner benchmarking (Computation Time, Travel Time, and Velocity Profile): DGP achieves avg. computation time of \best{13.5}~ms and avg. travel time of \best{15.5}~s; whereas Dynamic A* records \textcolor{red}{40.9}~ms and \textcolor{red}{16.6}~s. DGP demonstrates significantly faster computation while achieving faster travel time. Detailed performance results are summarized in Table~\ref{tab:dynus_global_planner_benchmarking}.}
  \label{fig:global_planner_benchmarking}
  \vspace{-1em}
\end{figure}

\subsection{DYNUS's Performance in Various Environments on UAV, Wheeled Robot, and Quadruped Robot Platforms}

\begin{figure}[htbp]
  \centering
  \includegraphics[width=\columnwidth]{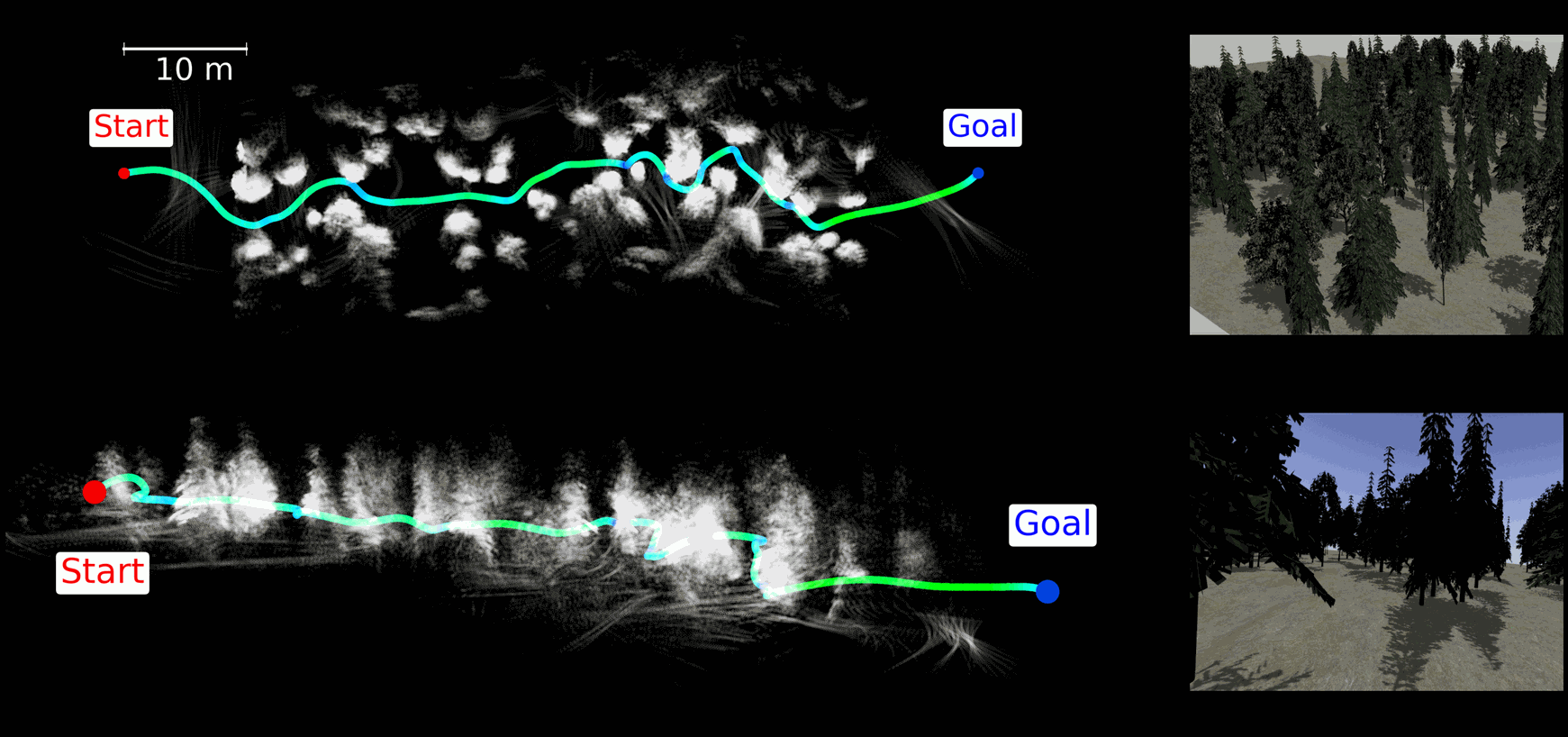}
  \caption{Photo-Realistic Forest: This figure shows the results of the photo-realistic forest simulation. The agent trajectory is color-coded according to the velocity profile.}
  \label{fig:single_agent_performance_big_forest_high_res}
  \vspace{-1em}
\end{figure}

\begin{figure}[htbp]
  \centering
  \includegraphics[width=\columnwidth]{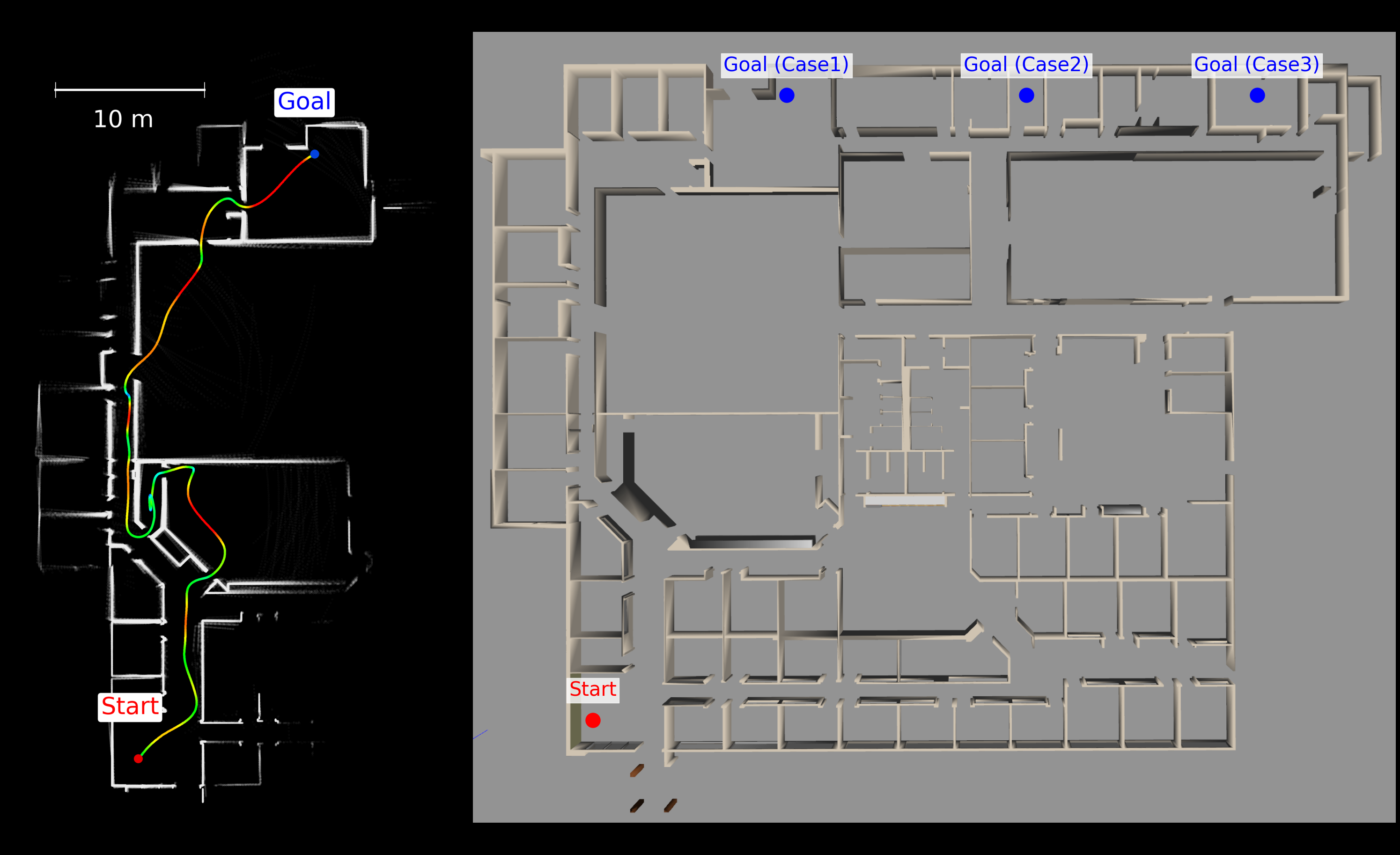}
  \caption{Office: The left figure shows the agent's trajectory and point cloud data in the office environment. The right figure shows the Gazebo simulation environment, start location, and goal locations for each case.}
  \label{fig:single_agent_performance_office_case1}
  \centering
  \includegraphics[width=\columnwidth]{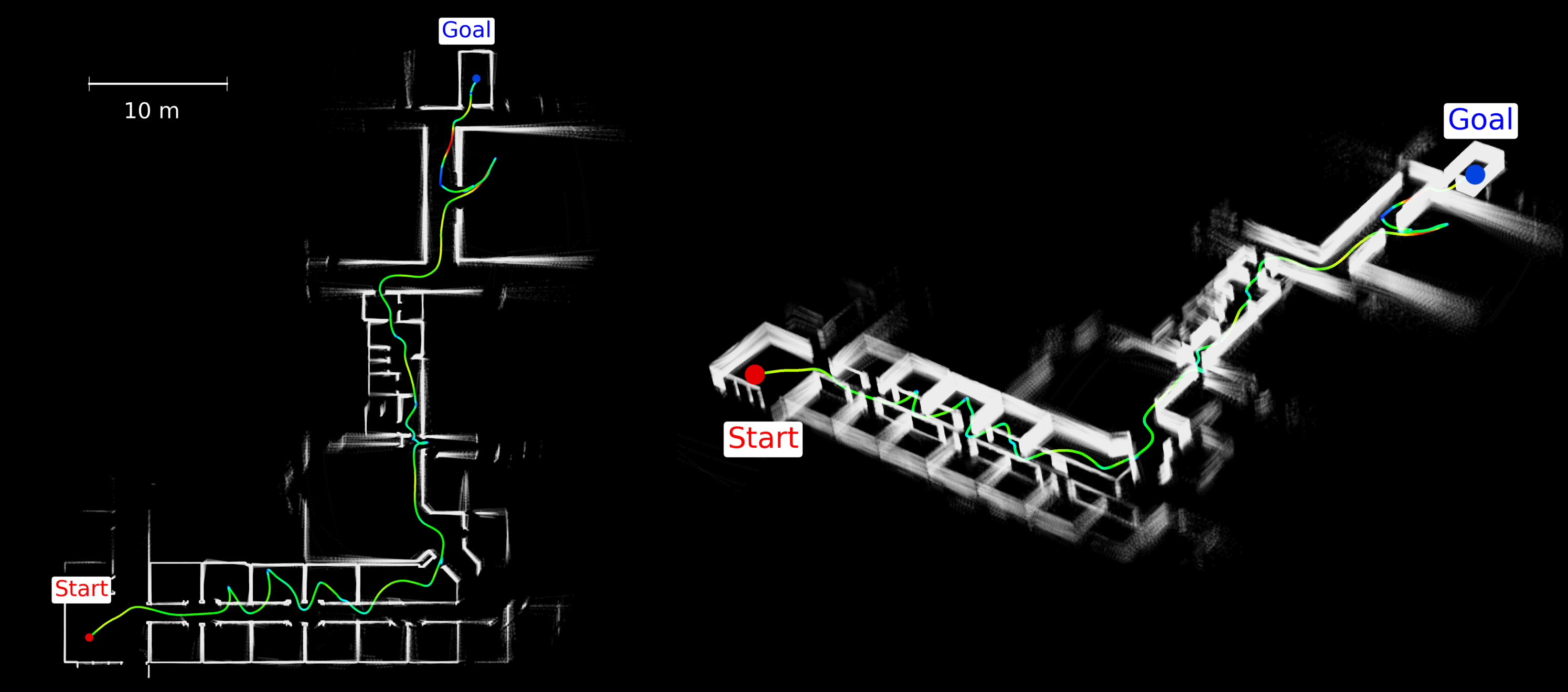}
  \caption{Office Case 2: We assign a goal point farther than in Case~1. DYNUS successfully recovers from multiple dead ends, including small rooms, and reaches the goal.}
  \label{fig:single_agent_performance_office_case2}
  \vspace{-1em}
\end{figure}

This section tests DYNUS's ability to navigate in various environments, including photorealistic forests, office, and cave environments, using the UAV, wheeled robot, and quadruped robot platforms.
For each environment, the agent's trajectory, point cloud, and camera view are presented.  
The trajectory is color-coded according to the velocity profile.

First, we test DYNUS in a \textbf{photo-realistic forest environment}.  
Unlike the previous simulation environments, this setting includes trees with branches and smaller obstacles.  
This test evaluates the ability of DYNUS to navigate \uline{highly cluttered, unknown, static environments}.  
Fig.~\ref{fig:single_agent_performance_big_forest_high_res} shows DYNUS navigating through a dense forest populated with realistic high-resolution trees.  

We then tested DYNUS in an \textbf{office} environment that contains many dead ends and walls.  
We gave the agent three different goals to test its ability to navigate the office environment from the same starting position.
This environment evaluates DYNUS's ability to fly in \uline{unknown, confined spaces} and its ability to escape and recover from dead-end situations. 
Fig.~\ref{fig:single_agent_performance_office_case1}, ~\ref{fig:single_agent_performance_office_case2}, and 
\ref{fig:single_agent_performance_office_case3} showcases the office simulation environment and results.
DYNUS encounters numerous dead-ends but successfully recovers and reroutes to reach the goal.

\begin{figure}[h]
  \centering
  \begin{tikzpicture}
    \node[anchor=south west,inner sep=0] (image) at (0,0) 
    {\includegraphics[width=\columnwidth]{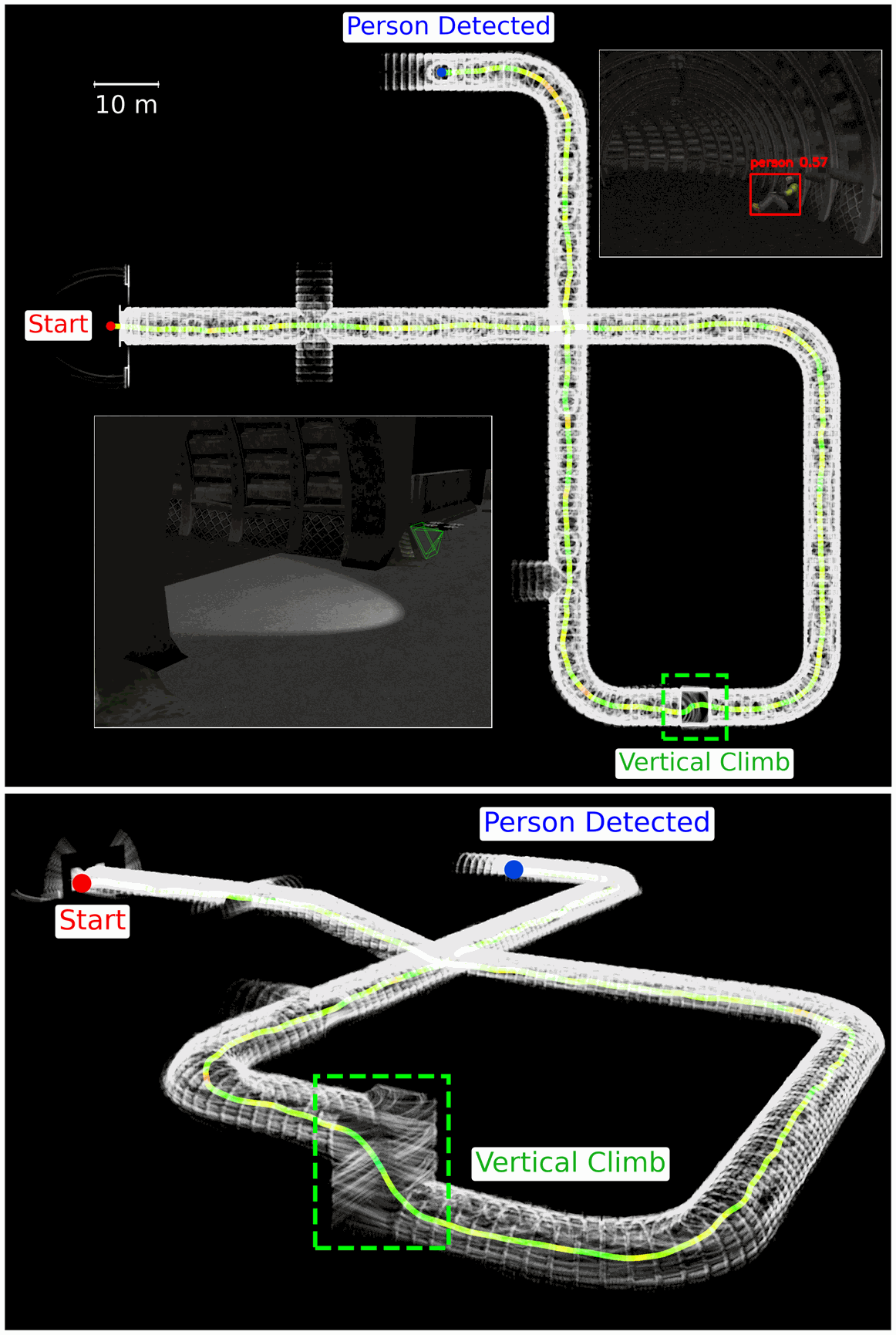}};
    \begin{scope}[x={(image.south east)},y={(image.north west)}]
    \end{scope}
  \end{tikzpicture}
  \caption{Cave: Figure illustrates DYNUS's performance in the cave environment.  
    It shows the point cloud data generated by the agent and the corresponding trajectory taken during exploration.  
    The top-right figure shows the detection of a person in the cave using YOLOv11, which serves as the termination condition for the exploration task. 
    The bottom-left figure in the top figure highlights the robot's onboard illuminating light.}
  \label{fig:single_agent_performance_cave}
  \vspace{-1em}
\end{figure}

We also test DYNUS's exploration capabilities in a \textbf{cave} environment, as shown in Fig.~\ref{fig:single_agent_performance_cave}. 
The cave is extremely confined, and DYNUS is tasked with exploring the space and locating a person inside.  
Person detection is performed in real-time using YOLOv11~\cite{khanam2024yolov11}.  
This test evaluates DYNUS's ability to \uline{explore unknown, confined environments} and detect target objects.  
For the cave simulation, the exploration algorithm described in Section~\ref{sec:frontier} is utilized to guide the agent through the environment.  
The figures in Fig.~\ref{fig:single_agent_performance_cave} display the point cloud data generated by the agent and the corresponding exploration trajectory.  
The top-right figure shows the detection of a person by the YOLOv11 model, and the bottom-left figure in the top figure shows the drone's onboard light illuminating the cave.  
DYNUS successfully explores the space while avoiding collisions with walls and safely ascending the vertical shaft.

\subsection{2D Performance \textemdash Wheeled Robot and Quadruped Robot}

\begin{figure}[htbp]
  \centering
  \includegraphics[width=\columnwidth]{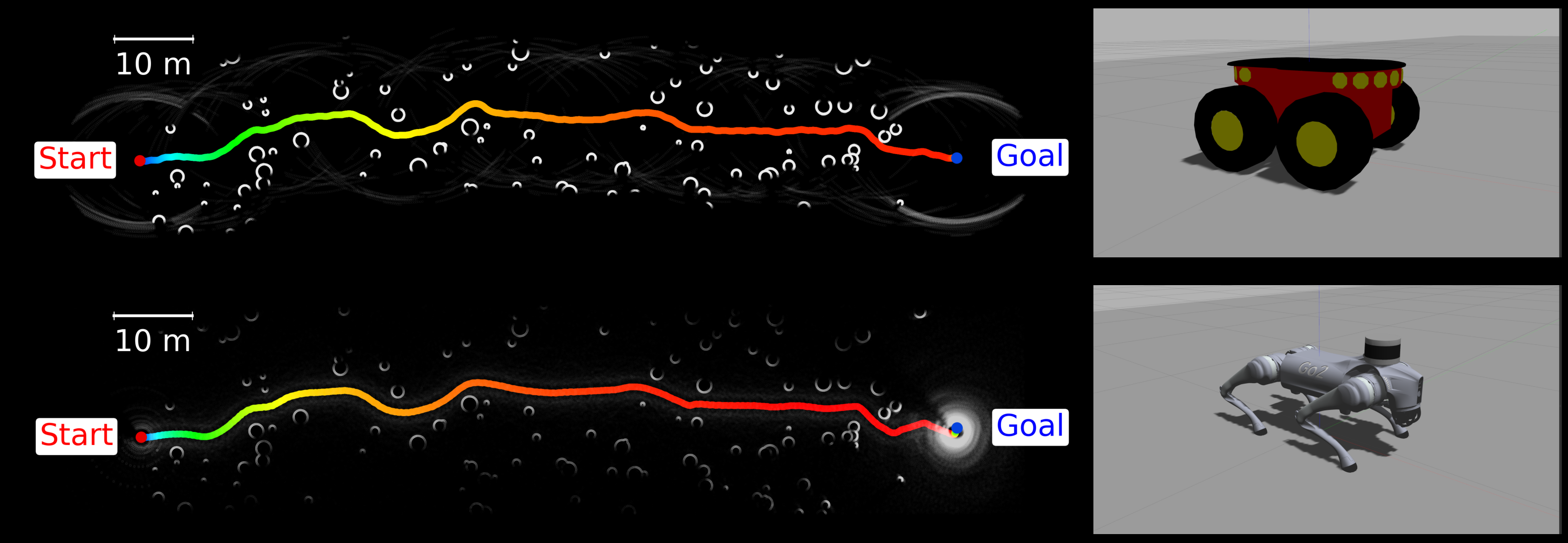};
  \caption{Ground Robot Performance: DYNUS enables both the wheeled robot and quadruped robot to navigate through a cluttered forest environment.  
    The left two figures show the point cloud data generated by the robots, and the right two figures show the corresponding trajectories.}
  \label{fig:ground_robot_performance}
  \vspace{-1em}
\end{figure}

To evaluate DYNUS's collision-avoidance ability on different platforms, we tested it on both a wheeled ground robot and a quadruped robot.  
DYNUS's trajectory for the robots is tracked using a geometric controller~\cite{rober20223d}.  
The simulations are performed in a cluttered static forest environment (Fig.~\ref{fig:sota_benchmarking_gazebo}). 
The quadruped robot is simulated using the Unitree Go2 ROS2 simulator.  
The dynamic constraints for the wheeled robot are set to \( \vect{v_{\text{max}}} = 1.0 \) \SI{}{\m/\s}, \( \vect{a_{\text{max}}} = 5.0 \) \SI{}{\m/\s\squared}, and \( \vect{j_{\text{max}}} = 10.0 \) \SI{}{\m/\s\cubed}.  
For the quadruped robot, the constraints are set to \( \vect{v_{\text{max}}} = 0.5 \) \SI{}{\m/\s}, \( \vect{a_{\text{max}}} = 5.0 \) \SI{}{\m/\s\squared}, and \( \vect{j_{\text{max}}} = 10.0 \) \SI{}{\m/\s\cubed}.  
These constraints are lower than those used for the quadrotor to mitigate tracking errors introduced by the lower-level controller.  
For the Unitree Go2, we used a Velodyne LiDAR sensor.  
Fig.~\ref{fig:ground_robot_performance} shows that DYNUS successfully enables both ground robots to navigate the environment.

%% file: paper/11_hardware_experiments.tex
\section{Hardware Experiments}\label{sec:hardware-experiments}

To evaluate the performance of DYNUS, we conduct hardware experiments on three platforms: a UAV, a wheeled robot, and a quadruped robot.  
For perception, we use a Livox Mid-360 LiDAR sensor, and for localization, we use onboard DLIO~\cite{chen2023dlio}. 
DYNUS runs on an Intel\textsuperscript{\texttrademark} NUC 13 across all platforms.  
For low-level UAV control, we use PX4~\cite{meier2015px4} on a Pixhawk flight controller~\cite{meier2011pixhawk}.  
All perception, planning, control, and localization modules run onboard in real time, enabling fully autonomous operations.

\subsection{UAV in Static Environments}

\begin{figure}[htbp]
    \centering
    \begin{minipage}[c]{0.35\columnwidth}
        \includegraphics[width=\linewidth, clip, trim=0 10 0 10]{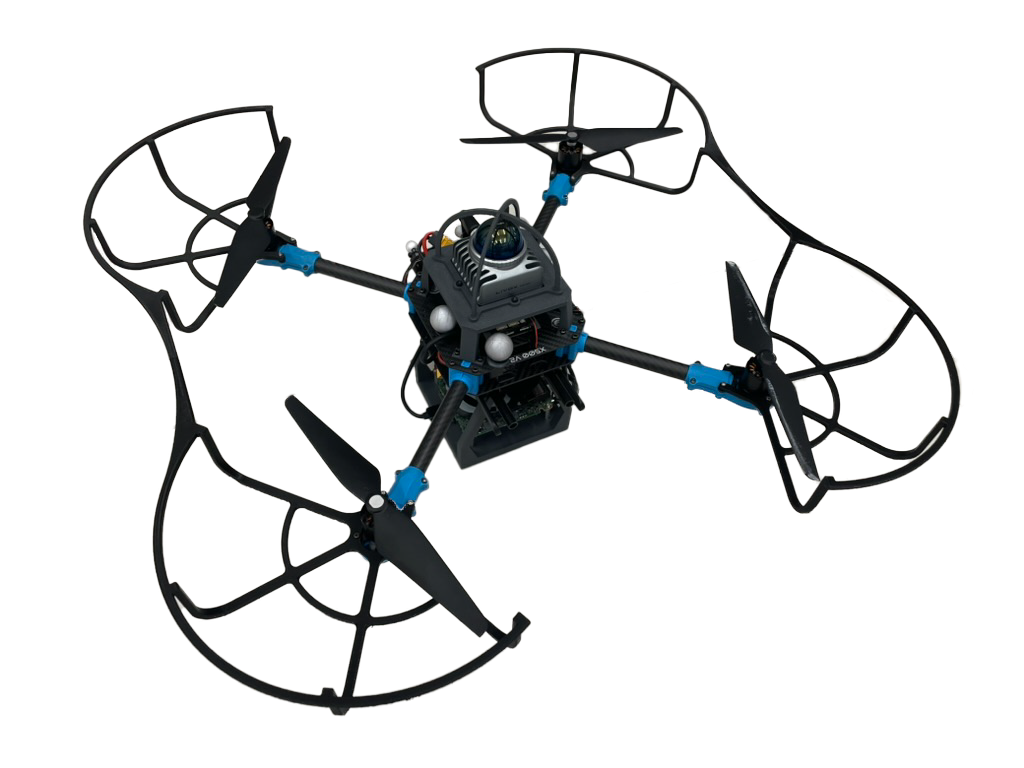}
    \end{minipage}%
    \hspace{1em}%
    \begin{minipage}[c]{0.6\columnwidth}
        \captionof{figure}{\textbf{UAV platform for hardware experiments.} 
        The Holybro PX4 Development Kit X500 is equipped with a Pixhawk flight controller and protective propeller guards. 
        A Livox Mid-360 LiDAR is mounted on top for 360-degree perception, and DLIO~\cite{chen2023dlio} is used for real-time lidar-inertial localization. 
        All modules—perception, planning, control, and localization—run onboard on an Intel\textsuperscript{\texttrademark} NUC 13, enabling fully autonomous navigation.}
        \label{fig:hardware_uav_picture}
    \end{minipage}
    \vspace{-3em}
\end{figure}

We first evaluate DYNUS on a UAV operating in static indoor environments. 
Fig.~\ref{fig:hardware_uav_picture} shows the UAV platform used in our experiments. 
We custom-designed and 3D-printed a protective frame for the propellers, a shelf for the Intel NUC, and a mounting stage for the Livox Mid-360 with additional housing for protection.

\begin{figure}[htbp]
    \centering
    \includegraphics[width=\columnwidth]{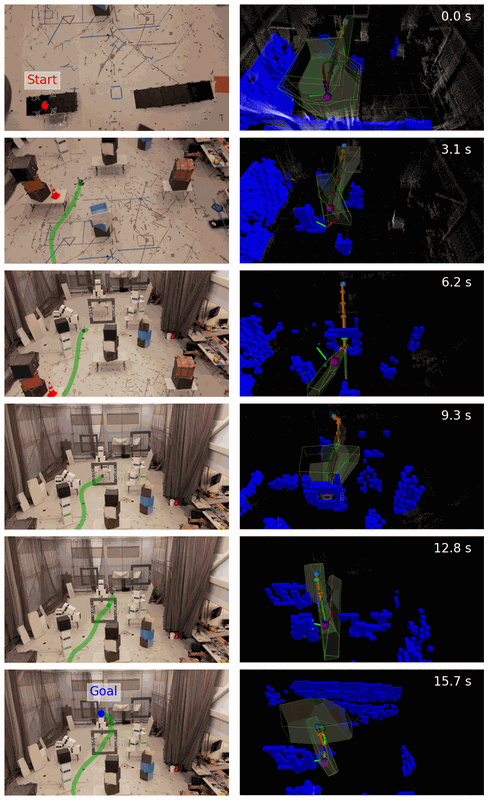}
    \caption{
        Experiment 1: UAV hardware experiment in a cluttered static environment.
        Left: Images showing the UAV flying through the obstacle course.  
        Right: RViz visualization. The purple dot marks the planning start position, the TF marker indicates the UAV's current pose, and the orange arrows show the planned global path.  
        The green dot corresponds to Point E (see Fig.~\ref{fig:trajectory_planning_framework}), and the blue dot denotes the global goal.  
        Green convex hulls represent the safe corridor used for safe trajectories, while red convex hulls indicate the safe corridor for exploratory trajectories.  
        The occupancy map and LiDAR point cloud are also shown. The UAV completes the mission successfully, navigating safely through all obstacles.
    }
    \label{fig:hardware_uav_static}
    \vspace{-1em}
\end{figure}

\begin{figure}[htbp]
    \centering
    \includegraphics[width=0.98\columnwidth, trim=0 30 0 40, clip]{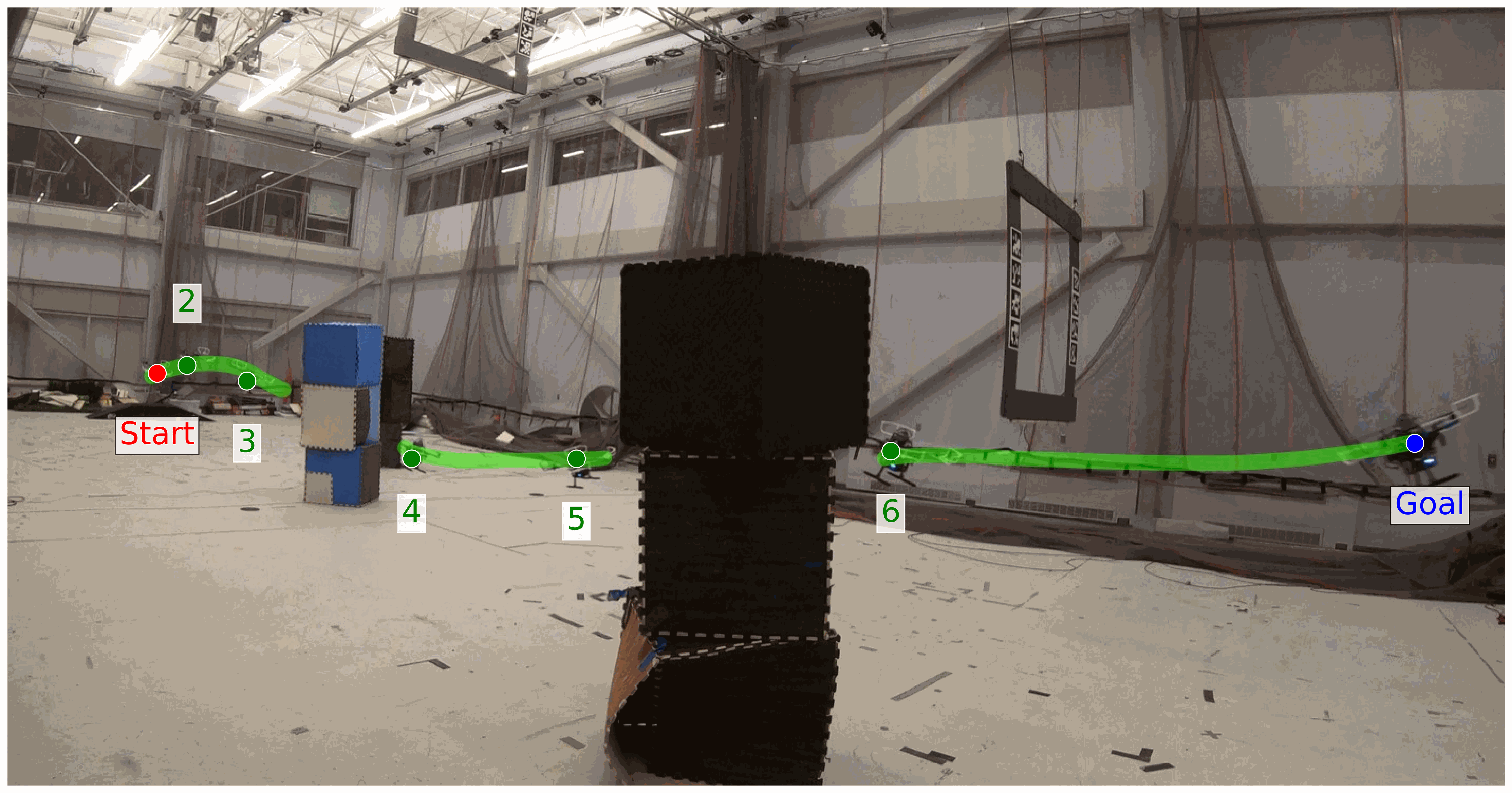}
    \caption{
    Experiment 2: UAV fast hardware experiment in a static environment.
    The UAV achieved a maximum of \SI{4.9}{\m/\s} while avoiding static obstacles in the environment.}
    \label{fig:hardware_uav_static_fast}
    \vspace{-1em}
\end{figure}

Static obstacles are placed in an $8\,\text{m} \times 20\,\text{m}$ area, and the UAV is tasked with flying from the start position at $(0, 0, 1.5)\,\text{m}$ to the goal at $(18, 0, 1.5)\,\text{m}$.  
Experiment 1 has more static obstacles, making the space cluttered, while Experiment 2 has fewer obstacles.
In Experiment 1, the dynamic constraints are set to $v_{\max} = 2.0\,\text{m/s}$, $a_{\max} = 5.0\,\text{m/s}^2$, and $j_{\max} = 10.0\,\text{m/s}^3$. 
In Experiment 2, the dynamic constraints are set to $v_{\max} = 8.0\,\text{m/s}$, $a_{\max} = 15.0\,\text{m/s}^2$, and $j_{\max} = 30.0\,\text{m/s}^3$ for faster flight.
Fig.~\ref{fig:hardware_uav_static} shows the resulting trajectory visualized in RViz, along with the occupancy map, point cloud, and safe corridors used for planning in Experiment 1.  
The UAV successfully completes the mission by navigating through narrow passages and avoiding all static obstacles.
Fig.~\ref{fig:hardware_uav_static_fast} shows the trajectory of Experiment 2, where DYNUS achieved a maximum of \SI{4.9}{\m/\s} and successfully reached the goal.

\subsection{UAV in Dynamic Environments}

To evaluate DYNUS in dynamic environments, we conducted hardware experiments involving one and two dynamic obstacles that were created by attaching approximately \SI{2.0}{\meter}-tall foam rectangular boxes to a wheeled robot.  
In Experiments 3 and 4, a single obstacle moved at a constant speed of \SI{0.4}{\m/\s} within an $8\,\text{m} \times 10\,\text{m}$ area, periodically blocking the UAV’s path.  
Experiments 5 and 6 have one dynamic obstacle and several static obstacles randomly placed in an $8\,\text{m} \times 20\,\text{m}$ area.  
In Experiments 7 to 10, two dynamic obstacles and multiple static obstacles were placed in the same $8\,\text{m} \times 20\,\text{m}$ environment.  

For Experiments 3 to 8, the dynamic constraints were set to $v_{\max} = 2.0\,\text{m/s}$, $a_{\max} = 5.0\,\text{m/s}^2$, and $j_{\max} = 30.0\,\text{m/s}^3$.  
In Experiments 9 and 10, these constraints were increased to $v_{\max} = 5.0\,\text{m/s}$, $a_{\max} = 10.0\,\text{m/s}^2$, and $j_{\max} = 30.0\,\text{m/s}^3$ for faster flight.

\begin{figure}[htbp]
    \centering
    \includegraphics[width=\columnwidth]{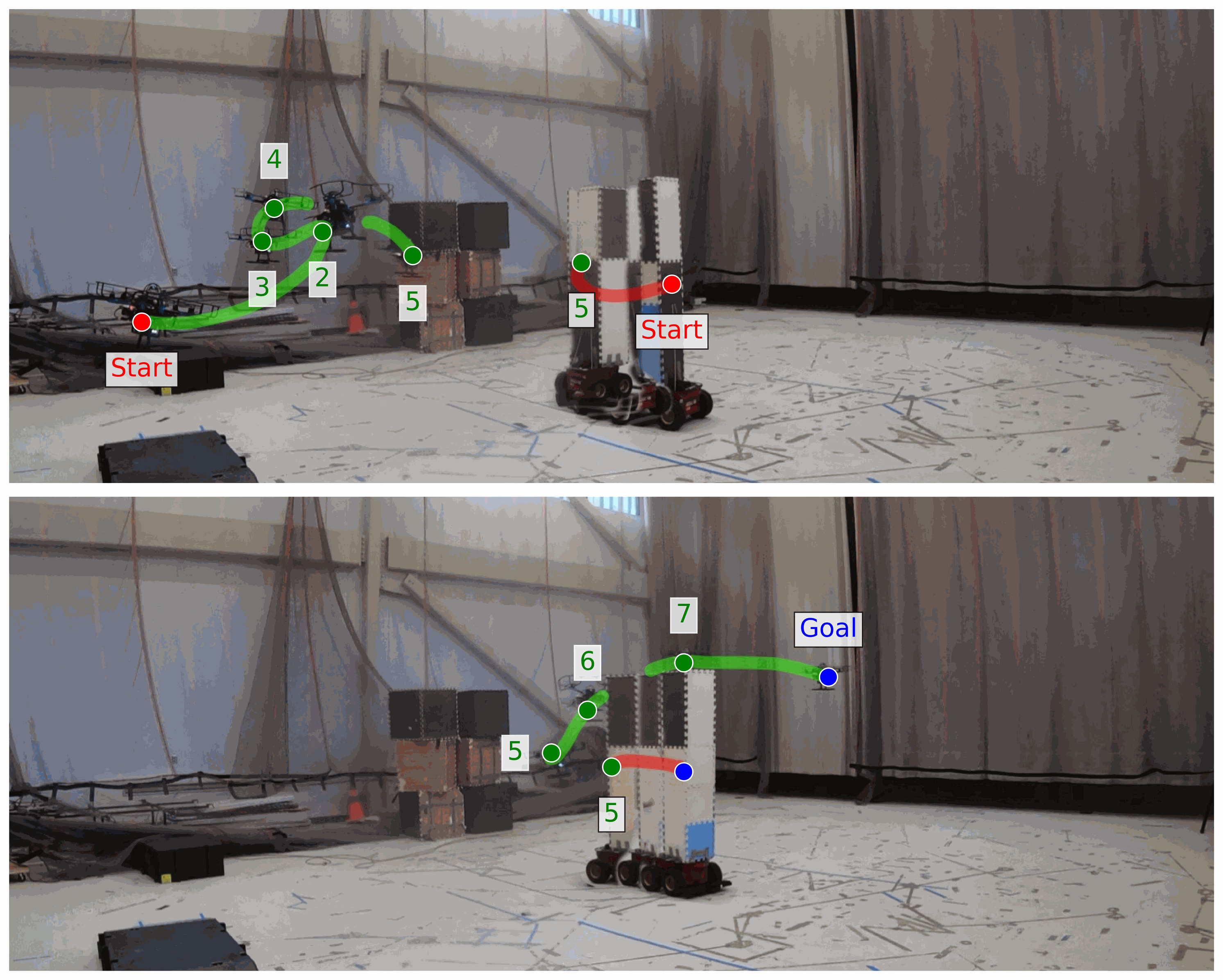}
    \caption{Experiment 3: UAV navigates around a single moving obstacle traveling at \SI{0.4}{\m/\s} in an open area.}
    \label{fig:hardware_uav_dynamic_exp3}
    \vspace{-1em}
\end{figure}

\begin{figure}[htbp]
    \centering
    \includegraphics[width=\columnwidth]{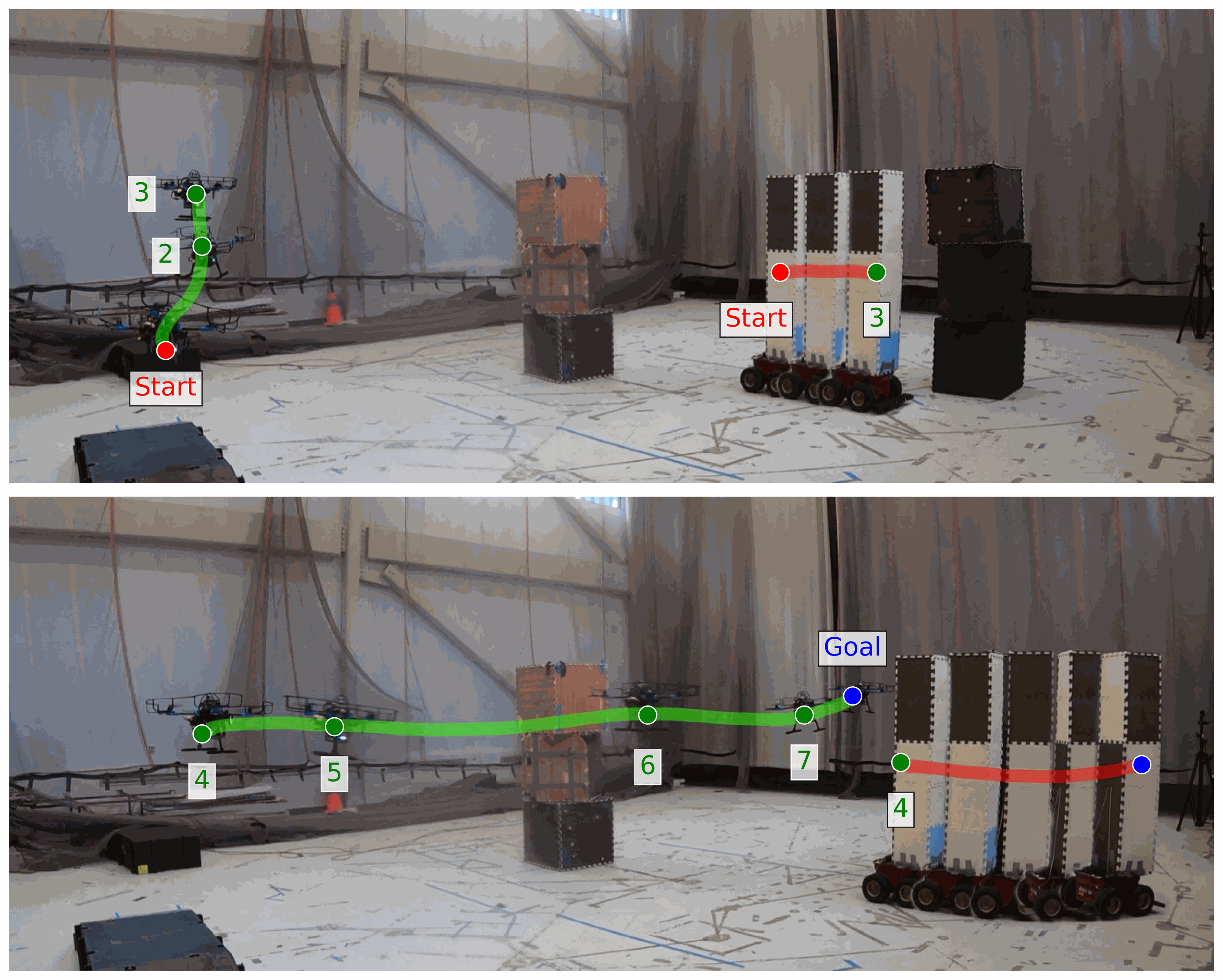}
    \caption{Experiment 5: UAV operates with one dynamic obstacle and randomly placed static obstacles.}
    \label{fig:hardware_uav_dynamic_exp5}
    \vspace{-1em}
\end{figure}

\begin{figure}[htbp]
    \centering
    \includegraphics[width=\columnwidth]{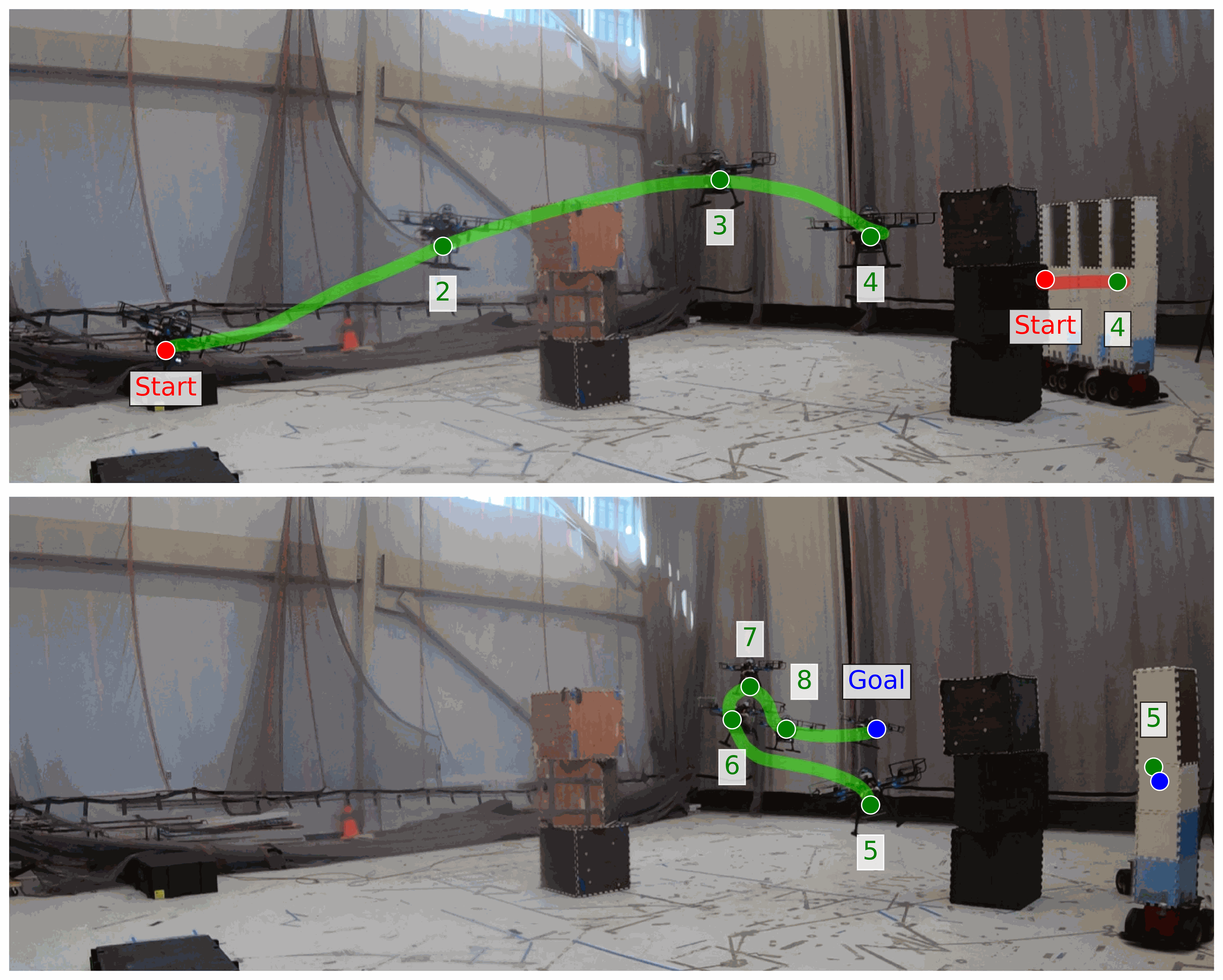}
    \caption{Experiment 6: Similar to Experiment 5, the UAV handles mixed static and dynamic obstacles in a cluttered environment.}
    \label{fig:hardware_uav_dynamic_exp6}
    \vspace{-1em}
\end{figure}

\begin{figure}[htbp]
    \centering
    \includegraphics[width=\columnwidth]{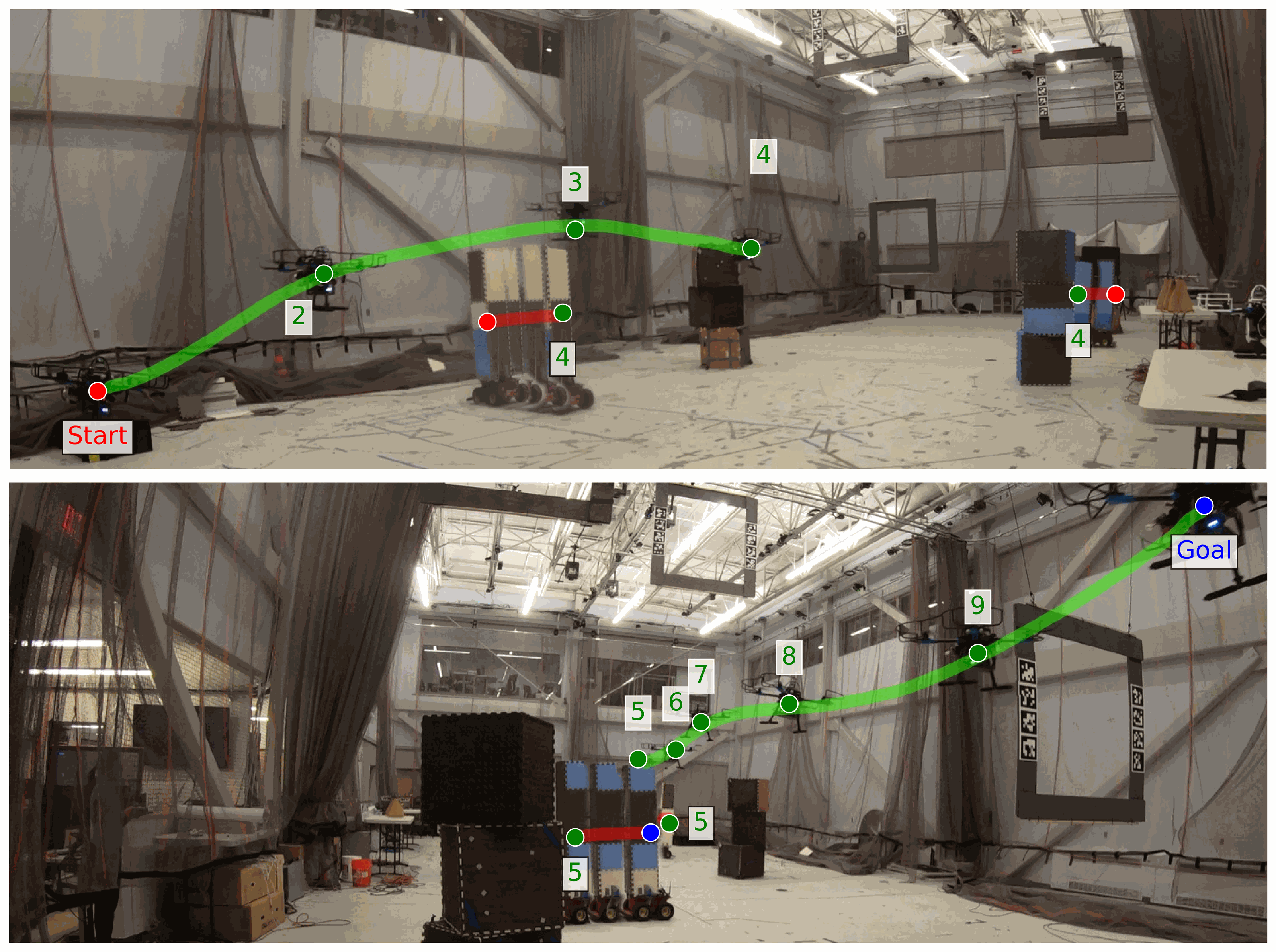}
    \caption{Experiment 7: UAV navigates through an environment with two moving obstacles and additional static obstacles.}
    \label{fig:hardware_uav_dynamic_exp7}
    \vspace{-1em}
\end{figure}

\begin{figure}[htbp]
    \centering
    \includegraphics[width=\columnwidth]{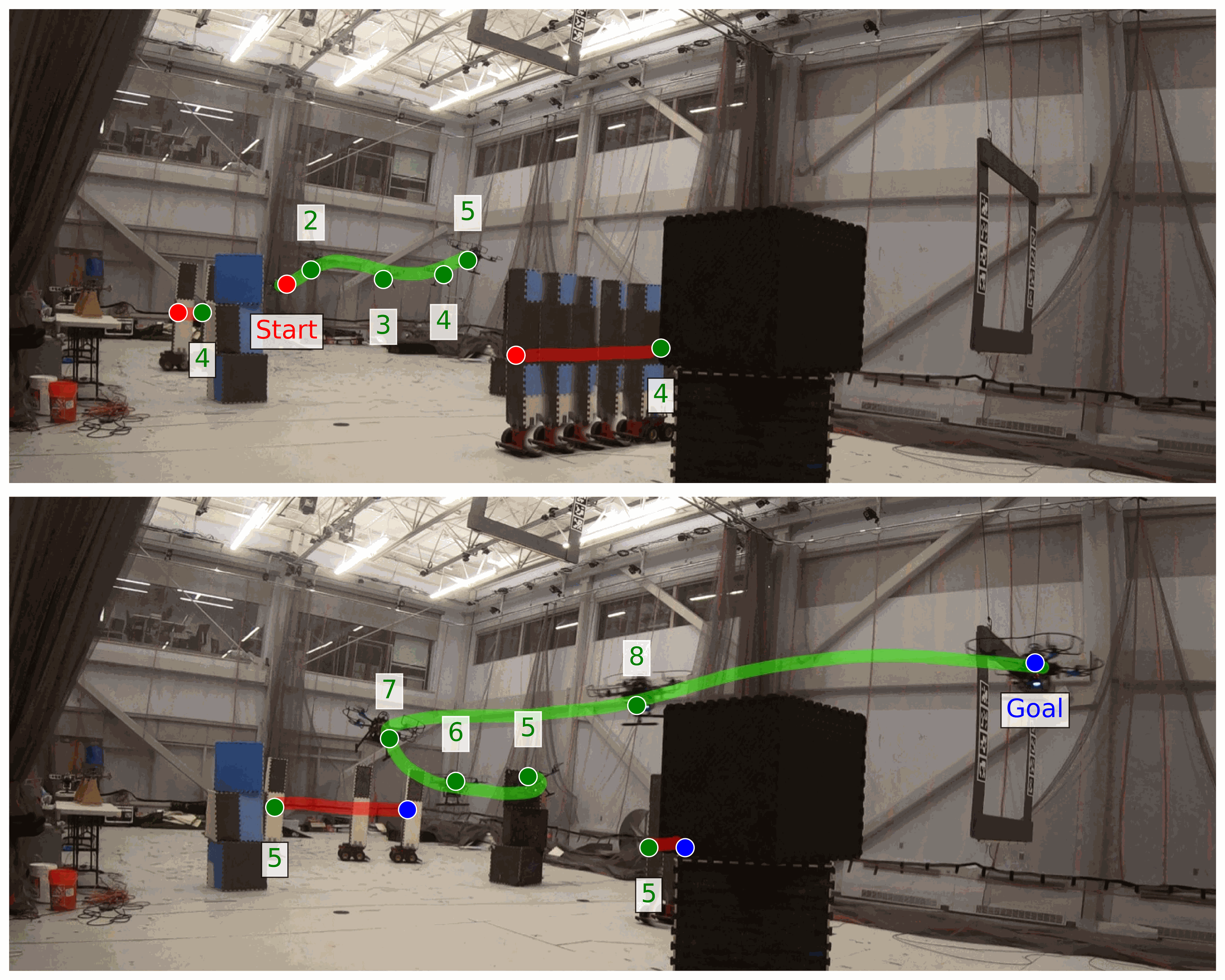}
    \caption{Experiment 8: UAV demonstrates safe flight with two dynamic and several static obstacles.}
    \label{fig:hardware_uav_dynamic_exp8}
    \vspace{-1em}
\end{figure}

\begin{figure}[htbp]
    \centering
    \includegraphics[width=\columnwidth, clip, trim=0 10 0 100]{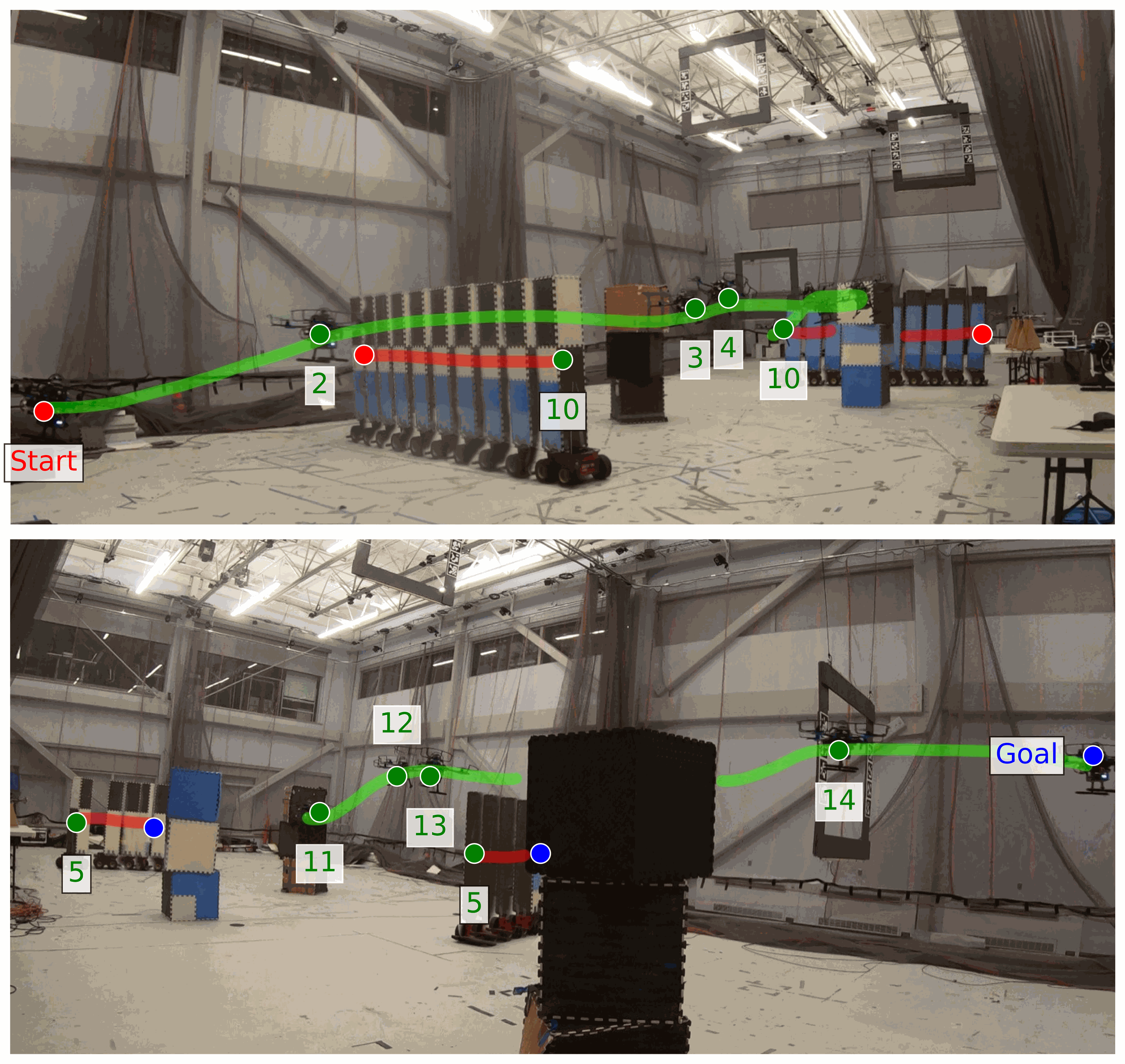}
    \caption{Experiment 9: UAV flies at higher speeds (\SI{5.6}{\m/\s} peak) in a dense environment with two dynamic obstacles.}
    \label{fig:hardware_uav_dynamic_exp9}
    \vspace{-1em}
\end{figure}

\begin{figure}[htbp]
    \centering
    \includegraphics[width=\columnwidth]{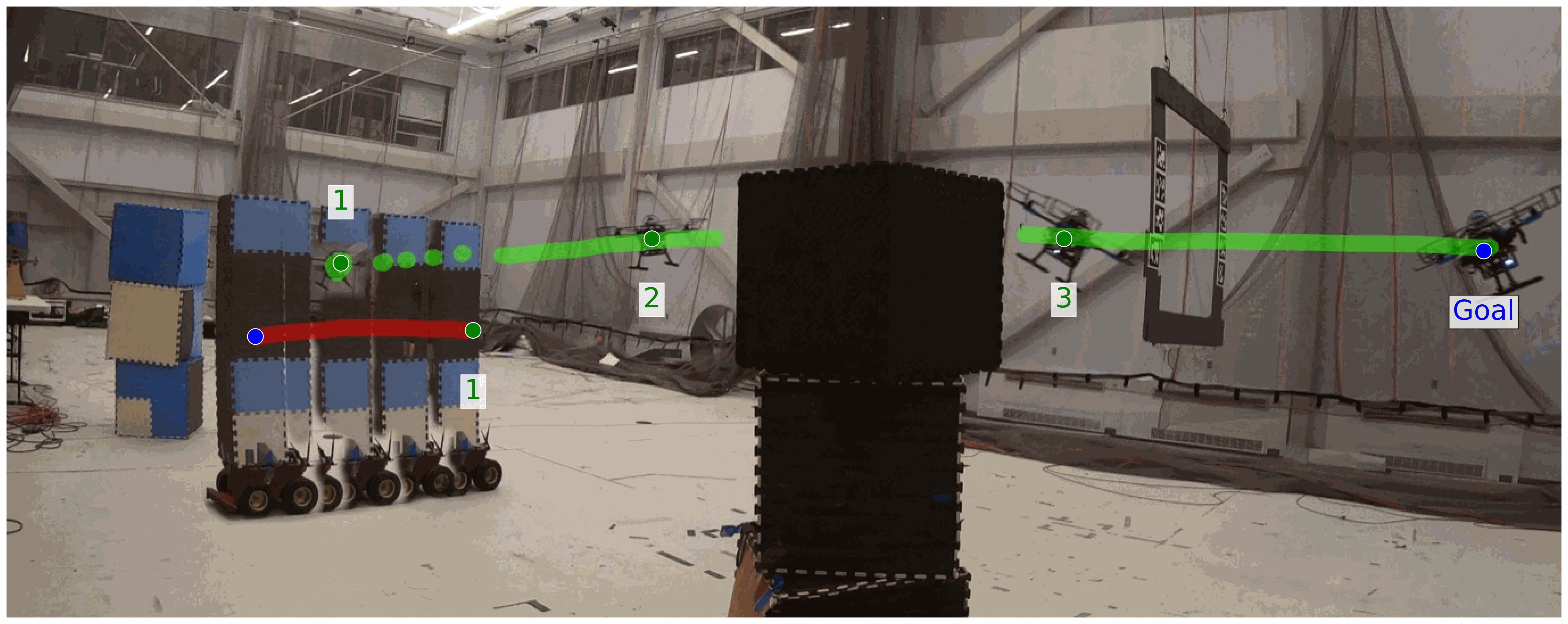}
    \caption{Experiment 10: High-speed UAV navigation with two moving and several static obstacles.}
    \label{fig:hardware_uav_dynamic_exp10}
    \vspace{-1em}
\end{figure}

\begin{figure}[htbp]
    \centering
    \includegraphics[width=\columnwidth, trim=0 0 0 20, clip]{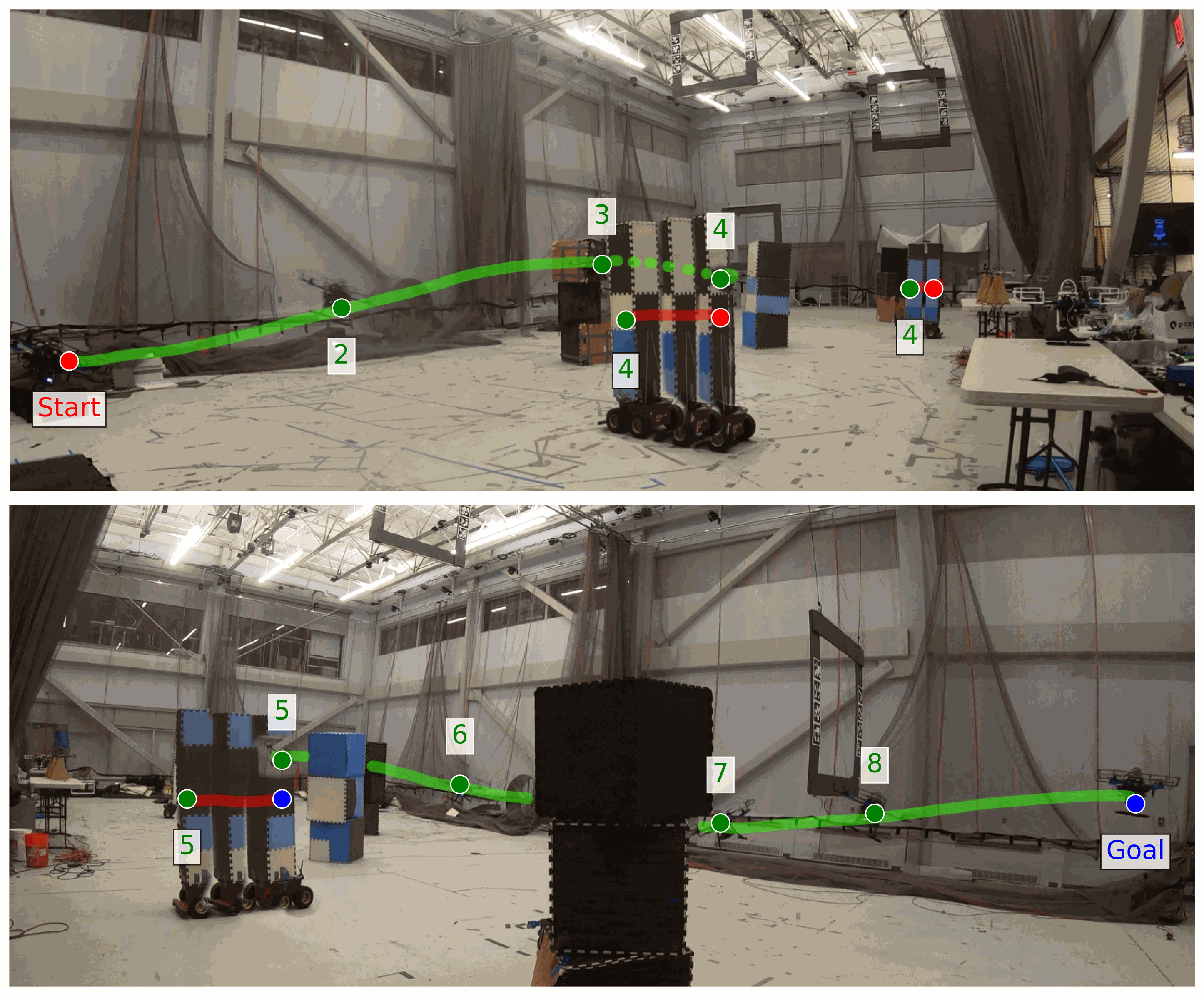}
    \caption{Experiment 11: Additional run demonstrating high-speed navigation in dynamic environments.}
    \label{fig:hardware_uav_dynamic_exp11}
    \vspace{-1em}
\end{figure}

Unlike simulations that utilize both LiDAR and depth cameras, the hardware experiments relied solely on a LiDAR sensor for perception to reduce computational load.  
Figs.~\ref{fig:hardware_uav_dynamic_exp3} to \ref{fig:hardware_uav_dynamic_exp11} illustrate the results of Experiments 3 through 11.  
Each figure presents time-lapse snapshots of the UAV’s trajectory as it navigates around dynamic obstacles.  
In all cases, DYNUS successfully plans and executes safe trajectories in dynamic environments.

For the faster flights in Experiments 9 and 10, the UAV reached maximum speeds of \SI{5.6}{\m/\s} and \SI{4.6}{\m/\s}, respectively.  
In Experiment 9, although DYNUS commanded a maximum velocity of \SI{4.6}{\m/\s} to respect the dynamic constraint of \SI{5.0}{\m/\s}, the lower-level controller commanded a peak velocity of \SI{5.6}{\m/\s} to maintain tracking performance.

\subsection{Ground Robots in Static Environments}

\begin{figure}[htbp]
    \centering
    \subfloat{\includegraphics[width=0.35\columnwidth, trim=50 18 50 0, clip]{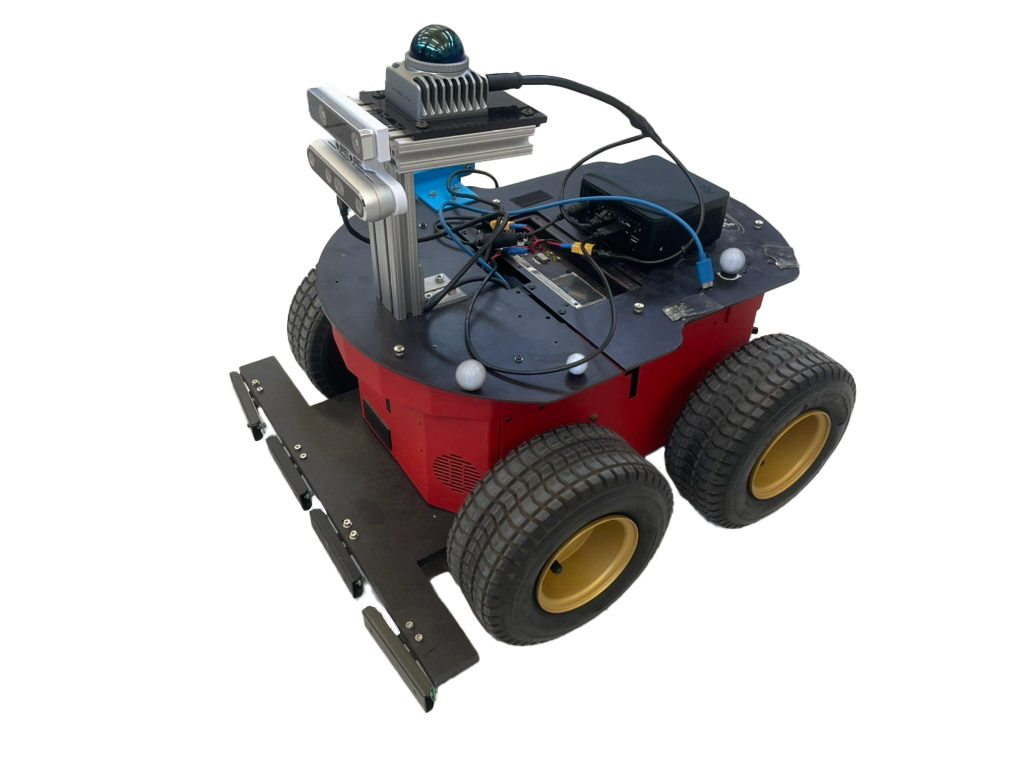}}
    \hfill
    \subfloat{\includegraphics[width=0.63\columnwidth]{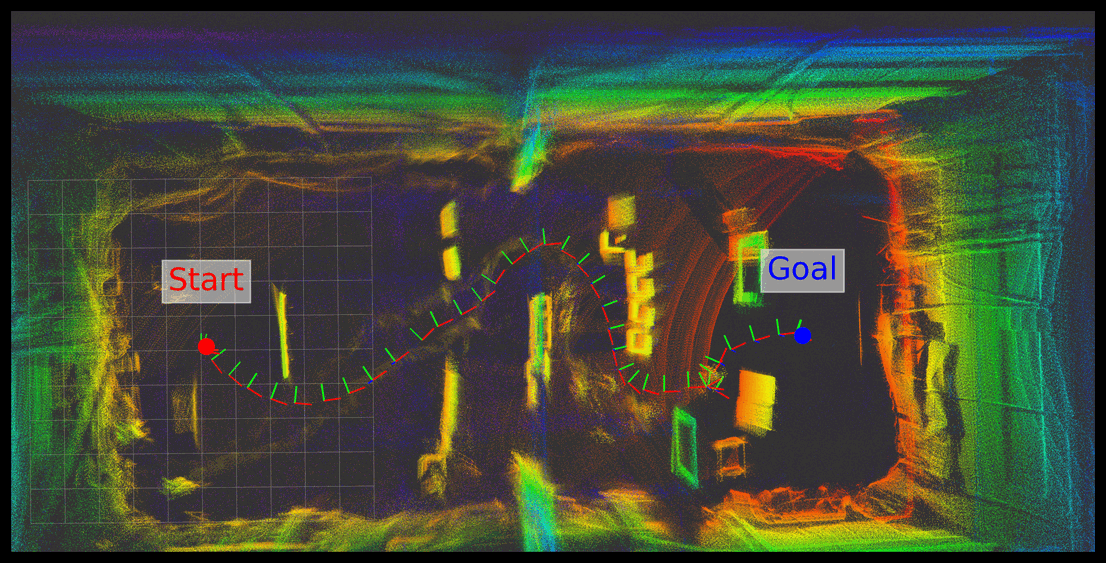}}
    \caption{
        Wheeled Robot in Static Environment.  
        Left: The wheeled robot platform.  
        Right: The history of robot's poses overlaid on the LiDAR point cloud, colored by height.
    }
    \label{fig:hardware_wheeled_static}
    \vspace{-1em}
\end{figure}

\begin{figure}[t]
    \centering    
    \subfloat{\includegraphics[width=0.35\columnwidth, trim=50 30 50 0, clip]{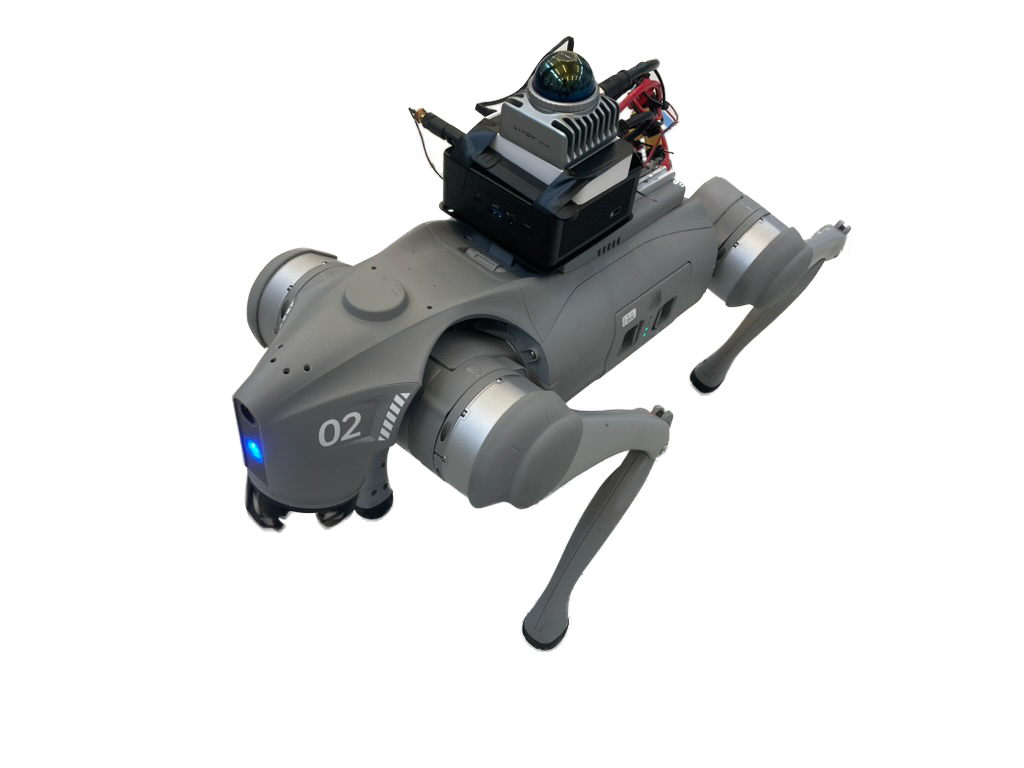}}
    \hfill
    \subfloat{\includegraphics[width=0.63\columnwidth]{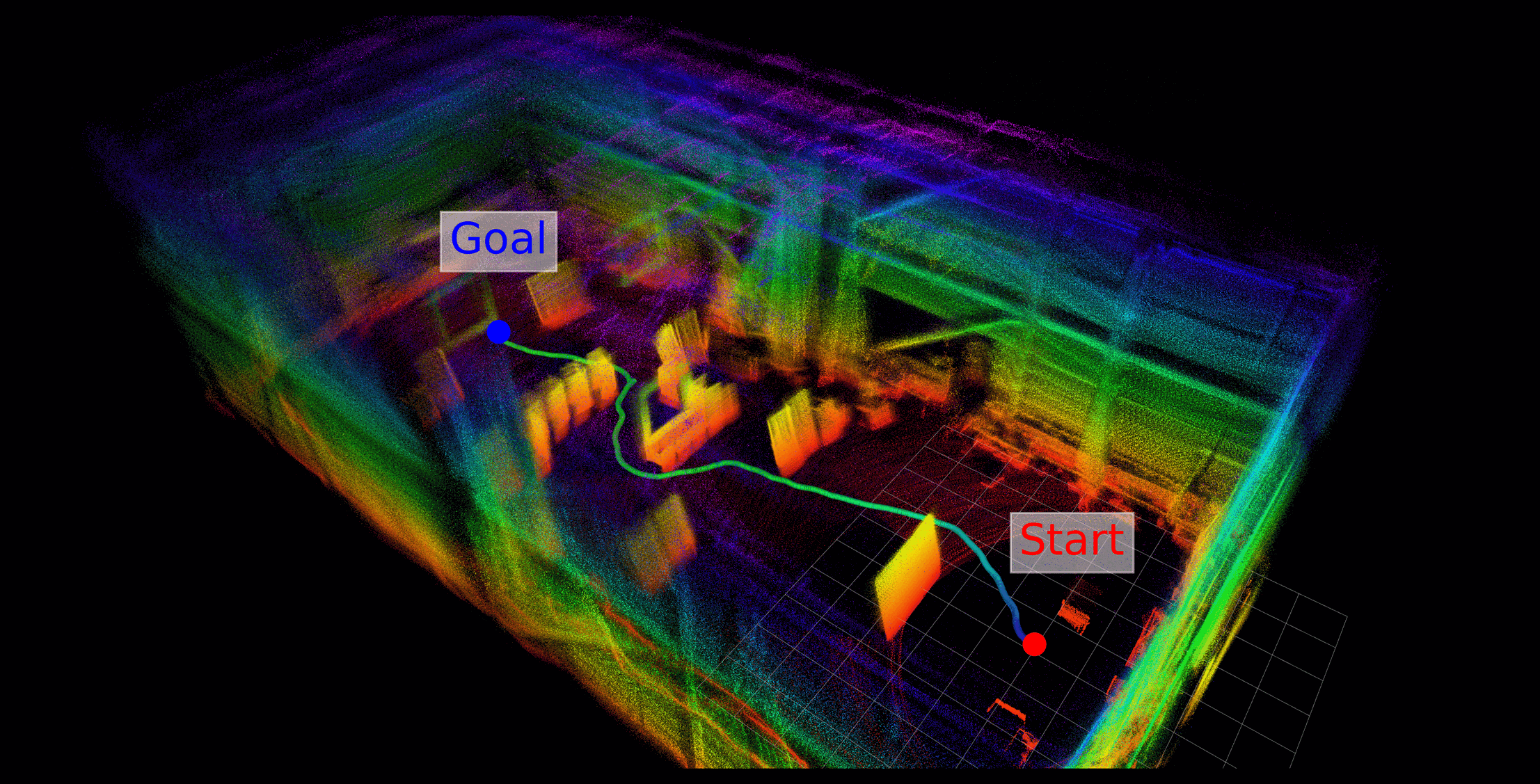}}
    \caption{
        Quadruped Robot in Static Environment.  
        Left: The Unitree Go2 platform.  
        Right: Executed trajectory colored by speed, overlaid on the LiDAR point cloud colored by height.
    }
    \label{fig:hardware_quadruped_static}
    \vspace{-1em}
\end{figure}

We also evaluate DYNUS on ground robots \textemdash a wheeled robot and a quadruped robot \textemdash operating in a static environment.  
Figs.~\ref{fig:hardware_wheeled_static} and \ref{fig:hardware_quadruped_static} show both robots successfully navigating while avoiding static obstacles.  
Each robot is equipped with a Livox Mid-360 LiDAR and an Intel\textsuperscript{\texttrademark} NUC 13.  
As with the UAV experiments, all modules \textemdash including perception, planning, control, and localization \textemdash run onboard in real time, enabling fully autonomous operation.

%% file: paper/12_conclusion.tex
\section{Conclusions}\label{sec:conclusion}

In this paper, we present DYNUS, an uncertainty-aware trajectory planning framework for dynamic, unknown environments.  
DYNUS navigates across diverse settings, including unknown, confined, cluttered, static, and dynamic spaces.  
It integrates a spatio-temporal global planner (DGP), a framework for handling dynamic obstacle unpredictability, and a variable elimination-based local optimizer for fast, safe trajectory generation.  
We validate DYNUS in simulation across forests, office spaces, and caves, and on hardware with UAV, wheeled, and legged robots.  
Future work will implement larger-scale deployments and further computational improvements.